\documentclass[12pt]{article}

\usepackage{tikz}
\usepackage{bm}

\usepackage[]{hyperref}
\newcounter{Chapcounter}

\newcommand{\Prologue}[1] 
{ {\centering          
  \large \underline{\textbf{\color{blue} Prologue:  What this survey is and is not about}} }   
  \addcontentsline{toc}{section}{ \color{blue} Prologue:~What this survey is and is not about}    
}

\newcommand{\chapter}[1] 
{ {\centering          
  \addtocounter{Chapcounter}{1} \large \underline{\textbf{\color{blue} Chapter \theChapcounter: ~#1}} }   
  \addcontentsline{toc}{section}{ \color{blue} Chapter:~\theChapcounter~~ #1}    
}

\hypersetup{colorlinks=true}
\hypersetup{linkcolor=black}

\newcommand{\norm}[1]{\left\lVert#1\right\rVert}

\usepackage[normalem]{ulem}
\usepackage{amsmath}
\usepackage{amssymb}
\usepackage{amsthm}
\usepackage{amscd}
\usepackage{amsfonts}
\usepackage{graphicx}
\usepackage{fancyhdr}
\usepackage{color}
\usepackage{enumerate}
\usepackage{amsfonts}
\usepackage{amsthm}
\usepackage{psfrag} 
\usepackage{epsfig}
\usepackage{subfigure}
\usepackage{graphicx}
\usepackage{amssymb}
\usepackage{epstopdf}
\usepackage{amsbsy}
\usepackage{latexsym}
\usepackage{moreverb}                      
\usepackage{textcomp}

\usepackage{multicol}
\usepackage{enumitem}

\usepackage{algorithm}
\usepackage{algorithmic}

\newtheorem{lemma}{Lemma}
\newtheorem{remark}{Remark}
\newtheorem{proposition}{Proposition}

\newtheorem{theorem}{Theorem}

\newtheorem*{example*}{Example}
\newtheorem{corollary}{Corollary}

\usepackage[top=2.8cm,bottom=2.8cm,left=2.5cm,right=2.5cm]{geometry}
\usepackage[T1]{fontenc}
\usepackage[utf8]{inputenc}

\usepackage[affil-it]{authblk} 
\usepackage{etoolbox}
\usepackage{lmodern}

\makeatletter
\patchcmd{\@maketitle}{\LARGE \@title}{\fontsize{18}{19.2}\selectfont\@title}{}{}
\makeatother

\usepackage{sectsty}
\sectionfont{\fontsize{13}{15}\selectfont}
\subsectionfont{\fontsize{12}{15}\selectfont}

\begin{document}

\title{Approximation Power of Deep Neural Networks \\
\bigskip
{\Large An explanatory mathematical survey}}

\date{\vspace{-5ex}}

\author[1]{Owen Davis\thanks{ondavis@sandia.gov (ORCID: 0009-0002-9818-5279)}}
\author[2]{Mohammad Motamed\thanks{motamed@unm.edu (ORCID: 0000-0002-1421-3694)}}

\affil[1]{Department of Optimization \& Uncertainty Quantification, Sandia National Laboratories, Albuquerque, USA}
\affil[2]{Department of Mathematics and Statistics, University of New Mexico, Albuquerque, USA}
\maketitle

{\centering\normalsize {\it Dedicated to Ra{\'u}l Tempone}\par}

\bigskip

\hrule

\bigskip
\noindent
{\bf Abstract}

\medskip
\noindent
This survey provides an in-depth and explanatory review of the approximation properties of deep neural networks, with a focus on feed-forward and residual architectures. The primary objective is to examine how effectively neural networks approximate target functions and to identify conditions under which they outperform traditional approximation methods. 
Key topics include the nonlinear, compositional structure of deep networks and the formalization of neural network tasks as optimization problems in regression and classification settings. The survey also addresses the training process, emphasizing the role of stochastic gradient descent and backpropagation in solving these optimization problems, and highlights practical considerations such as activation functions, overfitting, and regularization techniques. 
Additionally, the survey explores the density of neural networks in the space of continuous functions, comparing the approximation capabilities of deep ReLU networks with those of other approximation methods. It discusses recent theoretical advancements in understanding the expressiveness and limitations of these networks. A detailed error-complexity analysis is also presented, focusing on error rates and computational complexity for neural networks with ReLU and Fourier-type activation functions in the context of bounded target functions with minimal regularity assumptions. 
Alongside recent known results, the survey introduces new findings, offering a valuable resource for understanding the theoretical foundations of neural network approximation. Concluding remarks and further reading suggestions are provided.

\bigskip

\noindent
{\it keywords:} 
approximation theory of neural networks, mathematics of
deep learning, constructive approximation, error-complexity analysis

\bigskip

\hrule

\bigskip
\noindent
{\bf Intended audience.} The material in this survey should be accessible to undergraduate and graduate students, as well as researchers in mathematics, statistics, computer science, and engineering. It is aimed at anyone seeking a ``deeper'' understanding of ``deep'' neural networks.

\newpage

\tableofcontents

\newpage

\Prologue{}

\bigskip

Three key questions about deep neural networks, in the context of approximation, include:
\begin{enumerate}
\item {\textit{\textbf{Approximation Theory}}}: Given a target function space and a neural network architecture, what is the best the network can do in approximating the target function? Are there function spaces where certain neural network architectures outperform other methods of approximation?

\item {\it Optimal Experimental Design}: Given a target function space,  how can we effectively design the network architecture and generate data samples---by identifying and sampling from an optimal distribution---to achieve the best or near-best approximation with minimal computational effort?

\item {\it Learning Process}: Given a neural network architecture and a data set, how can we efficiently train generalizable networks, and when do optimization techniques like stochastic gradient descent work? Are there alternatives to gradient-based optimization methods?

\end{enumerate}

This survey focuses solely on the first question, offering an explanatory review of deep networks' approximation properties. It is structured into three chapters:

\begin{itemize}
    \item {\bf Chapter 1} introduces key concepts in deep networks, their compositional structure, and formalizes the neural network problem as an optimization problem for regression and classification. The chapter also briefly discusses stochastic gradient descent, backpropagation, activation functions, cost functions, overfitting, and regularization.

    \item {\bf Chapter 2} reviews the density of neural networks in the space of continuous functions. Starting with the concept of density in polynomial approximation, it explores the ability of feedforward networks to approximate continuous functions, including recent advances on deep ReLU networks' approximation capabilities.
    
    \item {\bf Chapter 3} focuses on error-complexity analysis, covering estimates for ReLU and Fourier-activated networks, with an emphasis on results for bounded target functions under minimal regularity assumptions.
\end{itemize}

We aim to provide an accessible yet rigorous explanatory review, balancing foundational insights with advanced topics. To achieve this, the chapters are structured to progressively increase in complexity, accommodating a broad audience that includes senior undergraduates, graduate students, and advanced researchers. 
While our focus is on key concepts and techniques, particularly those related to (residual) feed-forward networks with rectified linear and complex exponential activations, this survey is not an exhaustive account of all network types, architectures, or approximation estimates. Notable exclusions include popular architectures such as convolutional networks, sinusoidal representation networks, and Fourier neural operators, as well as advanced mathematical tools like Rademacher complexity. Instead, our goal is to present a clear and focused narrative that illuminates the fundamental mechanisms driving the success of deep networks, while staying within the constraints of scope and length.

The survey concludes with {\bf Chapter 4}, which offers further reading and final remarks.
\newpage

\chapter{Neural networks: formalization and key concepts}
\label{ch:1}

\bigskip
This chapter introduces the key ideas and
concepts underlying feed-forward neural networks in the context of supervised learning, applied to regression and classification tasks. It begins by outlining the compositional nonlinear structure of feed-forward networks and formulates the network problem as an optimization task. Next, it discusses the (stochastic) gradient descent algorithm and presents the backpropagation formulas used for solving the optimization problem. Finally, the chapter covers various topics concerning the performance of neural networks, including activation and cost functions, overfitting, and regularization.

\section{Feed-forward neural networks}
\label{sec:NN}

An $L$-hidden layer feed-forward neural network (NN) is a parametric map ${\bf f}_{\boldsymbol\theta}: {\mathbb R}^{d} \rightarrow
{\mathbb R}^{n_L}$ 
with parameters $\boldsymbol\theta \in {\mathbb R}^{n_{\theta}}$, formed by the composition of $L+1 \ge 2$ maps 
\begin{equation}\label{comp1}
{\bf f}_{\boldsymbol\theta}({\bf x}) = {\bf f}^{L} \circ \cdots \circ {\bf f}^1 \circ {\bf f}^0({\bf x}), \qquad
{\bf x} \in {\mathbb R}^{d}.
\end{equation}
Each individual map ${\bf f}^{\ell}: {\mathbb R}^{n_{\ell - 1}} \rightarrow
{\mathbb R}^{n_{\ell}}$, with $\ell = 0,1, \dotsc, L$, and $n_{-1}= d$, forms one layer of the network and is given by the component-wise application of a nonlinear activation function $\sigma_{\ell}$ to a multidimensional affine map, 
\begin{equation}\label{comp2}
{\bf f}^{\ell}({\bf z}) = \sigma_{\ell} (M^{\ell} \, {\bf z} + {\bf
  b}^{\ell}), \qquad {\bf z} \in {\mathbb R}^{n_{\ell-1}}, \quad 
M^{\ell} \in {\mathbb R}^{n_{\ell} \times n_{\ell-1}}, \quad
{\bf b}^{\ell} \in {\mathbb R}^{n_{\ell}}, \quad 
\ell = 0,1, \dotsc, L.
\end{equation}
The weight matrices $\{ M^{\ell} \}_{\ell=0}^L$ and bias vectors $\{ {\bf b}^{\ell} \}_{\ell=0}^L$ together form the network parameter vector $\boldsymbol\theta = \{ (M^{\ell}, {\bf b}^{\ell}) \}_{\ell = 0}^{L}$. 
The total number of network parameters is then calculated as $n_{\theta} = \sum_{\ell=0}^L n_{\ell} (n_{\ell - 1} + 1)$. 
Given a set of activation functions, a neural network is uniquely determined by its parameter vector. The network parameters are {\it tuned} during a process referred
to as {\it training} so that ${\bf f}_{\boldsymbol\theta}$ accomplishes a prescribed regression or classification task; we will discuss the training process in Sections \ref{sec:opt}-\ref{sec:solve_opt}.

\begin{figure}[!h]
\begin{center}
\includegraphics[width=0.7\linewidth]{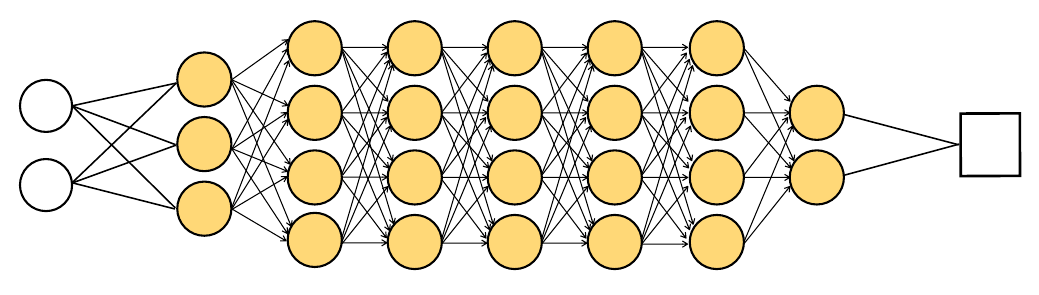}     
\caption{Graph representation of a feed-forward network with $L=7$ hidden layers, and two input and one output neuron. The number of neurons in hidden layers varies between 2 and 4.}
\label{graph_fig}
\end{center}
\end{figure}

The compositional structure of a network, given in \eqref{comp1}-\eqref{comp2}, can be illustrated by a graph. As an example, Figure \ref{graph_fig} depicts a graph representation of a feed-forward network with $L=7$ hidden layers and specific neuron counts: $d=2$, $n_0=3$, $n_1=n_2=n_3=n_4=n_5=4$, $n_6=2$, $n_7=1$.

In a graph representation, nodes correspond to neurons arranged in layers, while edges represent connections linking neurons across adjacent layers, with each edge indicating a scalar weight multiplication. 
Specifically, the network graph consists of an input layer with $d$ nodes, an output layer with $n_L$ nodes, and $L$ hidden layers with $n_0,n_1, \dotsc, n_{L-1}$ nodes, respectively. Input neurons receive the $d$ components of the independent variable vector ${\bf x}=(x_1, \dotsc, x_{d})$, while output neurons generate the $n_L$ components of ${\bf f}_{\boldsymbol\theta} = (f_{\boldsymbol\theta,1}, \dotsc, f_{\boldsymbol\theta,n_L})$.

We define the ``architecture" of a network by considering the number of layers, neurons, and their corresponding activation functions. The number of hidden layers $L$ in a network determines its ``depth"; in general, a higher $L$ indicates a deeper network. The number of neurons $n_{\ell}$ in each layer $\ell$ determines the ``width" of that layer and, consequently, the overall width of the network. Increasing the number of network parameters $n_{\theta}$ can be achieved by making the network wider, deeper, or both. 
It is worth noting that different layers within a network can have varying numbers of neurons and use different activation functions. 
For networks with a uniform width $W$, where $n_{\ell} = W$ for $\ell = 0, 1, \dotsc, L-1$, we find that $n_{\theta}$ scales approximately as $LW^2$. In other words, the number of network parameters increases linearly with the depth $L$ and quadratically with the width $W$.


\section{Applications of neural networks}
\label{sec:problem}

Neural networks, particularly in supervised learning, are extensively employed for solving two main types of problems: regression and classification.

\medskip
\noindent
{\bf Regression:} 
In multivariate regression, we aim to build a mapping ${\bf f}_{\boldsymbol\theta}: {\mathbb R}^{d} \rightarrow {\mathbb R}^{n_L}$, given a set of input-output data $\{ ({\bf x}^{(m)}, {\bf y}^{(m)}) \}_{m=1}^n \in {\mathbb R}^{d} \times {\mathbb R}^{n_L}$. This mapping should approximate ${\bf y}^{(m)}$ for ${\bf x}^{(m)}$, without overfitting, ensuring good performance beyond the provided dataset; see Section \ref{sec:reg} for a discussion on overfitting. Function approximation is a specific instance of regression, where we aim to approximate a target function ${\bf f}: {\mathbb R}^{d} \rightarrow {\mathbb R}^{n_L}$ using a dataset ${ ({\bf x}^{(m)}, {\bf f}({\bf x}^{(m)})) }_{m=1}^n$. 
In other words, our objective is to find a network ${\bf f}_{\boldsymbol\theta}$ such that ${\bf f}_{\boldsymbol\theta}({\bf x}) \approx {\bf f}({\bf x})$.

\medskip
\noindent
{\bf Classification.} 
In multi-classification, data are categorized into $n_L$ distinct classes, labeled as $c_1, \dotsc,
c_{n_L}$. Given an input dataset $\{ {\bf x}^{(m)} \}_{m=1}^n \in {\mathbb
  R}^{d}$, such as a set of images, with their corresponding class labels $\{
c^{(m)} \}_{m=1}^n$, where $c^{(m)} \in \{ c_1, \dotsc,
c_{n_L} \}$, the objective is to construct a mapping ${\bf f}_{\boldsymbol\theta}: {\mathbb
  R}^{d} \rightarrow [0,1]^{n_L}$ that provides the posterior probabilities of class membership for a given pattern ${\bf x} \in {\mathbb R}^{d}$, such as an image. 
  In other words, we aim to find a network ${\bf f}_{\boldsymbol\theta}$ such that ${\bf f}_{\boldsymbol\theta}({\bf x} )\approx (\text{Prob}(c_1 | {\bf
  x}), \dotsc, \text{Prob}(c_{n_L} | {\bf x}))$.


\section{Network training as an optimization problem}
\label{sec:opt}

Network training is the process of finding or tuning the parameters of a network. 
Given $n$ pairs of input-output training data points $\{ ({\bf x}^{(m)}, {\bf y}^{(m)}) \}_{m=1}^n$, the objective is to {\it train} a network with a predefined architecture to {\it learn} the underlying function from the available training data. This entails finding a parametric map ${\bf f}_{\boldsymbol\theta}: {\mathbb R}^{d} \rightarrow {\mathbb R}^{n_L}$, where $\boldsymbol\theta = \{ (W^{\ell}, {\bf b}^{\ell}) \}_{\ell=1}^{L}$ represents the network parameters, such that ${\bf f}_{\boldsymbol\theta}({\bf x})$ approximates the target ${\bf y}$. 
This is done by minimizing a specified loss (or empirical risk) function, denoted by $R_n$, that measures the ``distance'' between ${\bf f}_{\boldsymbol\theta}({\bf x})$ and ${\bf y}$ for the training data set. 
Training can thus be formulated as an optimization problem: find $\boldsymbol\theta$ as the solution of  
\begin{equation}\label{opt}
\min_{\boldsymbol\theta} R_n(\boldsymbol\theta), \qquad R_n(\boldsymbol\theta) = \frac{1}{n}
\sum_{m=1}^n C_m(\boldsymbol\theta), \qquad C_m(\boldsymbol\theta) =
C({\bf y}^{(m)}, {\bf f}_{\boldsymbol\theta} ({\bf x}^{(m)})).
\end{equation}
Here, the cost function $C({\bf a}, {\bf b})$ quantifies the discrepancy between vectors ${\bf a}$ and ${\bf b}$. For instance, a common choice is the quadratic cost function $C({\bf a}, {\bf b}) = || {\bf a} - {\bf b} ||_2^2$.

It is to be emphasized that training typically refers to the task of determining $\boldsymbol\theta$, given training data, network architecture, and the cost function. However, in practice, selecting training data (if not provided), defining network architecture, and choosing the cost function may also be viewed as integral parts of the training process.


\section{Solving the optimization problem}
\label{sec:solve_opt}

The optimization problem \eqref{opt} is often solved by a gradient-based method, such as stochastic gradient
descent \cite{SGD1,SGD2} or Adam \cite{Adam:17}. These methods compute the gradient of the loss function with respect to the network parameters through the chain rule, using a differentiation technique known as {\it backpropagation} \cite{BackProp:1986}. In this section, we will provide a concise overview of stochastic gradient descent and backpropagation. For more comprehensive insights, we refer to \cite{Bottou_etal:18}.

\subsection{Gradient descent approach}
\label{sec:gd}

Gradient descent (GD), also known as steepest descent, is an iterative optimization method. Given a fixed network architecture with unknown parameters $\boldsymbol\theta$ and a data batch $\{ ({\bf x}^{(m)}, {\bf y}^{(m)}) \}_{m=1}^n$ within the context of multivariate regression, GD aims to minimize the empirical risk function $R_n(\boldsymbol\theta)$, given in \eqref{opt}, through an iterative process. 
Starting with an initial guess $\boldsymbol\theta^{(0)}$ for the minimizer, GD consecutively updates $\boldsymbol\theta^{(k)}$, the set of parameters at iteration level $k \ge 0$, moving in the negative direction of the gradient of the empirical risk at $\boldsymbol\theta^{(k)}$,
\begin{equation}\label{gd}
\boldsymbol\theta^{(k+1)} = \boldsymbol\theta^{(k)} - \eta \,
\nabla_{\boldsymbol\theta} R_n(\boldsymbol\theta^{(k)}), \qquad k=0,
1, 2, \dotsc.
\end{equation}
Here, $\eta>0$ represents the ``learning rate'', a hyper-parameter that can be preselected or tuned through validation; see Section \ref{sec:validation}. As $\boldsymbol\theta^{(k)}$ gets updated, we monitor the value of
$R_n(\boldsymbol\theta^{(k)})$, expecting it to decrease with increasing $k$. The iteration continues until reaching a level, say $K$, where $R_n(\boldsymbol\theta^{(K)})$ is sufficiently small (below a specified tolerance). If $R_n(\boldsymbol\theta^{(k)})$ decreases slowly or even worse if it is increasing, adjusting the learning rate $\eta$ may be necessary.

\subsection{Backpropagation}
\label{sec:backprop}

The above gradient-based process involves computing the gradient $\nabla_{\boldsymbol\theta} R_n(\boldsymbol\theta^{(k)})$. This is primarily achieved through the backpropagation algorithm. 
Noting that the parameter set $\boldsymbol\theta$ consists of weights $\{ M^{\ell} \}_{\ell=0}^L$ and biases $\{ {\bf b}^{\ell} \}_{\ell=0}^L$, the backpropagation algorithm computes
$\nabla_{M^{\ell}} C_m(\boldsymbol\theta^{(k)})$ and $\nabla_{{\bf b}^{\ell}} C_m(\boldsymbol\theta^{(k)})$ for all
$\ell=0,1, \dotsc, L$. These computations are performed given a set of parameters $\boldsymbol\theta^{(k)}$ at iteration $k \ge 0$. 
This computation is repeated for all training data points $\{ ({\bf x}^{(m)}, {\bf y}^{(m)}) \}_{m=1}^n$ and averaged to compute:\begin{equation}\label{empirical_risk_gd}
\nabla_{M^{\ell}} R_n (\boldsymbol\theta^{(k)})= \frac{1}{n} \, \sum_{m=1}^n \nabla_{M^{\ell}}
C_m(\boldsymbol\theta^{(k)}), \qquad \nabla_{{\bf b}^{\ell}} R_n(\boldsymbol\theta^{(k)}) = \frac{1}{n} \, \sum_{m=1}^n
\nabla_{{\bf b}^{\ell}} C_m(\boldsymbol\theta^{(k)}).
\end{equation}
In practice, we approximate the sample averages \eqref{empirical_risk_gd} by randomly selecting a small batch of $n_b \ll n$ data points, a technique known as mini-batch stochastic GD; see Section \ref{sec:sgd-backprop}.

Before discussing the derivation of backpropagation algorithm, we establish some notations.

\begin{itemize}
\item $M_{j,k}^{\ell}$: weight of the connection from neuron $k$ in
  layer $\ell-1$ to neuron $j$ in layer $\ell$

\item $b_j^{\ell}$: bias of neuron $j$ in layer $\ell$

\item $z_j^{\ell} = \sum_{k=1}^{n_{\ell-1}} M_{j,k}^{\ell} \,
  a_k^{\ell -1} + b_j^{\ell}$: input of neuron $j$ in layer $\ell$

\item $a_j^{\ell} = \sigma_{\ell} (z_j^{\ell})$: output of neuron $j$ in layer $\ell$

\item $\delta_j^{\ell} : = \frac{\partial C_m}{\partial z_j^{\ell}}$: sensitivity of the cost function $C_m$ to changes in the $j$-th neuron in layer $\ell$.
\end{itemize} 
In matrix notation, we write
$$
M^{\ell} = ( M_{j,k}^{\ell} )  \in {\mathbb R}^{n_{\ell} \times n_{\ell-1}}, 
\ \ 
{\bf b}^{\ell} = ( b_{j}^{\ell} )  \in {\mathbb R}^{n_{\ell}}, 
\ \
{\bf z}^{\ell} =  ( z_{j}^{\ell} )  \in {\mathbb R}^{n_{\ell}},
\ \
{\bf a}^{\ell} = ( a_{j}^{\ell} )  \in {\mathbb R}^{n_{\ell}},
\ \
{\boldsymbol\delta}^{\ell} = (\delta_j^{\ell}) \in
{\mathbb R}^{n_{\ell}},
$$
$$
{\bf z}^{\ell} = M^{\ell}  \, {\bf a}^{\ell-1} + {\bf b}^{\ell},
\qquad
{\bf a}^{\ell} = \sigma_{\ell} ({\bf z}^{\ell}).
$$

We compute gradients $\nabla_{M^{\ell}} C_m$ and $\nabla_{{\bf b}^{\ell}} C_m$, evaluated at $\boldsymbol\theta^{(k)}$ for any fixed $m$, in three steps.

\medskip
\noindent
{\bf Step 1.} Compute sensitivity for the last layer using the chain rule and noting that $a_j^L = \sigma_L (z_j^L)$:
$$
\delta_j^L = \frac{\partial C_m}{\partial z_j^{L}} = \frac{\partial
  C_m}{\partial a_j^{L}} \, \sigma_{L}'(z_j^L).
$$
The terms $\partial C_m / \partial a_j^{L}$ and $\sigma_{L}'$ are obtained analytically, given the form of the cost function and the output activation function. Additionally, $z_j^L$ is computed by a
forward sweep, given $\boldsymbol\theta^{(k)}$.

\medskip
\noindent
{\bf Step 2.} Backpropagate sensitivities by expressing $\delta_j^{\ell}$ in terms of $\boldsymbol\delta^{\ell+1}$, for $\ell = L-1, \dotsc, 1,0$:
$$
\delta_j^{\ell} = \sum_{i =1}^{n_{\ell+1}} \delta_i^{\ell+1} \,
M_{i,j}^{\ell+1} \, \sigma_{\ell}'(z_j^{\ell}).
$$
Here, the weights $M_{i,j}^{\ell+1} $ are available via the given
$\boldsymbol\theta^{(k)}$, and neuron inputs $z_j^{\ell}$ are
computed by the forward sweep. This expression is derived using
the chain rule:
$$
\delta_j^{\ell} = \frac{\partial C_m}{\partial z_j^{\ell}} = \sum_{i =
1}^{n_{\ell+1}} \frac{\partial C_m}{\partial z_i^{\ell+1}} \,
\frac{\partial z_i^{\ell+1}}{\partial z_j^{\ell}} = \sum_{i =
1}^{n_{\ell+1}} \delta_i^{\ell+1} \,
\frac{\partial z_i^{\ell+1}}{\partial z_j^{\ell}},
$$
where $\partial z_i^{\ell+1} / \partial z_j^{\ell} = M_{i,j}^{\ell+1} \, \sigma_{\ell}'(z_j^{\ell})$ noting
$z_i^{\ell+1} = \sum_{j=1}^{n_{\ell}} M_{i,j}^{\ell+1} \,
  a_j^{\ell} + b_i^{\ell+1} = \sum_{j=1}^{n_{\ell}} M_{i,j}^{\ell+1} \,
  \sigma_{\ell}(z_j^{\ell}) + b_i^{\ell+1}$.

\medskip
\noindent
{\bf Step 3.} Having computed all sensitivity ratios $\boldsymbol\delta^0, \boldsymbol\delta^1,
\dotsc, \boldsymbol\delta^L$, compute gradients:
$$
\frac{\partial C_m}{\partial b_j^{\ell}} = \frac{\partial
  C_m}{\partial z_j^{\ell}} \, \frac{\partial z_j^{\ell}}{\partial
  b_j^{\ell}}  = \frac{\partial
  C_m}{\partial z_j^{\ell}} = \delta_j^{\ell},
  \qquad 
  \frac{\partial C_m}{\partial M_{j,k}^{\ell}} = \frac{\partial
  C_m}{\partial z_j^{\ell}} \, \frac{\partial z_j^{\ell}}{\partial
  M_{j,k}^{\ell}} = \delta_j^{\ell} \, a_k^{\ell-1},
$$ 
noting that $\partial z_j^{\ell} / \partial b_j^{\ell}=1$ and $\partial z_j^{\ell} / \partial
  M_{j,k}^{\ell} = a_k^{\ell-1}$.

\medskip

The pseudocode for backpropagation is given in Function 1. Backward movement in the algorithm is a natural consequence of the fact that the loss is a function of the network's output. By employing the chain rule, we need to move backward to compute all gradients.

\begin{algorithm}[!ht]

\renewcommand{\thealgorithm}{}
\floatname{algorithm}{}

\noindent
\caption{Function 1: Backpropagation algorithm}

{\bf function} $\{ \nabla_{{\bf b}^{\ell}} C_m , \,
\nabla_{{M}^{\ell}} C_m \}_{\ell=0}^L=$ BACKPRO(${\bf x}^{(m)}$, ${\bf
  y}^{(m)}$, $\, \{ (M^{\ell} ,  {\bf b}^{\ell} )\}_{\ell=0}^L , \, \{
\sigma_{\ell} \}_{\ell=0}^L, \,$ ``$C_m$'')

\begin{algorithmic} 
\medskip
\STATE
\begin{itemize} [topsep=1pt,leftmargin=0.7\labelsep]
\setlength\itemsep{-.1em}
\item[1:] Set ${\bf a}^{-1} = {\bf x}^{(m)}$ for the input
  layer.  

\medskip
\item[2:] {\bf Forward pass:} For each $\ell=0,1, \dotsc,
  L$ compute ${\bf z}^{\ell} = M^{\ell} {\bf a}^{\ell-1} + {\bf
    b}^{\ell}$ and ${\bf a}^{\ell} = \sigma_{\ell} ({\bf z}^{\ell})$.

\medskip
\item[3:] {\bf Loss:} Using the given function
  for loss 
  ``$C_m$'' and ${\bf y}^{(m)}$ and
  ${\bf a}^L$, compute $\nabla_{{\bf a}^L}C_m$. 

\medskip
\item[4:] {\bf Sensitivity:} Compute $\boldsymbol\delta^L = \nabla_{{\bf a}^L} C_m \odot \sigma_{L}'({\bf z}^L)$, where $\odot$ denotes element-wise multiplication.

\medskip
\item[5:] {\bf Backpropagate:} For each $\ell
  = L-1, \dotsc, 1, 0$ compute 
$\boldsymbol\delta^{\ell} = \bigl( {M^{\ell+1}}^{\top}  \,
\boldsymbol\delta^{\ell+1}  \bigr) \odot  \sigma_{\ell}'({\bf z}^{\ell})$.

\medskip
\item[6:] {\bf Outputs:} Compute 
$\nabla_{{\bf b}^{\ell}} C_m = \boldsymbol\delta^{\ell}, \ 
\nabla_{M^{\ell}} C_m = \boldsymbol\delta^{\ell} \, {{\bf a}^{\ell-1}}^{\top}$, for $\ell=0,1, \dotsc, L$.
  
\end{itemize}

\end{algorithmic}

\end{algorithm}

The computational complexity of backpropagation is primarily due the forward and backward passes (steps 2 and 5) of the algorithm. The cost of these steps is in turn contingent upon the number of layers and neurons in each layer, as well as the computational cost of activation functions. If the evaluation of each activation function or its derivative requires $N$ operations, the number of operations in steps 2 and 5 is $\sum_{\ell=0}^L n_{\ell}
(n_{\ell-1} + N + 1)$ and $\sum_{\ell=0}^L n_{\ell} (n_{\ell+1} + N + 1)$, respectively. 
If we further assume that the number of neurons in
each layer is a constant $W$, and that $N$ is also of the order of $W$, then the total cost of the algorithm is ${\mathcal O}(L \, W^2)$.

\subsection{Mini-batch stochastic GD and backpropagation}
\label{sec:sgd-backprop}

In combining the backpropagation algorithm with GD, one approach to compute the gradient of the empirical risk $\nabla_{\boldsymbol\theta} R_n(\boldsymbol\theta^{(k)})$, used in GD formula \eqref{gd}, is to apply the backpropagation algorithm $n$ times, each for
one training data point. Then one can take the sample average of
gradients for all $n$ data points to obtain the gradient of the
empirical risk by \eqref{empirical_risk_gd}. This leads to the
standard GD, for layers $\ell =0,1, \dotsc, L$, and with iterations $k=0, 1, \dotsc, K-1$:
\begin{equation}\label{gd2}
{M^{\ell}}^{(k+1)} = {M^{\ell}}^{(k)} - \eta \,
\nabla_{M^{\ell}} R_n (\boldsymbol\theta^{(k)}), \qquad 
{{\bf b}^{\ell}}^{(k+1)} = {{\bf b}^{\ell}}^{(k)} - \eta \,
\nabla_{{\bf b}^{\ell}} R_n (\boldsymbol\theta^{(k)}).
\end{equation}

A more efficient strategy is to replace GD with mini-batch stochastic GD (SGD), wherein a small subset of $n_b \ll n$ training data points, known as a mini-batch, is randomly selected from the entire dataset, and a gradient descent step is applied based on this mini-batch:

\begin{itemize}

\item Divide the set of $n$ training data points into $\lfloor n/n_b \rceil$ mini-batches of
  size $n_b \ll n$, where $\lfloor . \rceil$ means that we either take the floor or the ceil of the number depending on the number.

\item Loop over all mini-batches containing $n_b$ data points (note
  that the last batch may have fewer or larger number of points than $n_b$); 

\begin{itemize}

\item[$\circ$] For each training data point in the mini-batch,
  i.e. for $m=1, \dotsc, n_b$, call the backpropagation algorithm (BACKPRO) and compute $\{ \nabla_{{\bf b}^{\ell}} C_m , \,
\nabla_{{M}^{\ell}} C_m \}_{\ell=0}^L$;

\item[$\circ$] Approximate the gradients by taking the sample average
  of all $n_b$ gradients:
$$
\nabla_{M^{\ell}} R_n (\boldsymbol\theta^{(k)}) \approx \frac{1}{n_b} \, \sum_{m=1}^{n_b} \nabla_{M^{\ell}}
C_m(\boldsymbol\theta^{(k)}), \qquad \nabla_{{\bf b}^{\ell}} R_n(\boldsymbol\theta^{(k)}) \approx \frac{1}{n_b} \, \sum_{m=1}^{n_b}
\nabla_{{\bf b}^{\ell}} C_m(\boldsymbol\theta^{(k)});
$$

\item[$\circ$] Update the weights and biases according to \eqref{gd2}, using the approximate gradients.
\end{itemize}

\end{itemize}

The above procedure is referred to as one-{\it epoch} SGD, 
involving a single run of SGD over all $n$ data points with a specific mini-batch selection. In practice, this process is repeated for multiple epochs of training. This requires an outer loop iterating through $n_e$ epochs, wherein for each epoch, the training data points are randomly shuffled before being divided into mini-batches. 
Noting that each epoch of SGD involves $\lfloor n/n_b \rceil$ mini-batches, and each mini-batch forms one iteration of SGD, the total number of SGD iterations reads $K = n_e \lfloor n/n_b \rceil$. 
Implementing full SGD requires an initial guess for the parameters, denoted as ${M^{\ell}}^{(0)}$ and ${{\bf b}^{\ell}}^{(0)}$ for $\ell=0, 1, \dotsc, L$. Typically, this initialization is achieved through pseudo-random number generation.


\section{Activation functions}
\label{sec:activation}

Activation functions play a crucial role in neural networks by introducing nonlinearity, allowing the network to learn complicated patterns within data. Note that by solely stacking linear layers without applying an activation function, the network can produce polynomial nonlinearity. However, polynomials may not be sufficiently versatile or complex to capture intricate patterns or model complex functions. For the same reason, polynomial activation functions are also avoided. 
Activation functions further offer the capability to to limit and control a neuron's output as required. This aspect becomes particularly important in scenarios where the network output must adhere to specific constraints, such as being positive or bounded within a certain range, e.g., on [0,1].

In addition to the aforementioned features, activation functions should possess several desirable properties. Firstly, since activation functions are applied multiple times, typically $\sum_{\ell=0}^L n_{\ell}$ times in each forward pass, they should be computationally efficient to calculate, requiring only a few operations. Secondly, they should be (almost everywhere) differentiable to facilitate the computation of (generalized) gradients of the loss function; refer to \cite{Clarke:1990} for convergence analysis of gradient-based optimization methods, particularly in cases where activation functions are not differentiabile everywhere, such as Rectified Linear Unit (ReLU) functions. 
Moreover, activation functions should avoid the issue of vanishing gradients. In the backpropagation algorithm, multiple applications of the chain rule occur. With each multiplication of $\sigma'$, we have $\boldsymbol\delta^{\ell} \sim {(\sigma')}^{L+1 - \ell}$, as seen in the final formulas in Steps 1-2 in Section \ref{sec:backprop}. For example, if $\sigma_{\ell}'$ takes values between 0 and 1 across many layers, or if $\sigma_{\ell}'$ at a certain layer is close to zero (e.g., when $\sigma_{\ell}$ is flat), then $\boldsymbol\delta^{\ell}$ and consequently the gradient values at initial layers (with small $\ell$) become very small. Consequently, the weights and biases of these initial layers would learn slowly, leading to the vanishing gradient problem. To mitigate this issue, it is preferable to use an activation function that avoids diminishing gradients, such as an increasing function whose derivative remains greater than or equal to 1 over a large part of its domain.

Different layers of a network may use different activation functions. A few widely recognized activation functions includes: 
\begin{enumerate}
\item Identity: $\sigma(x)=x$.

\item Softmax: $\sigma(x_1, \dotsc, x_{n_{L}}) = (p_1, \dotsc, p_{n_L}), \quad  p_j =
  \frac{\exp(x_j)}{\sum_{j=1}^{n_L} \exp(x_j)}, \quad j=1, \dotsc,
  n_L$.

\item Sigmoid: $\sigma(x) = 1/ (1+ \exp(- x))$.

\item Hyperbolic tangent: $\sigma(x) = \tanh (x)$.

\item ReLU: $\sigma(x) = \max (0, x)$.

\item Leaky ReLU: $\sigma(x) = \max (\alpha \, x, x)$, where  $\alpha \in (0,1)$.
\end{enumerate}

In regression problems, the identity function is commonly employed in the output layer. 
In classification problems, the softmax function is typically used in the output layer. It transforms the network's output into a set of pseudo-probabilities $p_1, \dotsc, p_{n_L}$ with $\sum_{j=1}^{n_L} p_j = 1$. Each $p_j$ approximates the posterior probability $\text{Prob}(c_j | \text{data})$, representing the likelihood of the  $j$-th class. Ultimately, the class with the highest probability is selected.

Sigmoid and hyperbolic tangent functions are relatively inexpensive to compute but are susceptible to the vanishing gradient problem. Both functions are rather flat (with $\sigma '$ close to zero) over a large part of their domain. Specifically, for the sigmoid function, $0, \le \sigma' \le 1/4$, while for the hyperbolic tangent function, $0 \le \sigma' \le 1$. While these functions remain in use, their popularity has decreased in favor of ReLU.

ReLU stands out as one of the most favored activation functions in contemporary neural networks. Its computational efficiency and avoidance of the vanishing gradient issue make it highly desirable. ReLU's (weak) derivative is 0 for negative inputs and 1 for non-negative ones, but it introduces a phenomenon known as {\it dying ReLU}. This occurs when ReLU outputs zero for negative inputs, potentially causing some neurons to remain inactive. Despite this, {\it dying ReLU} contributes to model sparsity, akin to the sparsity observed in biological neural networks: among billions of neurons in a human brain, only a portion of them fire (i.e. are active) at a
time for a particular task. Sparse networks offer advantages: they are more concise, less prone to overfitting, and faster to compute due to fewer operations. 
The downside is the risk of some neurons becoming permanently inactive if they consistently receive negative inputs. However, this scenario is less likely in practice, as SGD typically involves multiple data points, providing opportunities for non-negative inputs.

Leaky ReLU offers also a solution to the {\it dying ReLU} problem by introducing a small slope in the negative range. Typically, the parameter $\alpha$ ranges between 0.01 and 0.05. Notably, setting $\alpha$ to 0 results in the standard ReLU function, while setting it to 1 yields the identity function.

\section{Loss functions}
\label{sec:cost_fun}

\noindent
{\bf Quadratic loss.} A common loss function in regression problems is the quadratic loss, 
$$
C_m(\boldsymbol\theta) = \frac{1}{2} \,
\sum_{j=1}^{n_L} (y_j^{(m)} - a_j^{L}(\boldsymbol\theta; {\bf x}^{(m)}) )^2.
$$

\medskip
\noindent
{\bf Cross-entropy loss.} The cross-entropy loss function is often used in classification problems. 
Given a set of labeled data $\{ ( {\bf x}^{(m)}, c^{(m)})
\}_{m=1}^n$ with class labels $c^{(m)} \in \{ c_1, \dotsc,
c_{n_L}  \}$, the cross-entropy loss reads
\begin{equation}\label{cross_entropy_cost} 
C_m(\boldsymbol\theta) = - \,
\sum_{j=1}^{n_L}  \delta_{m,j} \, \log
a_j^{L}(\boldsymbol\theta; {\bf x}^{(m)}), 
\ \
a_j^{L} = \text{Prob}(c_j
| {\bf x}^{(m)}),
\ \
\delta_{m,j} = 
\left\{ \begin{array}{l l}
1 & \ \text{if} \ \ c^{(m)} = c_j, \\
0 &  \ \text{otherwise}.
\end{array} \right.
\end{equation}
Here, $\delta_{m,j}$ is a 0-1 binary variable that indicates the ``true'' distribution of class membership, and $a_j^L$ is the $j$-th output of the softmax
activation function applied to the last layer of the network, indicating the ``predicted'' distribution. The cross-entropy loss is
related to the
Kulbacl-Leibler divergence between the true and predicted
distributions. It is easy to see that $C_m(\boldsymbol\theta) = - \, \log
( \text{Prob}(c^{(m)}
| {\bf x}^{(m)}) )$, 
that is, for each $m$, we find the output index $j$ for which $c_j =
c^{(m)}$.

\section{Overfitting and regularization}
\label{sec:reg}

The training strategy discussed so far may suffer from
{\it overfitting} (or overtraining), that is, the trained network ${\bf
  f}_{\boldsymbol\theta}({\bf x})$ may not perform well
in approximating ${\bf y}$ outside the set of training data $\{ ({\bf
  x}^{(m)}, {\bf y}^{(m)}) \}_{m=1}^n$.

\begin{figure}[!h]
\begin{center}
\includegraphics[width=0.45\linewidth]{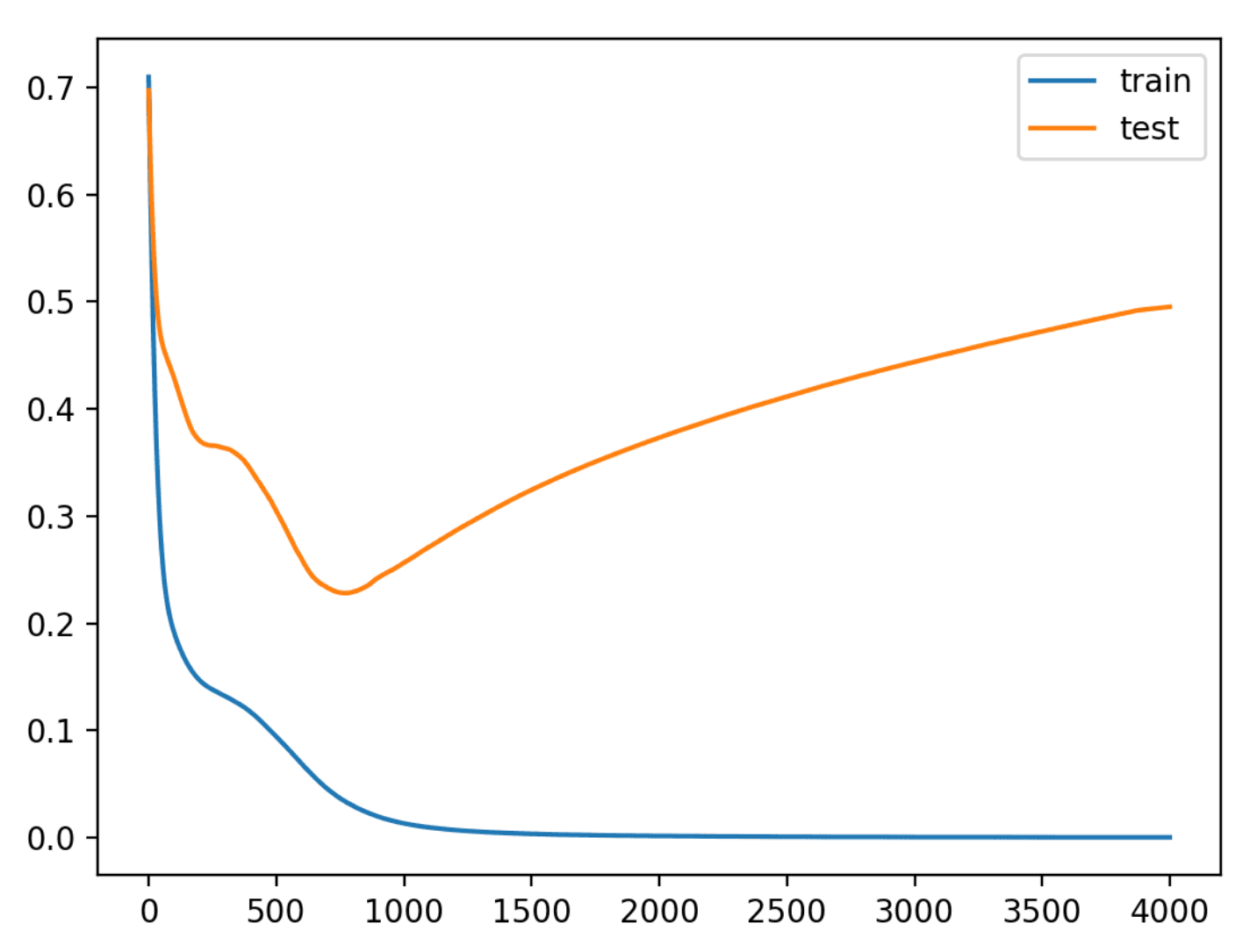}    
\caption{The loss on training data versus test data as the
  number of epochs increases.}
\label{monitor_loss_fig}
\end{center}
\end{figure}

The first step in avoiding overfitting is to detect overfitting. For this purpose, we usually split the available data into two categories: training data and test data (or sometimes validation data). 
While the network undergoes training using the training data, we monitor how the loss on this set evolves. Simultaneously, we track the loss on the test data to gauge the network's generalization ability. A scenario depicted in Figure \ref{monitor_loss_fig} illustrates a common observation: the loss on the training data decreases as the network learns, indicating a good fit to this dataset. However, the loss on the test data initially decreases but later starts to rise. This phenomenon signifies overfitting and suggests a lack of generalization.

Reducing overfitting can be achieved by increasing the amount of training data or by simplifying the network's complexity, i.e., decreasing the number of network parameters. 
However, these approaches are often impractical. Augmenting the training dataset may not be feasible due to limitations in availability or cost. Additionally, networks with a larger parameter count, such as deep networks, possess greater potential for power, making complexity reduction less desirable; we will
discuss the power of ``depth'' in more details in the next chapters. Fortunately, there are alternative strategies to mitigate overfitting within fixed constraints of data points and network parameters. These include introducing regularization terms to the loss function, early stopping, and employing dropout.

\subsection{Adding penalty terms}
\label{sec:reg1}

In this approach, we modify the loss function by adding a regularizing (or penalty) term. 
Commonly used penalty terms include the $L^1$-norm and $L^2$-norm of all weights (parameters excluding biases), scaled by a factor $\lambda$. This yields the regularized loss:
$$
R_n(\boldsymbol\theta) = \frac{1}{n} \sum_{m=1}^n
C_m(\boldsymbol\theta) + \lambda \, || \boldsymbol\theta_w ||^2.
$$ 
Here, $\boldsymbol\theta_w$ represents the vector of all weights excluding biases, and $|| . ||$ denotes either the $L^1$ or $L^2$ norm. The parameter $\lambda >0$ serves as the regularization parameter, a hyper-parameter that requires tuning.

This type of regularization introduces a compromise between minimizing the original loss and reducing weights, with $\lambda$ determining the balance. Smaller weights, encouraged by regularization, result in more stable outputs with less sensitivity to small changes in input, akin to simpler models which are more generalizable. With small weights, the network's output would not change much if we make small changes in the input. Additionally, regularization promotes convexity in the loss function, making optimization algorithms more robust against local minima and less dependent on the initial guess. Biases are excluded from regularization because large biases do not significantly affect network sensitivity to input.

The difference between $L^1$ and  $L^2$ regularization lies in how weights decrease during each iteration of GD/SGD. For a single weight $w:= M_{j,k}^{\ell}$, the updates are as follows:
\begin{itemize}
\item $L^1$ regularization: \ \ \ 
$w^{(k+1)} = w^{(k)} - \frac{\eta}{n} \sum_{m=1}^n \frac{\partial C_m}{\partial w} -  \eta \, \lambda \, \text{sgn}(w^{(k)})$,

\item $L^2$ regularization: \ \ \ 
$w^{(k+1)} = w^{(k)} - \frac{\eta}{n} \sum_{m=1}^n \frac{\partial C_m}{\partial w} - 2 \, \eta \, \lambda \, w^{(k)}$.
\end{itemize}
In both cases, the penalty terms cause $|w|$ to decrease. However, $L^1$ regularization decreases $|w|$ by a fixed amount, whereas $L^2$ regularization decreases $|w|$ proportionally to $|w|$. As a result, $L^2$ regularization reduces large weights more than $L^1$ regularization does, leading to a more balanced weight distribution resembling a mean, while $L^1$ regularization tends to keep some weights of high importance, akin to a median. 
Moreover, $L^2$ regularization maintains non-zero small weights, resulting in a dense weight vector, while $L^1$ regularization induces sparsity by driving small weights to zero.

\subsection{Early stopping}
\label{sec:reg2}

In this approach, we divide the data into three sets: training, test, and validation. The loss on the validation set is monitored at the end of each epoch. Training stops when the validation loss no longer decreases, indicating that further training may lead to overfitting. Refer to Section \ref{sec:validation} for a more detailed utilization of the validation set.

\subsection{Dropout}
\label{sec:reg3}

In the dropout approach \cite{Dropout:14}, we modify the network training process, rather than modifying the loss function. 
During each mini-batch iteration, we randomly deactivate a portion of hidden neurons, e.g. half. These deactivated neurons, along with their connections, are temporarily removed from the network. Forward and backward passes are then performed on the modified network, updating weights and biases only for the active neurons. This process is repeated for each mini-batch, with restoring the dropout neurons in the previous mini-batch and randomly selecting a new set of dropout neurons. After training, dropout is no longer used during prediction. Since the full network has twice as many active hidden neurons during training, the weights ``outgoing'' from these neurons are halved. 
More generally, during training, each hidden neuron is retained with a fixed probability $p$, and during testing, the outgoing weights of each hidden neuron are scaled by $p$. We note that to effectively remove a neuron from the network, its output weight is multiplied by zero. Hence, only the output weights are scaled by the selected rate, while biases remain unchanged. 
Conceptually, dropout simulates training numerous sparse neural networks in parallel, introducing regularization through the averaging of diverse model behaviors: different models may overfit in different ways, and hence averaging helps mitigate overfitting.

\section{Validation and hyper-parameter tuning}
\label{sec:validation}

Every optimization/regularization technique involves hyper-parameters such as learning rates, number of epochs, mini-batch sizes, dropout rates, etc. These hyper-parameters are often optimized using a validation set, separate from the training data. For further details, refer to \cite{Bengio:12,GoodBengCour16}.

\medskip
\noindent
{\bf Learning rate $\eta$.} 
The learning rate determines the step size in GD/SGD. If $\eta$ is too large, overshooting the minimum is likely, while a too small $\eta$ slows down convergence. Typically, we start with a small $\eta$ and adjust it based on the training-validation loss behavior. We often decrease $\eta$ gradually as training progresses, especially after validation loss stops decreasing. 
Note that decreasing $\eta$ is natural because we anticipate getting closer to the minimum as training advances. 
This adaptive approach helps find an optimal learning rate.

\medskip
\noindent
{\bf Number of epochs $n_e$}. Early stopping is often employed to determine the number of epochs for training. By monitoring the validation loss at the end of each epoch, we stop training when the validation loss exhibits consistent stagnation or increase for a user-defined number of epochs---an indicator of potential overfitting. While early stopping offers a convenient way to automate the training duration, it should not serve as the primary method of regularization. Neural network loss landscapes are often highly complex, with many sub-optimal local minima. To ensure robustness against getting trapped in such minima, early stopping must be complemented with effective learning rate adaptation and well-designed loss functions. When used alongside other regularization techniques, it not only streamlines the training process but also helps mitigate overfitting.

\medskip
\noindent
{\bf Regularization parameter $\lambda$}. Similar to $\eta$, we initially select a value for $\lambda$ based on a few epochs of training. We then adjust and fine-tune $\lambda$ as necessary by monitoring the validation loss. Specifically, if validation loss stagnates or begins to increase, a sign of overfitting, then $\lambda$ can be increased to compensate.

\medskip
\noindent
{\bf Mini-batch size $n_b$}. In general, smaller batch sizes accelerate parameter updates but may slow down learning due to less accurate gradient approximations. 
Noting that the mini-batch size is relatively independent of other hyper-parameters, to strike a balance between speed and convergence we typically start with a set of reasonable hyper-parameters and then use validation data to optimize the mini-batch size for the fastest performance improvement in terms of CPU-time. To this end, by plotting the validation error against CPU-time for various batch sizes, we can identify the optimal mini-batch size before fine-tuning other hyperparameters.

\medskip
\noindent {\bf Network architecture:} 
The network architecture (width and depth) is a key factor that influences the quality of the network's approximation. 
While it is not necessarily a hyperparameter, it can be treated as one, as the choice of architecture may affect the selection of other hyperparameters. 
A common approach is to use an overparameterized network, with more training parameters than training samples. This strategy has been supported by empirical success, though the mathematical investigation of this phenomenon remains an exciting research area. However, we will not address further architectural considerations here, as they can be viewed as part of optimal experimental design. For more details, we refer readers to
\cite{nakkiran2021deep, allen2019convergence}.


\section{Procedure summary}
\label{sec:summary}

A concise summary of the general procedure follows:
\begin{itemize}[topsep=3pt,leftmargin=2.5\labelsep]
\setlength\itemsep{-.05em}

\item Data: 
Split the data into three non-overlapping sets: training, validation, and test sets, such as a $70\% - 20\% - 10 \%$ split, ensuring their representativeness for the problem

\item Architecture: Select the number of layers, neurons, and activation functions based on the problem requirements.

\item Training: 
Utilize the training set for computing cost gradients and the validation set for hyper-parameter tuning to train the network parameters.

\item Evaluation: 
Assess the trained network's performance using the test set. If satisfactory, it is deemed ready for prediction.

\item Prediction: 
Deploy the trained network for making predictions.

\end{itemize}


\newpage

\chapter{Approximation theory and ReLU networks}

\bigskip
This chapter reviews classical concepts and recent advancements in approximating functions using neural networks with depth $L=1$ and $L\geq 2$, respectively referred to as shallow and deep networks. It primarily focuses on the density of neural networks in the space of continuous functions, examining the ability of feed-forward neural networks to approximate continuous functions. Beginning with an overview of the concept of density in polynomial approximation, it investigates the density of shallow feed-forward networks and discusses recent developments in understanding the approximation capabilities of deep ReLU networks.

\section{Main questions of approximation theory}
\label{sec:intro}

\noindent
{\bf Target functions.} Let $X \subset {\mathbb R}^d$ be a compact domain, and let $f: X \rightarrow {\mathbb R}$ be a real-valued traget function we aim to approximate. Throughout this chapter, we assume $f$ is continuous, denoted $f \in C(X)$. We consider the scenario where evaluating $f$ at any point ${\bf x} \in X$ is computationally expensive, making numerous evaluations across $X$ infeasible. 
However, our computational budget permits a limited number of evaluations. 
Such scenarios are common in computational science, for instance when the target function is available through a system of parametric differential equations (ODEs/PDEs). In this case, one evaluation of the target function for a fixed set of parameters requires an expensive ODE/PDE solve.

\smallskip
\noindent
{\bf Neural network surrogates.} Given a continuous target function $f \in C(X)$, we 
aim to use our limited computational resources to perform a (small) number of evaluations of $f$ to generate a training dataset and then train a neural network $f_{\boldsymbol\theta}: {\mathbb R}^{d} \rightarrow {\mathbb R}$ with $n_{\boldsymbol\theta}$ parameters to approximate $f$. 
The goal is to replace expensive evaluations of $f({\bf x})$ with less expensive yet accurate NN evaluations $f_{\boldsymbol\theta}({\bf x})$. 
In this context, $f_{\boldsymbol\theta}$ serves as an accurate and fast-to-evaluate surrogate for $f$. 
Note that evaluating a neural network typically involves simple matrix-vector operations and is hence cost-effective.

\smallskip
\noindent
{\bf Approximation theory.} 
When the approximant is a neural network, the main task of approximation theory is to study the approximability of the
target function by neural networks. Specifically, these studies aim to
address three major questions:
\begin{itemize}
\item {\bf Density} (or convergence): if $n_{\boldsymbol\theta} \rightarrow \infty$, is there
  $f_{\boldsymbol\theta}$ that can approximate $f$ arbitrarily well?

\item {\bf Error} (or convergence rate):  if $n_{\boldsymbol\theta} < \infty$ is kept fixed, how close can $f_{\boldsymbol\theta}$ get to $f$?

\item {\bf Complexity}: how large should $n_{\boldsymbol\theta}$ be to achieve a desired accuracy constraint $|| f - f_{\boldsymbol\theta} || <
  \varepsilon$?
\end{itemize}

In this chapter, we primarily focus on the first question: the question of density. We will address error-complexity analysis in the next chapters. Density concerns the theoretical ability of $f_{\boldsymbol\theta}$ to approximate $f$ arbitrarily well. Specifically, can we approximate any target function in a given function space as accurately as we wish using neural networks? In other words, is the family of neural networks ``dense" in the space of continuous functions?

To better understand the concept of density, we will start with the simpler case where the approximants are algebraic polynomials (Section \ref{sec:density-polynomials}). We will then discuss a classical density result for shallow feedforward networks with one hidden layer (Section \ref{sec:density-shallow}), followed by more recent results on deep ReLU networks (Sections \ref{sec:relu-net}-\ref{sec:depth-power}).

We note that one important issue not addressed here is the selection and preparation of training data $\{ ({\bf x}^{(m)}, f({\bf x}^{(m)}))
\}_{i=m}^n$. Specifically, to achieve a desired accuracy $|| f - f_{\boldsymbol\theta} || < \varepsilon$, we wish to know how to optimally select the number $n$ of data points, the location
(or distribution) of input data points $\{ {\bf x}^{(m)} \}_{m=1}^n$ in
$X$, and the quality (or accuracy) of output data points $\{f({\bf x}^{(m)}) \}_{m=1}^n$. This is a complex problem, possibly beyond the scope of approximation theory alone. For simplicity, throughout this review paper, we assume the availability of abundant high-quality data. 
For further reading on approximation theory outside the context of neural networks,
refer to \cite{Trefethen_book:2019,Powell_book:1996, DeVore_Lorentz_book:1992,DeVore_Acta:98}.

\begin{remark}
 (Approximation theory vs. numerical analysis) Achieving more accurate approximations of a target function generally requires increasing the complexity of the approximants. Understanding this trade-off between accuracy and complexity is the main goal of ``constructive" approximation, focusing on how the approximation can be constructed effectively. In this regard, 
 both approximation theory and numerical analysis share similar goals. However, approximation theory tends to prioritize theoretical aspects and is less concerned with computational issues. In numerical computation, target functions are often implicitly defined through differential, integral, or integro-differential equations, presenting additional constraints and challenges not typically addressed in approximation theory. 
\end{remark}

\section{Density in the space of continuous functions}
\label{sec:density-polynomials}

Here, we briefly review the concept of uniform convergence, essential for exploring density in the space of continuous functions. We then present some results on polynomial density, serving as an example to better grasp the concept of density.

\subsection{Uniform convergence}

For studying density, e.g., in the space of continuous functions, we need to consider a notion of ``distance'' that measures the ``closeness'' of the target function to the approximant. Here, we will consider uniform norms and hence review the notion of uniform convergence. 

A sequence $\{ f_m \}_{m=1}^{\infty}$ of functions is said to converge uniformly to a limiting
function $f$ on a set $X$ if given any $\varepsilon>0$, there exists a
natural number $N$ such that $| f({\bf x}) - f_m({\bf x}) | < \varepsilon$ for all
$m \ge N$ and for all ${\bf x} \in X$. We write
\begin{equation}\label{uniform_conv1}
\hskip 1.5cm f_m \rightarrow f \ \ \ \ \text{uniformly}.
\end{equation}

An equivalent formulation for uniform convergence can be given in
terms of the supremum norm (also called infinity norm or uniform norm). Let the supremum norm of $f$ be
$$
|| f ||_{\infty} := \sup_{{\bf x} \in X} | f({\bf x}) |.
$$
Then the uniform convergence \eqref{uniform_conv1} is equivalent to
\begin{equation}\label{uniform_conv2}
|| f - f_m ||_{\infty} \xrightarrow{m \rightarrow \infty} 0.
\end{equation}

We note that uniform convergence is a stronger form of convergence than pointwise convergence. In pointwise convergence, for each point ${\bf x} \in X$, there exists an integer $N = N(\varepsilon, {\bf
  x})$ such that for all $m \ge N$,  $| f({\bf x}) - f_m({\bf x}) | < \varepsilon$, with the rate of convergence potentially varying across the domain. 
  Uniform convergence, on the other hand, requires a single integer $N =
N(\varepsilon)$, independent of the point ${\bf x}$, such that the same condition holds for all ${\bf x} \in X$. 
This ensures that the rate of convergence of $f_m({\bf x})$ to $f({\bf x})$ is consistent throughout the domain $X$, regardless of the location of ${\bf x}$. While uniform convergence implies pointwise convergence. 
For example, consider $f_m(x) =
x^m$ on $X=[0,1]$. This sequence converges pointwise to a limit function $f(x)$ defined by $f(x) = 0$ for $x \in [0,1)$ and $f(x) = 1$ for $x =1$, but it does not converge uniformly. 
At $x = (3/4)^{1/m} <1$, the sequence $f_m(x)$ can remain far from the limit function, showing that the convergence is not uniform. Furthermore, the limit function $f(x)$ is discontinuous, which aligns with the Uniform Limit Theorem: if a sequence of continuous functions converges uniformly, the limit must also be continuous. This example illustrates that pointwise convergence does not necessarily imply uniform convergence.

\subsection{Density of polynomials}

In 1885, Karl Weierstrass proved that algebraic polynomials are dense in the space of continuous functions on closed intervals. In 1937, Marshall Stone generalized this to higher dimensions.

Let $C([a,b])$ denote the space of real-valued continuous functions on
the closed interval $[a,b]$. The Weierstrass approximation
theorem states that the set of real-valued algebraic polynomials on
$[a,b]$ is dense in $C([a,b])$ with respect to the supremum norm, i.e. in the topology of uniform convergence on compact sets. 
That is, given any function $f \in C([a,b])$, there exists a sequence
of algebraic polynomials $\{ p_m \}_{m=1}^{\infty}$ such that $p_m
\rightarrow f$ uniformly. 
We say that $f \in C([a,b])$ can be uniformly approximated as
accurately as desired by an algebraic polynomial. 
The precise statement of the theorem follows.

\smallskip
\noindent
{\bf Weierstrass Theorem.} Given a function $f \in C([a,b])$ and an
arbitrary $\varepsilon >0$, there exists an algebraic polynomial $p$
such that $| f(x) - p(x) | < \varepsilon$ for all $x \in [a,b]$, or
equivalently $|| f - p ||_{\infty} < \varepsilon$.

\smallskip

A constructive proof of the theorem can be established using Bernstein polynomials; see e.g. \cite{Davidson_Donsig_book}. 
We also note that a similar result holds for $2 \, \pi$-periodic continuous functions and trigonometric polynomials, also due to Karl Weierstrass. Specifically, trigonometric polynomials are uniformly dense in the class of $2 \, \pi$-periodic continuous functions. 

Stone generalized and proved the Weierstrass approximation theorem by replacing the closed interval $[a,b]$ with any compact Hausdorff
space $X$, i.e. a compact space where any two distinct points have disjoint neighborhoods. Note that all metric spaces, including the Euclidean space ${\mathbb R}^d$, are Hausdorff. 

It is worth noting that polynomials of finite degree $m < \infty$ are not dense in the space of continuous functions, recalling that the space $P_m = \text{span} \{ 1, x, \dotsc, x^m \}$ of algebraic polynomials and the space $T_m = \text{span} \{ 1, \sin x, \cos x \dotsc, \sin mx, \cos mx \}$ of trigonometric polynomials of degree at most $m < \infty$ are finite-dimensional spaces.

\section{Density of shallow networks}
\label{sec:density-shallow}

In this section, we discuss the main density result for feedforward networks with one hidden layer. 

\subsection{One-hidden-layer feedforward networks}
\label{sec:two-layer-network}

We will consider the family of feedforward networks, discussed in Section \ref{sec:NN}, with $d$ input neurons, one hidden layer ($L=1$) containing $n_0=W$ neurons, all using the same activation function $\sigma: {\mathbb R} \rightarrow
{\mathbb R}$, and one output neuron ($n_L=1$) without activation and without bias. 
Letting ${\bf x} = (x_1, \dotsc, x_d) \in {\mathbb R}^d$ be the input variable, the network's output reads
$$
y = \sum_{i=1}^W M_{1,i}^1 \, \sigma ( \sum_{j=1}^d M_{i,j}^0 x_j  + b_i),
$$
where $M^0 \in {\mathbb R}^{W \times d}$ and $M^1 \in {\mathbb R}^{1 \times W}$ are the weight matrices, and ${\bf b} = (b_1, \dotsc, b_W) \in {\mathbb R}^W$ is the bias vector. 
Denoting the $i$-th row
of $M^0$ by ${\bf w}^{(i)} := (M_{i,1}^{1}, \dotsc, M_{i,d}^1) \in
{\mathbb R}^d$ and the $i$-th
column of $M^1$ by $c_i := M_{1,i}^2 \in {\mathbb R}$, we write the network's output
more succinctly as
$$
y = \sum_{i=1}^W c_i \, \sigma (  {\bf w}^{(i)} \cdot {\bf x} + b_i),
$$
where $ {\bf a} \cdot {\bf b} := \sum_{j=1}^d a_j \, b_j$ is the inner
product of ${\bf a} \in {\mathbb R}^d$ and ${\bf b} \in {\mathbb R}^d$.

\subsection{Pinkus theorem}

We address the density question by considering
the network space (i.e. the case $W \rightarrow \infty$),
$$
{\mathcal M}(\sigma) := \text{span} \{ \sigma (  {\bf w} \cdot
{\bf x} + b) \, : \, {\bf w} \in {\mathbb R}^d, \, b \in {\mathbb R} \},
$$
and ask for which class of activation functions the network space
${\mathcal M}(\sigma)$ is dense in the space $C({\mathbb R}^d)$ of
continuous functions with respect to the supremum norm in the
topology of uniform convergence on compact sets. Equivalently, given a target function $f \in
C({\mathbb R}^d)$ and a compact subset $X \subset {\mathbb R}^d$ and an
arbitrary $\varepsilon >0$, we ask for what $\sigma$ there exists $g \in {\mathcal M}(\sigma)$ such that $\sup_{{\bf x} \in X} | f({\bf x}) - g({\bf x}) | < \varepsilon$. 
Formally, we state the main density result (Theorem 3.1 in \cite{Pinkus:99}).

\medskip
\noindent
{\bf Pinkus Theorem.} Let $\sigma \in C({\mathbb R})$. Then ${\mathcal
  M}(\sigma)$ is dense in $C({\mathbb R}^d)$ with respect to the
supremum norm on compact sets if and only if $\sigma$ is not a polynomial.

\medskip
As an intuitive example, let $d=1$ and $X=[a,b] \subset {\mathbb R}$, and consider $\sigma (x) = \cos x$, which satisfies the conditions of the above theorem; it is continuous and not a polynomial. We obtain
$$
g(x) = \sum_{i \ge 1} c_i \, \cos (w_i \, x + b_i),
$$
which is the amplitude-phase form of Fourier series from which we can recover the more familiar sine-cosine form by the identity $\cos(\alpha + \beta) =
\cos \alpha \, \cos \beta - \sin \alpha \, \sin \beta$. Recall that if we have a
continuous function, we can expand it in Fourier series, and hence it follows that $g$ is dense in the space of continuous functions.

\subsection{Proof sketch of Pinkus theorem}
\label{sec:proof-Pinkus}

First, we need to show that if ${\mathcal M}(\sigma)$ is dense, then
$\sigma$ is not a polynomial. Suppose $\sigma \in P_m ({\mathbb R})$
is a polynomial of degree $m$, then for every choice of ${\bf w} \in {\mathbb R}^d$ and
$b \in {\mathbb R}$, $\sigma (  {\bf w} \cdot {\bf x} + b)$ is a
multivariate polynomial of total degree at most $m$, and thus
${\mathcal M}(\sigma)$ is the space of all polynomials of total degree
at most $m$, that is ${\mathcal M}(\sigma) = P_m ({\mathbb R}^d)$, which does
not span $C({\mathbb R}^d)$, contradicting the density. 
We next show the converse result: if $\sigma$ is not a polynomial,
then ${\mathcal M}(\sigma)$ is dense. This is done in four steps:

\begin{itemize}

\item Step 1. Consider the 1D case (i.e. $d=1$) and the
  1D counterpart of ${\mathcal M}(\sigma)$:
$$
{\mathcal N}(\sigma) = \text{span} \{ \sigma (w \, x + b), \, w, b \in
{\mathbb R} \}.
$$

\item Step 2. Show that for $\sigma \in C^{\infty}({\mathbb R})$ and
  not a polynomial, ${\mathcal N}(\sigma)$ is dense in $C({\mathbb R})$.

\item Step 3. Weaken the smoothness demand on $\sigma$, using
  convolution by mollifiers, and show that for $\sigma \in C({\mathbb R})$ and
  not a polynomial, ${\mathcal N}(\sigma)$ is dense in $C(\mathbb R)$.

\item Step 4. Extend the result to multiple dimensions: show that if ${\mathcal N}(\sigma)$ is dense in
  $C({\mathbb R})$ then ${\mathcal M}(\sigma)$ is dense in $C({\mathbb R}^d)$.
 
\end{itemize}

\medskip
\noindent
{\it Proof of step 2.} 
We will use a very interesting lemma asserting that if a smooth
function (i.e. $C^{\infty}$) on an interval is such that its Taylor
expansion about every point of the interval has at least one
coefficient equal to zero, then the function is a polynomial. 

\vskip 0.1cm
\noindent
{\bf Lemma.} 
Let $\sigma \in C^{\infty}((\alpha,\beta))$, where $(\alpha,\beta) \subset {\mathbb R}$ is an
open interval on the real line. If for every point $x \in (\alpha,\beta)$ on
the interval there exists an integer $k=k(x)$ such that the $k$-th
derivative of $\sigma$ vanishes at $x$, i.e. $\sigma^{(k)}(x) = 0$, then $\sigma$ is a polynomial. 

\vskip .1cm
\noindent
For a proof of this lemma see page 53 in \cite{Donoghue:1969}.

\vskip .1cm
\noindent
The above lemma implies that since $\sigma$ is smooth and not a
polynomial, there exists a point $b_0 \in {\mathbb R}$ at which
$\sigma^{(k)} \neq 0$, $k=0, 1, 2, \dotsc$.  
Now we note that
$$
\frac{\sigma((w + h) \, x + b_0) - \sigma(w \, x + b_0)}{h} \in
{\mathcal N}(\sigma), \qquad \forall \, h \neq 0,
$$
where $w \in {\mathbb R}$. Taking the limit (as $h \rightarrow 0$), it
follows that the derivative of $\sigma$ with respect to $w$ is in $\overline{{\mathcal N}(\sigma)}$, which
is the closure of ${\mathcal
  N}(\sigma)$. In particular, for $w=0$ we get
$$
\lim_{h \rightarrow 0} \frac{\sigma((w + h) \, x + b_0) - \sigma(w \, x + b_0)}{h} \Big|_{w=0} = \frac{d}{d w} \sigma(w \, x + b_0) \Big|_{w=0} = x \, \sigma'(b_0) \in
\overline{{\mathcal N}(\sigma)},
$$
Similarly (by considering $k+1$ terms of $\sigma$ and taking the limit
$h \rightarrow 0$) we can show that
$$
\frac{d^k}{d w^k} \sigma(w \, x + b_0) \Big|_{w=0} = x^k \, \sigma^{(k)}(b_0) \in
\overline{{\mathcal N}(\sigma)}.
$$ 
Since $\sigma^{(k)} \neq 0$, $k=0, 1, 2, \dotsc$, the set
$\overline{{\mathcal N}(\sigma)}$ contains all monomials. By the
Weierstrass Theorem this implies that $\overline{{\mathcal
    N}(\sigma)}$ and hence ${\mathcal N}(\sigma)$ is dense in
$C({\mathbb R})$, because if the closure of a function space is dense, the function
space will be dense too: we can approximate any function in the
closure space by functions in the space as accurately as we wish. 

\medskip
\medskip
\noindent
{\it Proof of step 3.} The proof utilizes the classical technique of
convolution by mollifiers to weaken the smoothness requirement of
$\sigma$. In this technique we consider a mollified activation
function $\sigma_{\phi} (x)$ obtained by convolving $\sigma \in
C({\mathbb R})$ with a smooth and compactly supported mollifier $\phi \in
C_0^{\infty}({\mathbb R})$, 
$$
\sigma_{\phi} (x) = \int_{\mathbb R} \sigma (x-y) \, \phi(y) \, dy.
$$
Since both $\sigma$ and $\phi$ are continuous and $\phi$ has compact
support, the above integral exists for all $x \in {\mathbb
R}$. We also have $\sigma_{\phi} \in C^{\infty}({\mathbb
R})$; this can be simply shown using the integration by parts. Moreover, taking the limit of Riemann sums, one can easily show that
$\sigma_{\phi}$ and ${\mathcal
    N}(\sigma_{\phi})$ are contained in $\overline{{\mathcal
    N}(\sigma)}$: 
$$
\sigma_{\phi}(wx+b) = \int_{\mathbb R} \sigma (wx+b-y) \, \phi(y) \, dy = \lim_{N \rightarrow \infty} \sum_{i=1}^N \sigma(wx+b - y_i)  \, \phi(y_i) \, \Delta y_i \in \overline{{\mathcal
    N}(\sigma)},
$$
where the support of $\phi$, say the closed interval $[a,b]$, is decomposed into $N$ subintervals of length $\Delta y_i$, with $i=1, \dotsc, N$, by inserting $N$ points such that $a=y_1 < y_2 < \dotsc < y_N < b$. 
Now since $\sigma_{\phi} \in C^{\infty}({\mathbb
R})$, and provided we choose $\phi$ such that $\sigma_{\phi}$ is not a
polynomial (also note that $\sigma$ is not a polynomial), then by the
method of proof of Step 2, all monomials are contained in
$\overline{{\mathcal N}(\sigma_{\phi})}$. Hence ${\mathcal
    N}(\sigma_{\phi})$ and therefore $\overline{{\mathcal
    N}(\sigma)}$ and ${\mathcal
    N}(\sigma)$ are dense in $C({\mathbb R})$.

\begin{figure}[!h]
\centering
    \begin{tabular}{c c c}
\vspace{-.4cm}
        \includegraphics[width=0.32\linewidth]{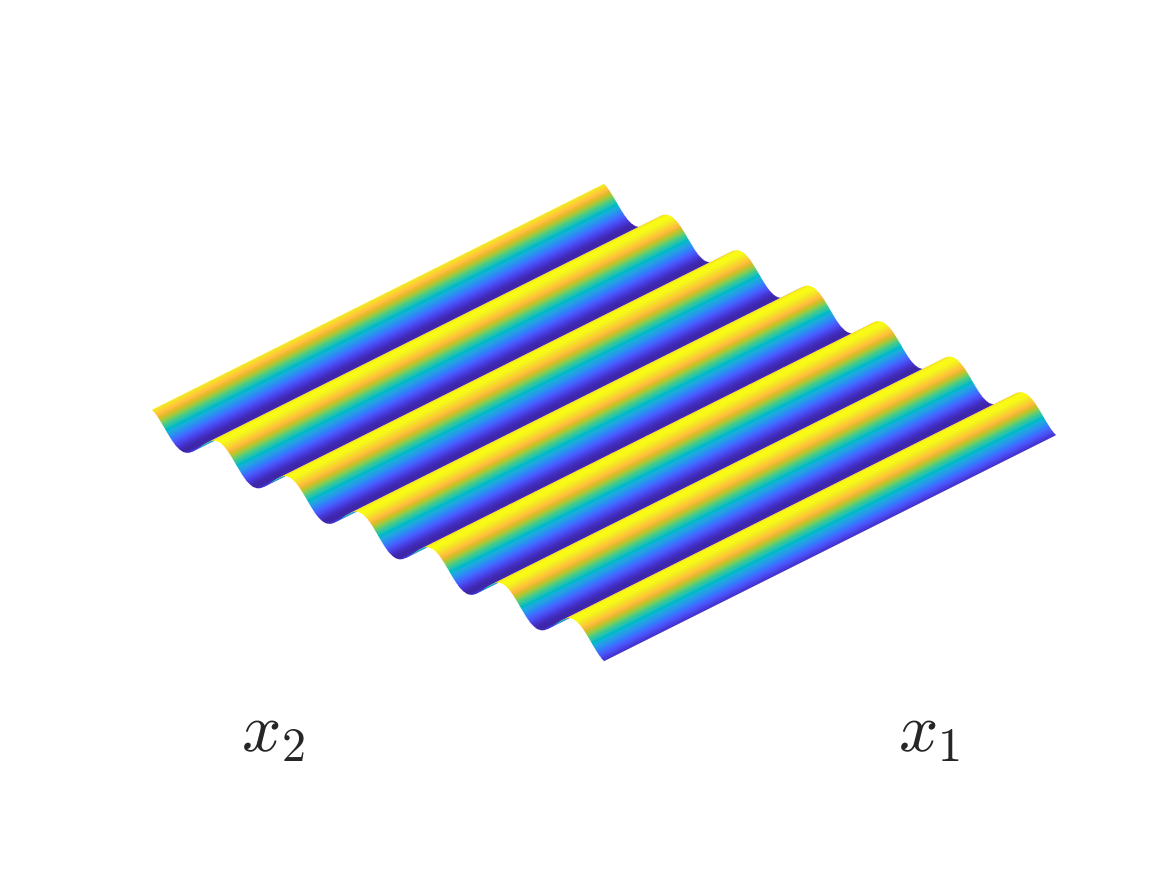}
                                   & \includegraphics[width=0.32\linewidth]{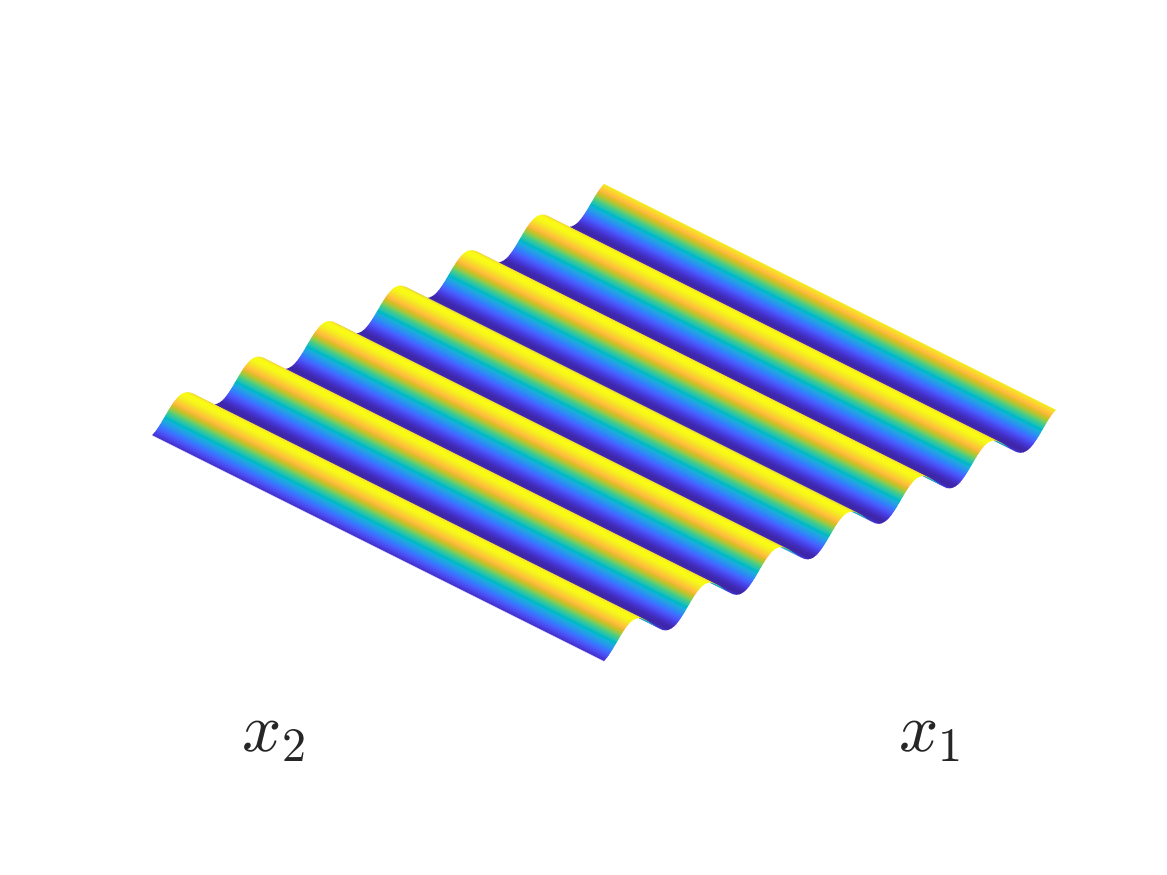}
      & \includegraphics[width=0.32\linewidth]{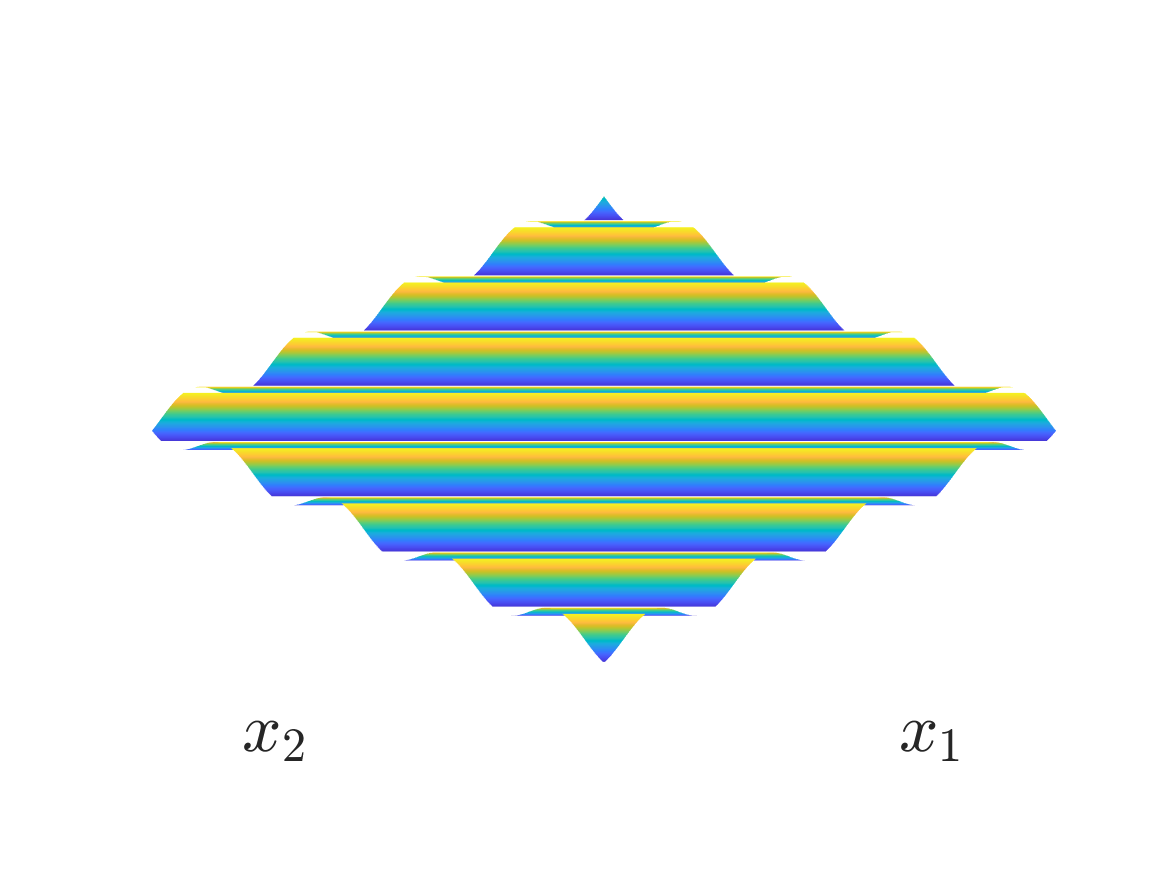}\\
$\sin (0 . x_1 + 1 . x_2)$ & $\sin (1 . x_1 + 0 . x_2)$ & $\sin
                                                          (\frac{1}{\sqrt{2}}
                                                          . x_1 +
                                                          \frac{1}{\sqrt{2}}
                                                          . x_2)$ 
    \end{tabular}
\caption{An example of a ridge function $g({\bf w} \cdot {\bf
  x}) = \sin {\bf w} \cdot {\bf x}$ in two dimensions, where ${\bf x} = (x_1, x_2) \in
{\mathbb R}^2$, and for three different direction vectors ${\bf w} = (0,1)$ (left), ${\bf w}
= (1,0)$ (middle), and ${\bf w} = (1/\sqrt{2},1/\sqrt{2})$ (right).}
\label{Ridge_fig}
\end{figure}

\medskip
\medskip
\noindent
{\it Proof of step 4.} One interesting technique to ``reduce
dimension'' (here we want to reduce $d$ to 1) is to utilize {\it ridge
functions}, also known as plane waveforms in the context of hyperbolic
PDEs. A ridge function $g: {\mathbb R} \rightarrow {\mathbb R}$ is a
multivariate function $f: {\mathbb R}^d \rightarrow {\mathbb R}$ of the form
$$
f({\bf x}) = g({\bf w} \cdot {\bf x}), \quad {\bf w} \in {\mathbb R}^d
\setminus \{ {\bf 0} \}.
$$
Figure \ref{Ridge_fig} displays an example of a ridge function $g({\bf w} \cdot {\bf
  x}) = \sin {\bf w} \cdot {\bf x}$, where ${\bf x} = (x_1, x_2) \in
{\mathbb R}^2$, with three choices ${\bf w} = (0,1)$ (left), ${\bf w}
= (1,0)$ (middle), and ${\bf w} = (1/\sqrt{2},1/\sqrt{2})$ (right). 

\noindent
As we observe, the vector ${\bf w} \in {\mathbb R}^d \setminus \{ {\bf
  0} \}$ determines the direction of the plane wave. Importantly, for each ${\bf w} \in
{\mathbb R}^d \setminus \{ {\bf 0} \}$ and $b \in {\mathbb R}$, the
function $\sigma ({\bf w} \cdot {\bf x} + b)$ that is the building
block of ${\mathcal M}(\sigma)$ is also a ridge
functions. Now let
$$
{\mathcal R} := \text{span} \{ g({\bf w} \cdot {\bf x}): \, {\bf w}
\in {\mathbb R}^d, \, || {\bf w} ||_2 = 1, \, g \in C({\mathbb R})
\}. 
$$
We first note that ${\mathcal R}$ is dense in $C({\mathbb R}^d)$,
because it contains all functions of the form $\cos({\bf w} \cdot {\bf
  x})$ and  $\sin({\bf w} \cdot {\bf
  x})$ which are dense on any compact subset of $C({\mathbb
  R}^d)$. Note that this does not directly imply that ${\mathcal
  M}(\sigma)$ will also be dense, but if ridge functions in ${\mathcal
  R}$ were not dense in $C({\mathbb R}^d)$, then it would not be
possible for ${\mathcal M}(\sigma)$ to be dense in $C({\mathbb
  R}^d)$. Now let $f \in C(X)$ be a given target function on some compact set $X \subset
{\mathbb R}^d$. Since ${\mathcal R}$ is dense in $C(X)$, then given
$\varepsilon >0$ there exist $\{ g_i \}_{i=1}^N \in
C({\mathbb R})$ and ${\bf w}^{(i)} \in {\mathbb R}^d$ with $|| {\bf
  w}^{(i)} ||_2 = 1$, $i=1, \dotsc, N$ (for some $N$) such that
$$
| f({\bf x}) - \sum_{i=1}^N g_i ({\bf w}^{(i)} \cdot {\bf x}) | <
\frac{\varepsilon}{2}, \qquad \forall \, {\bf x} \in X.
$$
Since $X$ is compact, then for each $i=1, \dotsc, N$, we have $\{ {\bf w}^{(i)} \cdot {\bf x}: \, {\bf x} \in
X \} \subseteq [a_i, b_i]$ for some bounded interval $[a_i, b_i]$. We
next utilize the fact that ${\mathcal N}(\sigma)$ is dense in $C([a_i,
b_i])$, for all $i=1, \dotsc, N$, which implies that there exist
$c_{i,j}, \, w_{i, j}, \, b_{i, j} \in {\mathbb R}$, with $j=1,
\dotsc, n_i$ for some $n_i$, such that
$$
| g_i (t) - \sum_{j=1}^{n_i} c_{i,j} \, \sigma( w_{i,j} \, t +
b_{i,j}) | < \frac{\varepsilon}{2 \, N}, \qquad \forall t \in [a_i,
b_i], \quad i=1, \dotsc, N.
$$
Hence, combining the above two inequalities, we get
\begin{align*}
&| f({\bf x}) - \sum_{i=1}^N \sum_{j=1}^{n_i} c_{i,j} \, \sigma( w_{i,j} \, {\bf w}^{(i)} \cdot {\bf x} +
b_{i,j}) | = \\ 
&| f({\bf x}) - \sum_{i=1}^N g_i ({\bf w}^{(i)} \cdot
{\bf x}) + \sum_{i=1}^N g_i ({\bf w}^{(i)} \cdot {\bf x})  - \sum_{i=1}^N \sum_{j=1}^{n_i} c_{i,j} \, \sigma( w_{i,j} \, {\bf w}^{(i)} \cdot {\bf x} +
b_{i,j}) | \le \\
&| f({\bf x}) - \sum_{i=1}^N g_i ({\bf w}^{(i)} \cdot
{\bf x}) | + \sum_{i=1}^N |g_i ({\bf w}^{(i)} \cdot {\bf x})  - \sum_{j=1}^{n_i} c_{i,j} \, \sigma( w_{i,j} \, {\bf w}^{(i)} \cdot {\bf x} +
b_{i,j}) | < \\
&\frac{\varepsilon}{2} + N \, \frac{\varepsilon}{2 \, N} =
  \varepsilon, \qquad \forall \, {\bf x} \in X.
\end{align*}
Clearly, this implies that there exist $c_{k}, \, b_{k} \in
{\mathbb R}$ and ${\bf w}^{(k)} \in {\mathbb R}^d$, with $k = 1,
\dotsc, W$ for some $W = \sum_{i=1}^{N} n_i$, such that  
$$
| f({\bf x}) - \sum_{k=1}^W c_{k} \, \sigma({\bf w}^{(k)} \cdot {\bf x} +
b_{k}) | < \varepsilon, \qquad \forall \, {\bf x} \in X.
$$
This in turn means that there exist $f_{\boldsymbol\theta} \in
{\mathcal M}(\sigma)$, given by $f_{\boldsymbol\theta} ({\bf x}) =
\sum_{k=1}^W c_{k} \, \sigma({\bf w}^{(k)} \cdot {\bf x} + b_{k})$ for some $W$, such that 
$$
| f({\bf x}) - f_{\boldsymbol\theta} ({\bf x}) | < \varepsilon, \qquad \forall \, {\bf x} \in X.
$$
Hence, ${\mathcal M}(\sigma)$ is dense in $C(X)$. This complete the proof of Pinkus theorem.
\qed

\section{Deep ReLU networks with uniform width}
\label{sec:relu-net}

For studying the approximation power of deep networks, we define two specific classes of feed-forward networks: (I) uniform-width standard ReLU networks and (II) uniform-width special networks. 
All approximation results in the remainder of this chapter pertain to uniform-width standard ReLU networks. Uniform-width special ReLU networks will be primarily used to prove these results.

\medskip
\noindent
{\bf I. Uniform-width standard ReLU networks}. A standard ReLU network on $X \subset {\mathbb R}^d$, with $d$ input neurons, $n_L$ (often 1) output neurons, and $L \ge 1$ hidden layers, each
with $W \ge 1$ neurons, is given by a sequence of matrix-vector tuples (or weight-bias tuples),
$$
\Phi := \{ (M_0,b_0), (M_1,b_1),  \dotsc, (M_L,b_L) \} \in {\mathbb R}^{(L-1) W^2 + (d+n_L)W} \times {\mathbb R}^{LW + n_L}, 
$$ 
where $M_0 \in {\mathbb R}^{W \times d}$, $M_1, \dotsc,
M_{L-1} \in {\mathbb R}^{W \times W}$, $M_L \in {\mathbb R}^{n_L \times W}$, $b_0, \dotsc, b_{L-1} \in {\mathbb R}^{W}$, and $b_L \in {\mathbb R}^{n_L}$. 
We denote by $f_{\Phi}: {\mathbb R}^d \rightarrow {\mathbb R}^{n_L}$ the real-valued
function that the network $\Phi$ realizes, 
$$
f_{\Phi}({\bf x}) = A_L \circ \sigma \circ A_{L-1} \circ \dotsc \circ
\sigma \circ A_0 ({\bf x}), \qquad {\bf x} \in X, 
$$
with affine maps formed by a linear transformation followed by a translation,
$$
A_0 ({\bf x})= M_{0} \, {\bf x} + b_0, \qquad 
A_{\ell} ({\bf z})= M_{\ell} \, {\bf z} + b_{\ell}, \qquad {\bf z} \in {\mathbb R}^W, \qquad \ell = 1, \dotsc, L,
$$
and with ReLU activation $\sigma(x) = \max \{0,x\}$, where $x \in {\mathbb R}$. 
We denote by ${\mathcal N}_{W,L}$ the set of all functions generated by uniform-width standard ReLU networks,
$$
{\mathcal N}_{W,L} = \{ f_{\Phi}: X \subset {\mathbb R}^d\rightarrow {\mathbb R}^{n_L}, \ \ \Phi = \{ (M_{\ell} , b_{\ell}) \}_{\ell
  = 0}^{L}  \in {\mathbb R}^{(L-1) W^2 + (d+n_L)W} \times {\mathbb R}^{LW + n_L} \}.
$$

\medskip
\noindent
{\bf An example: identity ReLU network.} Let $I_d \in {\mathbb R}^{d
\times d}$ be the identity matrix, i.e. $I_d \, {\bf x} = {\bf x}$. Consider the standard ReLU
network,
$$
\Phi_I := \{ (M_0, b_0), (M_1, b_1) \},
$$ 
with one hidden layer ($L=1$) and $W= 2 \, d$ and $n_L = d$, where
$$
M_0 = \left( \begin{array}{c}
I_d \\
- I_d
\end{array} \right) \in {\mathbb R}^{2d \times d}, \qquad 
b_0 = {\bf 0} \in {\mathbb R}^{2 d}, \qquad 
M_1 = ( I_d \ | \ - I_d ) \in {\mathbb R}^{d \times 2 d}, \qquad 
b_1 = {\bf 0} \in {\mathbb R}^d.
$$
Then, using the identity relation $\sigma(x) - \sigma(-x) = x$, it can
easily be shown that
$$
f_{\Phi_I} ({\bf x}) = I_d \, {\bf x} = {\bf x}.
$$
We call $\Phi_I$ the identity ReLU network. As we will see later, the identity ReLU network is handy in our constructive proofs when for instance we need to recover negative values (for example when ${\bf x}$ contains negative values) after going through ReLU activation.

\begin{figure}[!h]
\centering
\includegraphics[width=9cm,height=3.5cm]{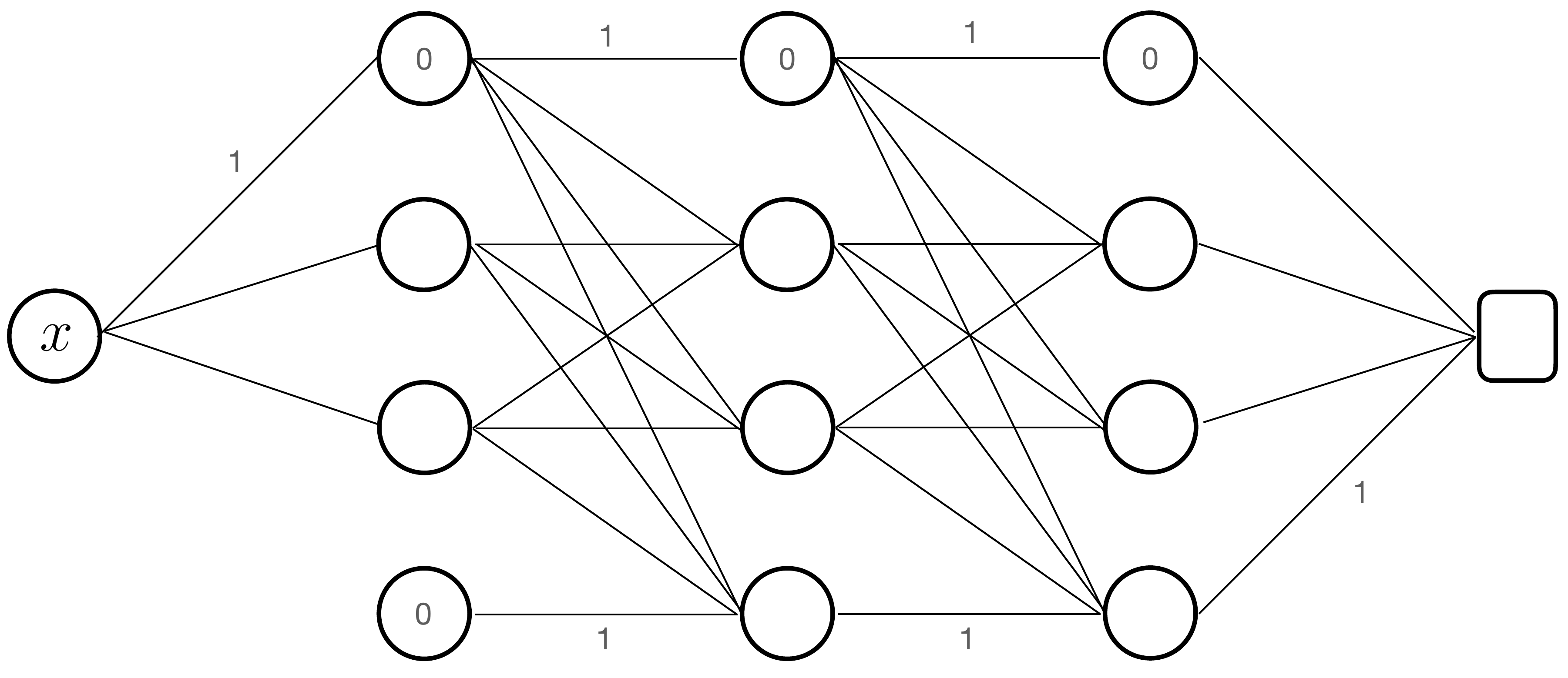}
\vspace{-.2cm}
\caption{Graph representation of a special ReLU network in $\hat{\mathcal N}_{4,3}$; network weights are indicated adjacent to their corresponding edge and biases internal to their corresponding neuron.}
\label{fig:special_relu}
\end{figure}

\medskip
\noindent
{\bf II. Uniform-width special ReLU networks.} 
A special ReLU network on $X \subset {\mathbb R}_+^d$ can be considered as a special ``subset'' of standard ReLU networks with the same depth, comparable width, and where special roles are reserved for the top and bottom neurons of each hidden layer. Specifically, we define three types of channels that appear in special ReLU networks. 
\begin{enumerate}

\item A source channel formed by the top $d$ neurons in each hidden layer that are assumed to be ReLU-free with unit weights and zero bias. The neurons in a source channel do not take any input from neurons in other channels and do not do any computation. This channel simply carries forward the input ${\bf x}$ and may feed ${\bf x}$ into the neurons in the subsequent hidden layer. 

\item A collation channel formed by the bottom neuron in each hidden layer that are also assumed to be ReLU-free. This channel is used to collect intermediate computations (i.e. outputs of hidden layers). The neurons in a collation channel do not feed into subsequent calculations. They only take outputs of neurons in previous layers and carry them over with unit weight to subsequent bottom neurons. 

\item Standard computational channels with ReLU activation.
\end{enumerate}
Figure \ref{fig:special_relu} displays the graph representation of a
special ReLU network when $d=1$, $W=4$, $L=3$. 

Let $\hat{\Phi} =\{ (\hat{M}_{\ell}, \hat{b}_{\ell}) \}_{\ell
  = 0}^{L}$ represent the set of matrix-vector tuples of a special ReLU network, and let $f_{\hat{\Phi}}: {\mathbb R}^d \rightarrow {\mathbb R}^{n_L}$ denote the function it realizes. We define $\hat{\mathcal N}_{W,L}$ as the set of functions generated by special
ReLU networks:
$$
\hat{\mathcal N}_{W,L} = \{ f_{\hat{\Phi}}: X \rightarrow {\mathbb R}^{n_L}, \ \
\hat{\Phi} =\{ (\hat{M}_{\ell}, \hat{b}_{\ell}) \}_{\ell
  = 0}^{L} \in {\mathbb R}^{(L-1) W^2 + (d+n_L)W} \times {\mathbb R}^{LW + n_L} \}.
$$

Note that since top and bottom neurons are ReLU-free,
special networks are not a direct subset of ReLU
networks. However, we make the following observations:
\begin{itemize}
\item When the input is non-negative (i.e. $X \subset {\mathbb R}_+^d$), then $x=\sigma
  (x)$, and hence the assumption that top neurons are ReLU-free is not restrictive. 

\item The first bottom neuron in the first hidden layer (corresponding to $\ell = 0$) is simply taking
  zero and since $0 = \sigma (0)$, the ReLU-free assumption is not restrictive. 

\item Any other bottom neuron in the remaining $L-1$ hidden layers (for $\ell = 1,
  \dotsc, L-1$) takes an input function, say $g_{\ell} (x)$ that
  depends continuously on $x$. Hence there is a constant, say
  $C_{\ell}$, such that $g_{\ell} (x) + C_{\ell} \ge 0$ for every $x
  \in [0,1]$, e.g., we may set $C_{\ell} := - \min_{x} g_{\ell}(x)$. This implies that $g_{\ell} (x) = \sigma(g_{\ell} (x) +
  C_{\ell} )- C_{\ell} $.   

\end{itemize}
Consequently, with $X \subset {\mathbb R}_+^d$, given any function $f_{\hat{\Phi}} \in \hat{\mathcal
  N}_{W,L}$ corresponding to a special network $\hat{\Phi}=\{ \hat{M}^{(\ell)} , \hat{b}^{(\ell)} \}_{\ell
  = 0}^{L}$, one can construct a standard ReLU network $\Phi$ with the same complexity
that produces the same function $f_{\Phi} \equiv f_{\hat{\Phi}}$. The
parameters $\Phi = \{ {M}^{(\ell)} , {b}^{(\ell)} \}_{\ell
  = 0}^{L}$ of such standard ReLU network are given in terms of the parameters
of the special network $\hat{\Phi}$ by
$$
{M}^{(\ell)} = \hat{M}^{(\ell)}, \qquad \ell=0, \dotsc, L,
$$
$$
b_j^{(\ell)} = \hat{b}_j^{(\ell)}, \quad j=1, \dotsc, W-1, \quad
  b_W^{(\ell)} = \hat{b}_W^{(\ell)}+C_{\ell}, \quad \ell=1, \dotsc,
  L-1, \quad 
b^{L} = \hat{b}^{L} - \sum_{\ell=1}^{L-1} C_{\ell}.
$$
Hence, although special networks are not direct subsets of standard networks, for non-negative inputs the inclusion $\hat{\mathcal N}_{W,L} \subset {\mathcal N}_{W,L}$ holds in terms of sets of functions the two networks produce.

For inputs that can take negative values, when $X \subset {\mathbb R}^d$, a special network in $\hat{\mathcal N}_{W,L}$ can be converted into a corresponding standard network by using an identity ReLU network. This requires two sets of top channels in the standard network: one for ${\bf x} \in {\mathbb R}^d$ and one for $-{\bf x} \in {\mathbb R}^d$, to represent the single set of source channels in the special network. The resulting standard network will belong to ${\mathcal N}_{W+d, L}$, maintaining the same number of hidden layers but with $d$ additional neurons per layer to handle $-{\bf x}$. 
Despite these additions, the complexity of the standard network remains comparable to that of the original special network.

As we will see later, a key use of special networks is in concatenating standard networks. This approach is employed in constructive proofs (i.e., proofs by construction) of mathematical properties of networks, such as density and complexity. The following proposition exemplifies the use of special networks in constructing the sum of outputs from multiple standard networks. 
\begin{proposition}\label{prop:sum}
For any $f_{{\Phi}_j} \in {\mathcal N}_{W, L_j}$ with $j=1 ,
\dotsc, J$, the following holds:
$$
f_{{\Phi}_1} + \dotsc + f_{{\Phi}_J} \in \hat{\mathcal N}_{W+2,L_1 + \dotsc + L_J}.
$$
\end{proposition}
\begin{proof} We concatenate the standard ReLU networks $\Phi_1$ and $\Phi_2$ by adding a source and a
collation channel, as shown in Figure \ref{fig:concat_sum}, to construct a special ReLU network, say
$\Phi_{1+2}$, with width $W+2$ and depth $L_1 + L_2$, that outputs $f_{\Phi_1} + f_{\Phi_2}$. 
\begin{figure}[!h]
\centering
        \includegraphics[width=0.65\linewidth]{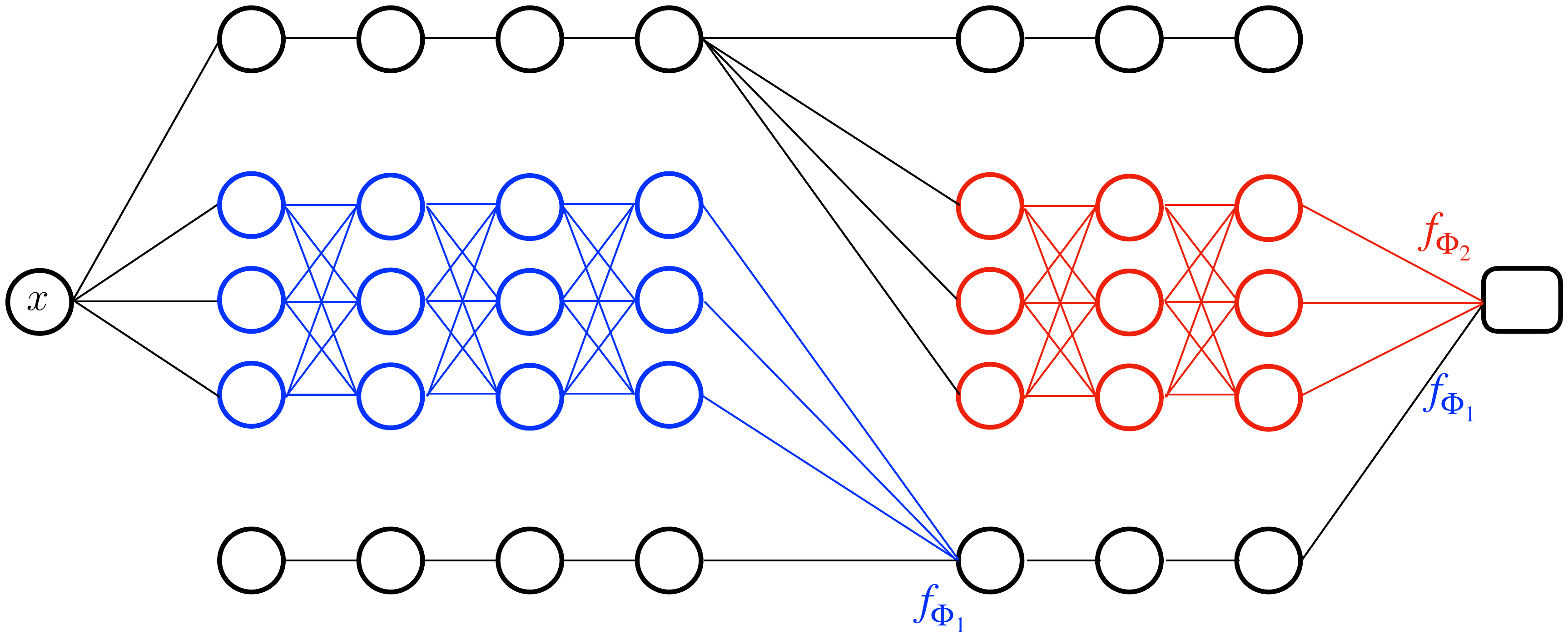}
        \vspace{-0.2cm}
\caption{By adding source and collation channels we can concatenate two
  standard ReLU networks $\Phi_1$ (in blue) and $\Phi_2$ (in red) and
  generate a special ReLU network that outputs $f_{\Phi_1} + f_{\Phi_2}$.}
\label{fig:concat_sum}
\end{figure} 
This procedure can be repeated $J-1$ times to
build a special network with
width $W+2$ and depth $L_1 + \dotsc + L_J$ that outputs $f_{\Phi_1} +
\dotsc + f_{\Phi_J}$. This completes the proof.
\end{proof}

It is worth noting that an alternative approach to concatenation is to stack standard networks ``vertically'' rather than ``horizontally'', absorbing the complexity of the concatenated network into the network's width while keeping the depth fixed. This can be done without the use of special networks, as summarized in the following proposition.
\begin{proposition}\label{prop:sum_width}
For any $f_{{\Phi}_j} \in {\mathcal N}_{W_j, L}$ and $a_j \in {\mathbb R}$, with $j=1 ,
\dotsc, J$, the following holds:
$$
\sum_{j=1}^J a_j \, f_{{\Phi}_j} \in {\mathcal N}_{W_1 + \dotsc + W_J,L}.
$$
\end{proposition}
\begin{proof}
Let $(M_j^{(L)},b_j^{(L)})$ be the output weight-bias pair for the network $\Phi_j$. The desired network can be constructed by stacking the $\Phi_j$'s vertically, such that all networks share a single output neuron with a bias of $\sum_{j=1}^J a_j \, b_j^{(L)}$. The output weight matrix is then obtained by scaling each $M_j^{(L)}$ by the corresponding $a_j$.    
\end{proof}

We end this section by a simple proposition that will be useful in our constructive proofs.
\begin{proposition}\label{prop:scalar_mult}
Let $\Phi \in {\mathcal N}_{W,L}^{{\mathbb R}}$ be a ReLU network realizing the function $f_{\Phi} = f_{\Phi}(x)$, with $x \in {\mathbb R}$. 
Then, for any scalar $\gamma \in \mathbb{R}$, there exists a network $\tilde{\Phi} \in {\mathcal N}_{W,L}^{\mathbb{R}}$, referred to as the scalar multiplication network, with the same complexity as $\Phi$ and realizing the function $f_{\tilde{\Phi}}(x) = \gamma \, f_{\Phi}(x)$. 
\end{proposition}
\begin{proof}
Let $(M^{(L)},b^{(L)})$ be the output weight-bias pair for the network $\Phi$. 
To construct the scalar multiplication network $\tilde{\Phi}$, we keep the architecture and all weight-bias pairs of the hidden layers identical to $\Phi$. For the output layer, we replace the weight-bias pair $(M^{(L)}, b^{(L)})$ with $(\gamma \, M^{(L)}, \gamma \, b^{(L)})$. This ensures that the output of the network $\tilde{\Phi}$ is $f_{\tilde{\Phi}}(x) = \gamma \, f_{\Phi}(x)$, thereby completing the construction while preserving the overall network complexity. 
\end{proof}

\section{Density of deep ReLU networks}
\label{sec:density-deep1}

So far, we have discussed the density question regarding the approximation of continuous target functions by networks with one hidden layer and non-polynomial continuous activation functions. 
In the remainder of this chapter, we will consider the more practical case of ``deep" networks with ReLU activation functions. We will discuss the expressive power of depth and show that deep, narrow ReLU networks are also dense in the space of continuous functions.

We closely follow \cite{Yarotsky:17} and \cite{Daubechies_etal:21,Elbrachter_etal:21}, utilizing the basic elements such as the ``sawtooth" construction developed in \cite{Telgarsky:15}. However, there will be minor changes, such as in the construction of neural networks, compared to the proofs presented in those references. Other relevant references will be cited when an idea, formula, or concept is used. Throughout this section, we limit our discussion to one-dimensional input domains where $X = [0,1]$. Extension to multiple dimensions for general $X \subset {\mathbb R}^d$ can be done similarly, using identity ReLU networks and the strategy presented in the previous section. 
We will start with a simple example to illustrate the power of depth.

\subsection{Expressive power of deep ReLU networks: an intuitive 1D example}
\label{sec:relu-x2}

As an intuitive example, we consider the quadratic function on the interval $[0,1]$, 
$$
f(x) = x^2, \qquad x \in [0,1].
$$
Our goal is to construct a ReLU
network approximant, say $f_{\Phi}$, of $f$ with the error
\begin{equation}\label{accuracy_x2}
|| f - f_{\Phi} ||_{L^{\infty}([0,1])} \le \varepsilon,
\end{equation}
with minimal complexity, i.e. with the number of neurons and layers as small as possible. Indeed, we will show that this can be
achieved with complexity $W=3$ and $L = {\mathcal O}(\log_2 \varepsilon^{-1})$. Our
construction will use the ``sawtooth'' function that (first) appeared in \cite{Telgarsky:15}.

\begin{figure}[!h]
\vspace{-0.6cm}
\centering
\includegraphics[width=0.26\linewidth]{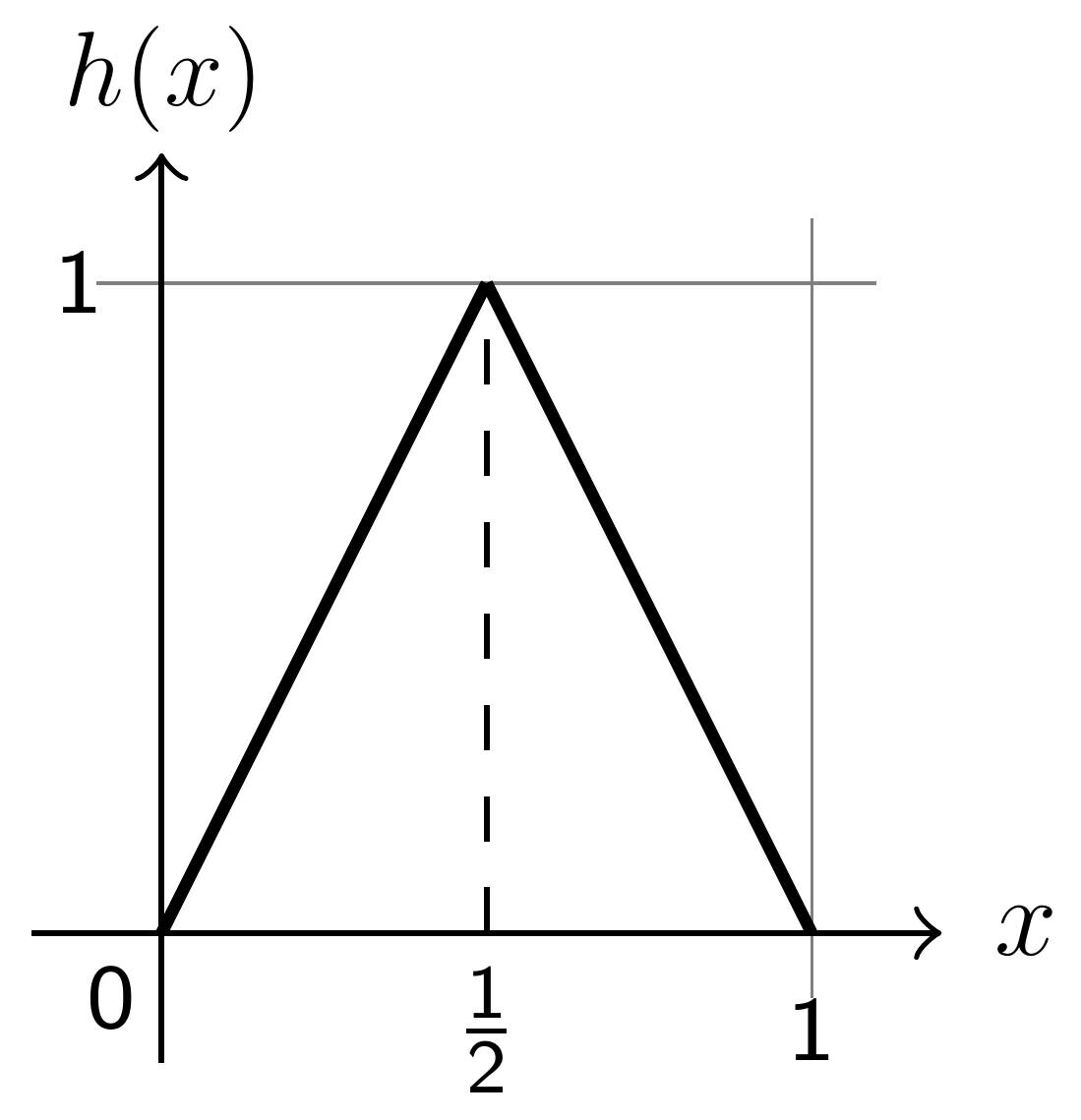}
        \vspace{-0.4cm}
\caption{The hat function $h(x)$ on the interval $[0,1]$.}
\label{fig:hat}
\end{figure}

\medskip
\noindent
{\bf Sawtooth function made by the composition of hat functions.}
Consider the hat function (also known as tent or triangle function), displayed in Figure \ref{fig:hat},
$$
h(x) = \left\{ \begin{array}{l l}
2 \, x, & \qquad 0 \le x < 1/2, \\
2 (1-x), &  \qquad 1/2 \le x \le 1.
\end{array} \right.
$$

\begin{figure}[!h]
\vspace{-0.1cm}
\centering
\includegraphics[width=0.47\linewidth]{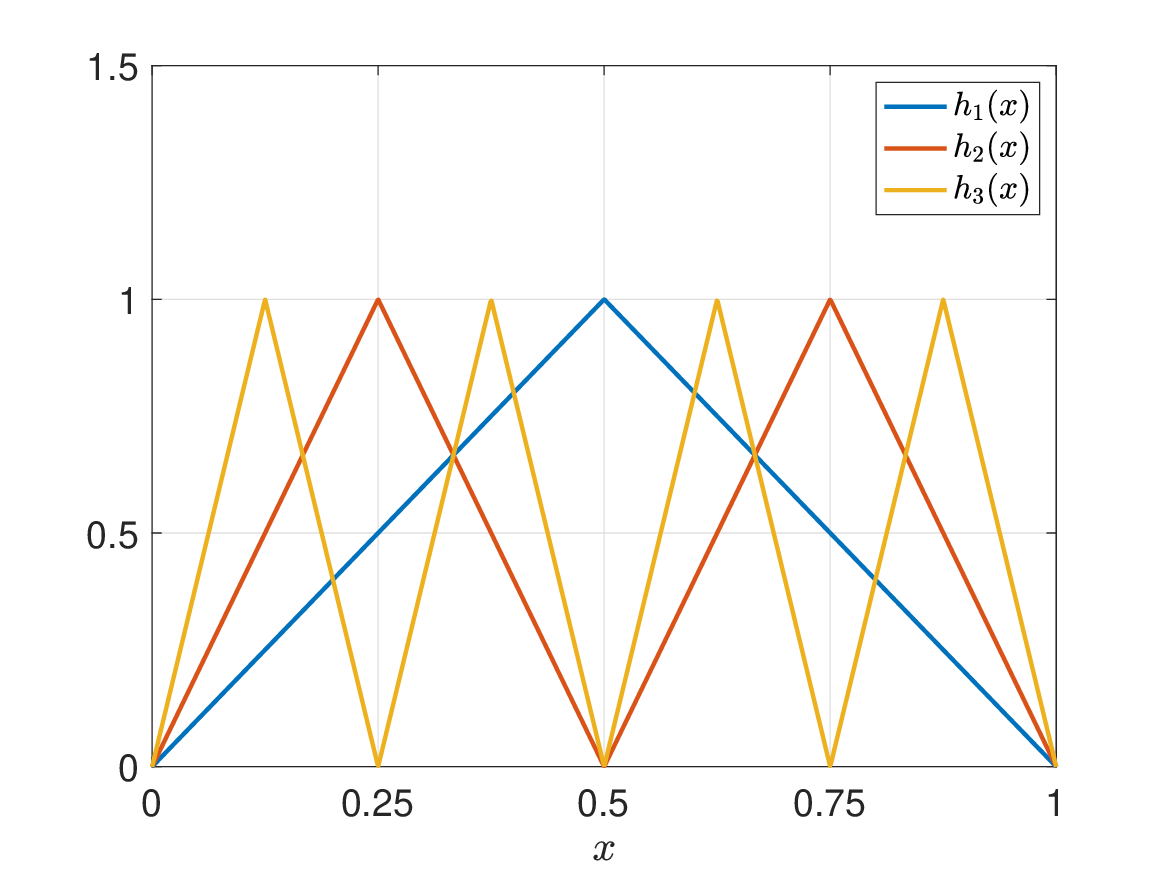}
        \vspace{-0.4cm}
\caption{The sawtooth function $h_s(x)$ on the interval $[0,1]$ with $s=1,2,3$.}
\label{fig:sawtooth}
\end{figure}

We let 
\begin{equation}\label{eqn:sawtooth}
h_0(x)=x, \qquad h_1(x)=h(x), \qquad h_s(x) = \underbrace{{h \circ h \circ \dotsc \circ h}}_{s-1 \, \, \text{compositions}} (x), \qquad s \ge 2,
\end{equation}
where $h_s(x)$, with $s \ge 2$, denotes by  the $s$-fold composition of $h$ with
itself. 
Telgarsky \cite{Telgarsky:15} has shown that $h_s$ is a sawtooth
function with $2^{s-1}$ evenly distributed teeth (or tents), where
each application of $h$ doubles the number of teeth; see Figure
\ref{fig:sawtooth}.

We make two observations and will discuss each in turn:
\begin{itemize}
\item $f(x)=x^2$ can be approximated by linear combinations of $h_0,
  h_1, h_2, \dotsc$. 

\item $h_0, h_1, h_2,\dotsc$ and hence their linear combinations can be expressed by a ReLU network.
\end{itemize}

\begin{figure}[!h]
\vspace{-0.5cm}
\centering
    \includegraphics[width=0.53\linewidth]{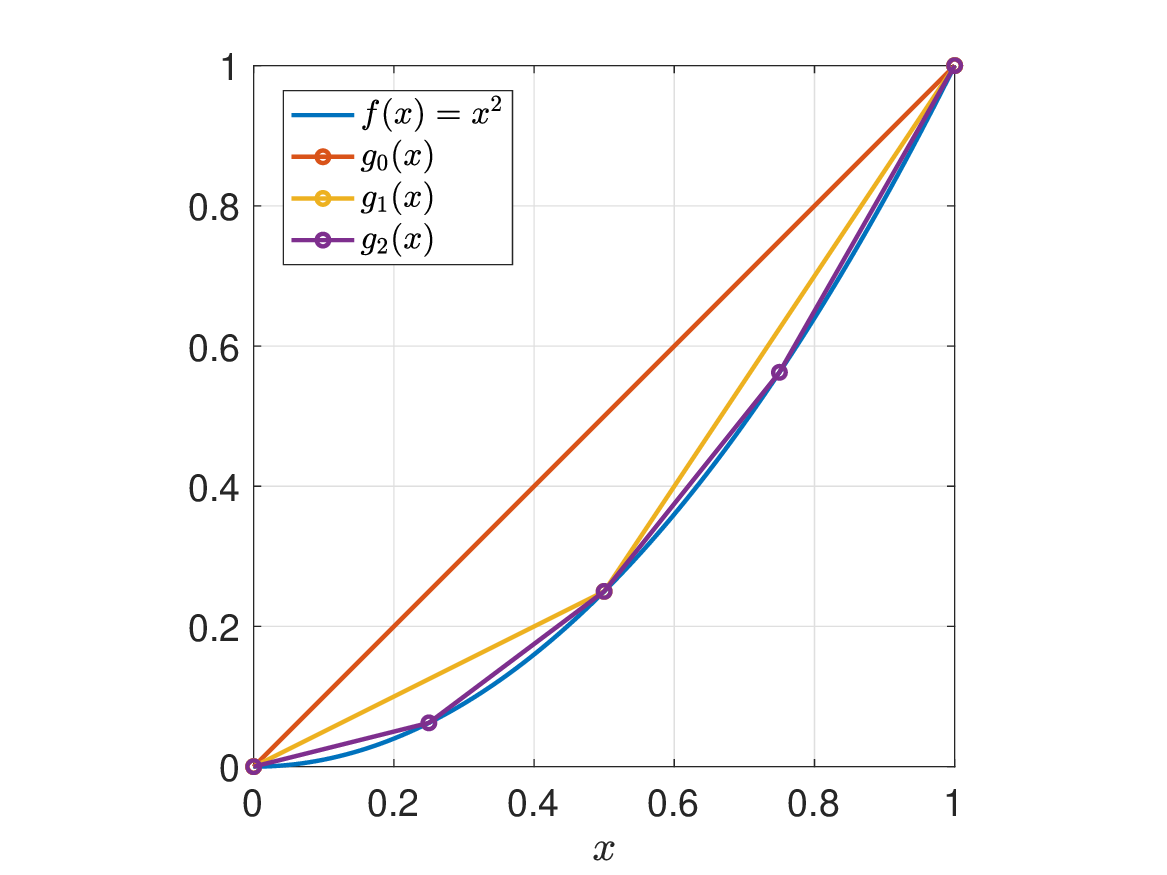}
    \vspace{-0.2cm}
\caption{Piecewise-linear approximation $g_m(x)$ of $f(x) = x^2$ at $2^m+1$
  evenly distributed grid points on the
  interval $[0,1]$, with $m=0,1,2$.}
\label{fig:x2_piecewise_lin}
\end{figure}

\medskip
\noindent
{\bf Observation 1.} 
We start with approximating $f(x)=x^2$ by a piecewise-linear interpolation $g_m$ of $f$ on a uniform grid of $2^m+1$ evenly distributed grid points $\{
\frac{i}{2^m} \}_{i=0}^{2^m}$; see Figure
\ref{fig:x2_piecewise_lin}. Note that by definition of interpolation, we have
$$
g_m(x) = x^2, \qquad x=\frac{i}{2^m}, \qquad i=0, 1,
\dotsc, 2^m.
$$

It is not difficult to see that refining the interpolation from
$g_{m-1}$ to $g_m$ amounts to adjusting it by a function proportional
to a sawtooth function, 
$$
g_{m-1}(x) - g_{m}(x) = \frac{h_m(x)}{2^{2 \, m}}.
$$
We hence obtain
\begin{equation}\label{x2_lin_comb}
g_m(x) = x - \sum_{s=1}^m \frac{h_s(x)}{2^{2 \, s}}.
\end{equation}
That is, the function $f(x)=x^2$ can be approximated by linear combinations of $h_0,
  h_1, h_2, \dotsc$. 

It is to be noted that the construction of $g_m$ only involves ${\mathcal O}(m)$
linear operations, including compositions of $h$. 
This demonstrates the power of composition. For instance, building $h_m$ (and thus $g_m$) without composition would require ${\mathcal O}(2^m)$ operations. However, with composition, we need only ${\mathcal O}(m)$ operations, significantly reducing the complexity from exponential to linear. As we will see later, this reduction is a major reason why deep learning, which relies on multiple applications of composition, has been so successful in solving science and engineering problems.

\begin{figure}[!h]
\centering
    \includegraphics[width=0.5\linewidth]{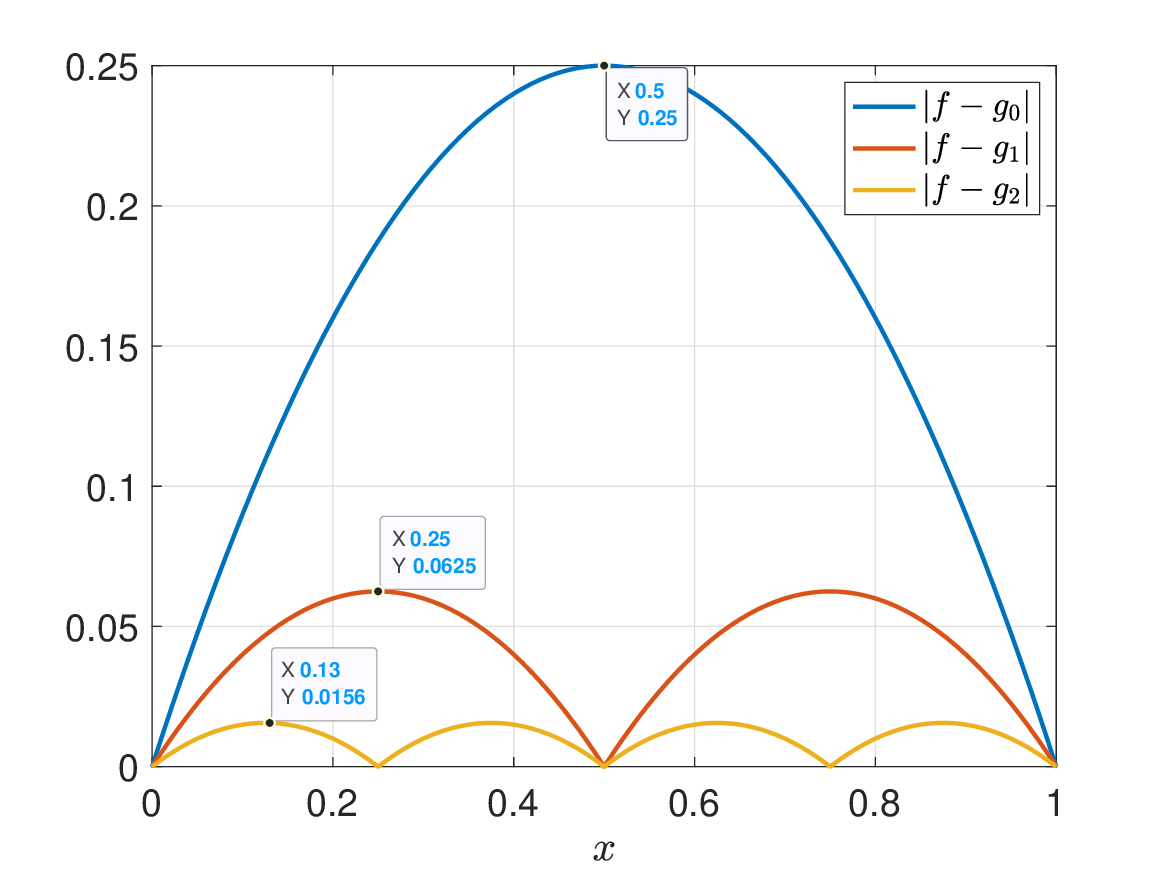}
    \vspace{-0.25cm}
\caption{The error $|f(x)-g_m(x)|$ in the approximation of $f(x) =
  x^2$ by a piecewise linear interpolant $g_m$ at $2^m+1$
  evenly distributed grid points on the
  interval $[0,1]$, with $m=0,1,2$. This suggests that with increasing
  $m$, the maximum
  error decays as $2^{-2(m+1)}$.}
\label{fig:x2_error}
\end{figure}

We can also derive a formula for the approximation error $|| f(x) - g_m(x) ||_{L^{\infty}([0,1])}$. 
For this purpose, we start with conducting a simple numerical experiment that aids in deriving a theoretical bound. Specifically, we compute and plot the error $|f(x)-g_m(x)|$ versus $x \in [0,1]$ for the
first three approximations $g_m$ with $m=0,1,2$. 
The results are depicted in Figure \ref{fig:x2_error}. 

Clearly, the initial maximum error is
$$
\max_{x \in [0,1]} |f(x)-g_0(x)| = \max_{x \in [0,1]} |x^2-x| = 1/4.
$$
This can be easily shown by setting the derivative of
$x^2-x$ to zero (i.e. $2 x - 1 = 0$), implying that the maximum occurs at $x=1/2$ and is equal to $|(1/2)^2 - (1/2)| = 1/4$. Figure
\ref{fig:x2_error} also illustrates that increasing $m$ by one multiplies the maximum error by $1/4$. Consequently, the figure suggests that the maximum error is $2^{-2(m+1)}$, i.e.,
\begin{equation}\label{x2_piecewiselin_error}
|| f(x) - g_m(x) ||_{L^{\infty}([0,1])} = 2^{-2(m+1)}, \qquad m=0,1,
2, \dotsc.
\end{equation}
Interestingly, this numerical result also provides insight into how to theoretically prove \eqref{x2_piecewiselin_error}. For every $m=0, 1, 2, \dotsc$, the error $|f(x)-g_m(x)|$ is identical on each subinterval between two consecutive points of interpolation. Specifically, we have
$$
\max_{x \in [0,1]} |f(x)-g_m(x)| = \max_{x \in [0,1/2^m]} |f(x)-g_m(x)|.
$$
Using the right-hand side of this formula, \eqref{x2_piecewiselin_error} follows.

\begin{figure}[!h]
\vspace{-0.7cm}
\centering
\includegraphics[width=0.45\linewidth]{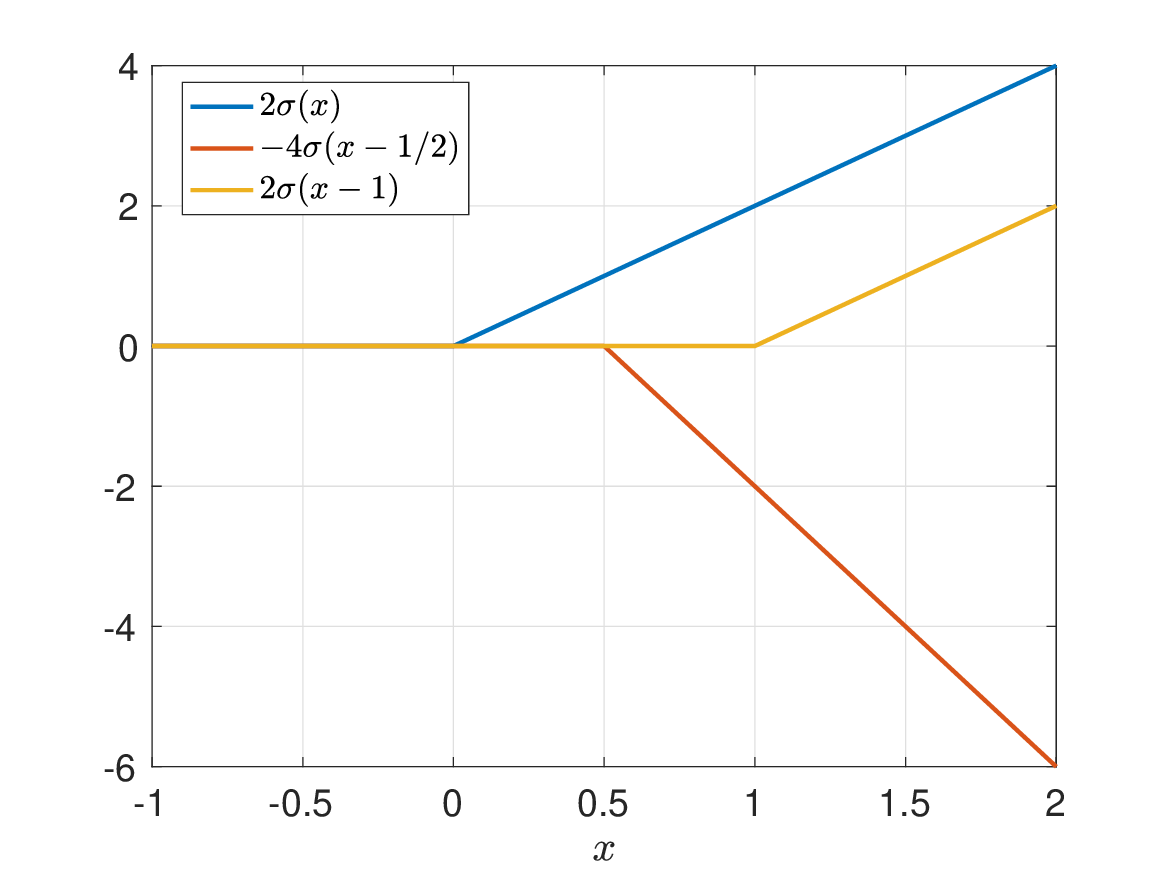}
 \includegraphics[width=0.45\linewidth]{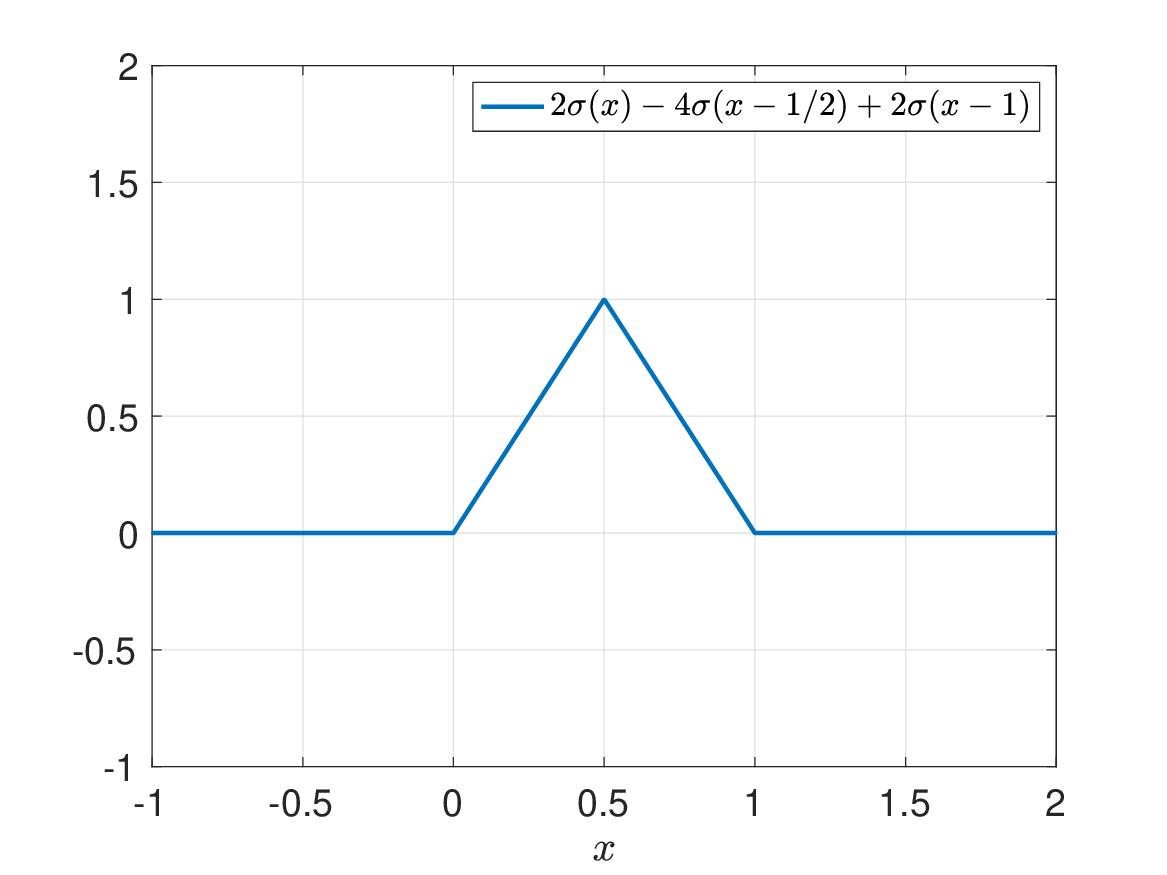}
 \vspace{-0.3cm}
\caption{The three ReLU functions (left) that combine to form the hat function (right).}
\label{fig:relu_hat}
\end{figure}

\medskip
\noindent
{\bf Observation 2.} 
We further note that the hat function
$h=h(x)$ can be exactly realized (or expressed) by a ReLU network. To derive such a network, we first observe that $h$ can be expressed  as a linear combination of three ReLU functions\footnote{Note that two ReLU functions will be enough when $x \in [0,1]$. In this case we have $h(x) = 2 \, \sigma(x) - 4 \, \sigma(x - 1/2)$.} (see Figure \ref{fig:relu_hat}):
$$
h(x) = 2 \, \sigma(x) - 4 \, \sigma(x - 1/2) + 2 \, \sigma(x-1), \qquad x \in {\mathbb R}. 
$$

\begin{figure}[!h]
\centering
    \includegraphics[width=0.4\linewidth]{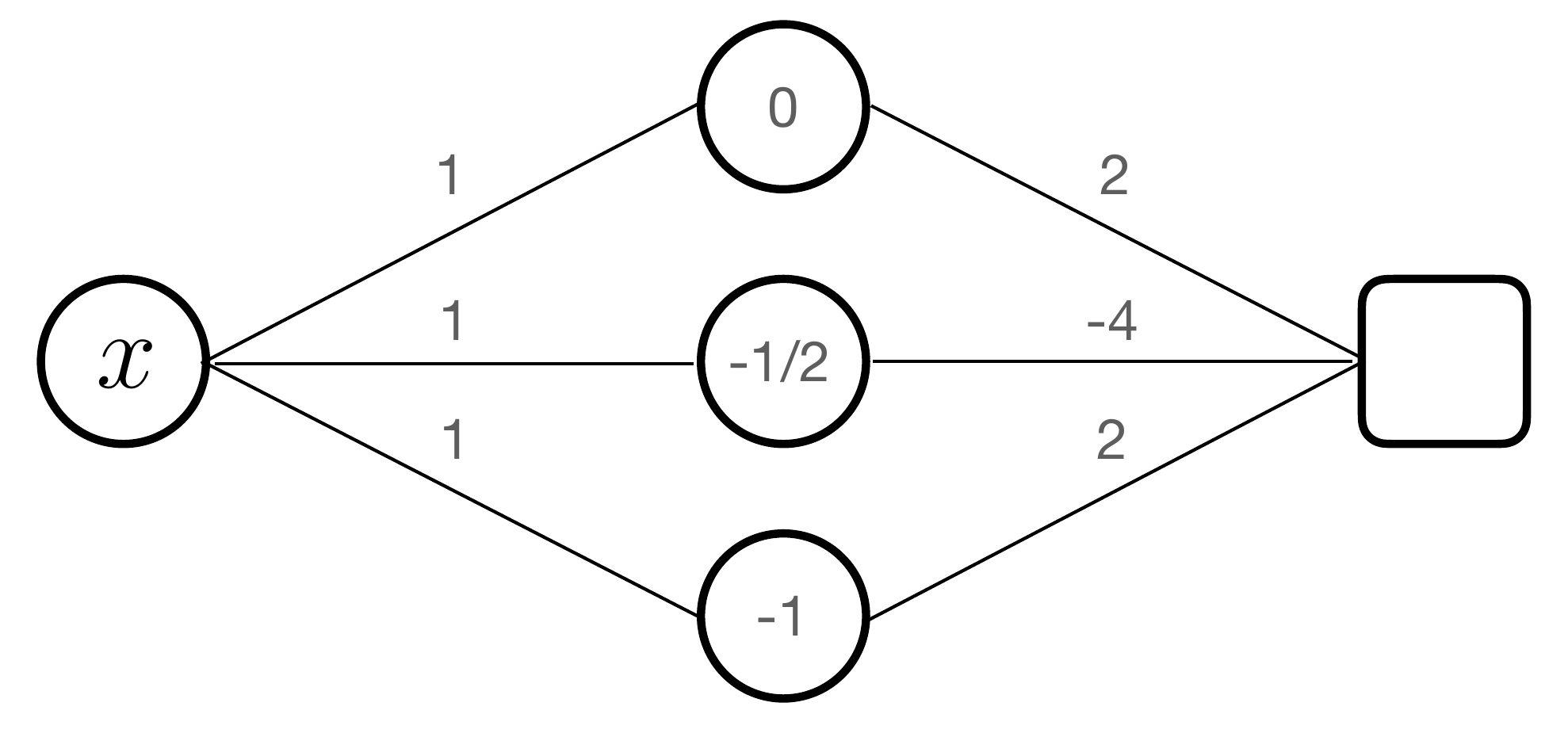}
        \vspace{-0.25cm}
\caption{The ReLU network that realizes $h(x) = 2 \, \sigma(x) - 4 \, \sigma(x - 1/2) + 2 \, \sigma(x-1)$.}
\label{fig:relu_net_hat}
\end{figure}

One can then easily see that $h(x)$ is given by a ReLU network, say $\Phi_h$, with one hidden layer ($L=1$) and $W=3$ neurons in the hidden layer(or $W=2$ when $x \in [0,1]$),
$$
f_{\Phi_h}(x) := A_1 \circ \sigma \circ A_0(x),
\quad 
A_0 (x) = 
\left( \begin{array}{c}
1 \\
1\\
1
\end{array} \right) \, x +
\left( \begin{array}{c}
0 \\
-1/2\\
-1
\end{array} \right), 
\quad
A_1({\bf z}) = (2 \ \ -4 \ \ 2) \, 
\left( \begin{array}{c}
z_1 \\
z_2\\
z_3
\end{array} \right).
$$
Figure \ref{fig:relu_net_hat} shows a graph representation of
$f_{\Phi_h}(x) = h(x)$. 

\begin{figure}[!h]
\centering
        \includegraphics[width=0.65\linewidth]{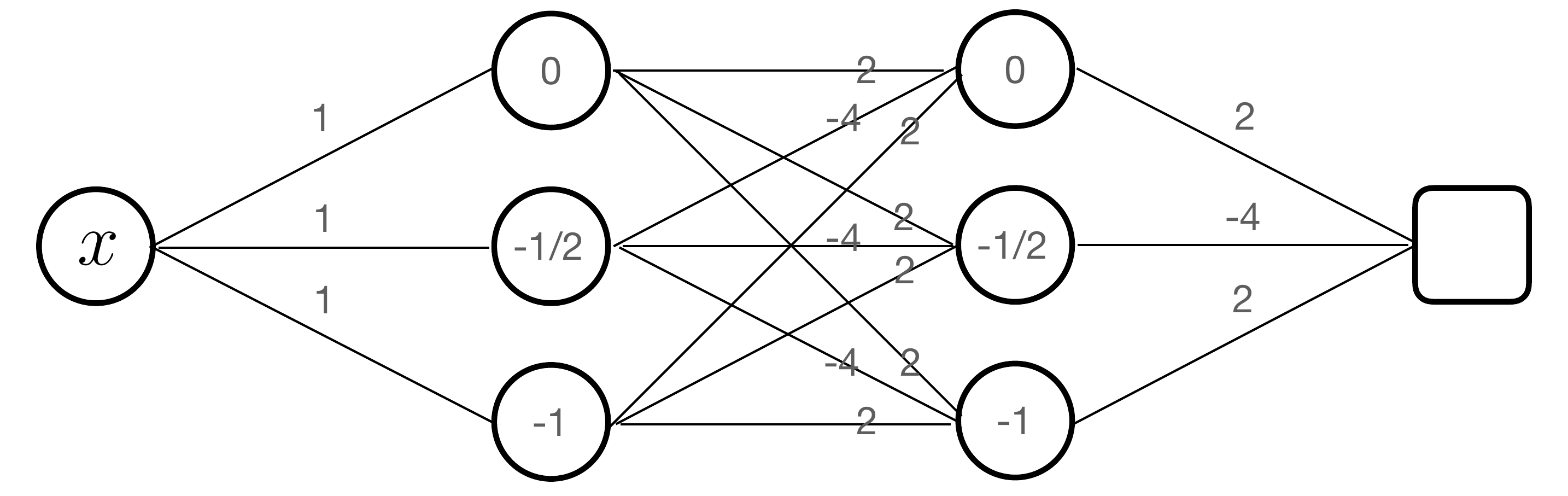}
\caption{The three-layer ReLU network that realizes $h_2(x)$.}
\label{fig:relu_net_h2}
\end{figure}

It will now be straightforward to construct a ReLU network, say $\Phi_{h_s}$, with $L=s$ hidden layers and $W=3$ neurons in each hidden layer (or $W=2$ when $x \in [0,1]$) that
realizes $h_s(x)$, for any $s\ge 2$. One such network is given by
\begin{equation}\label{eqn:network_hs}
f_{\Phi_{h_s}}(x) := A_{s} \circ \sigma \circ A_{s-1} \circ \sigma \circ
\dotsc \circ A_1 \circ \sigma \circ A_0(x), 
\end{equation}
where
$$
A_0 (x) = 
\left( \begin{array}{c}
1 \\
1\\
1
\end{array} \right) \, x +
\left( \begin{array}{c}
0 \\
-1/2\\
-1
\end{array} \right), 
\quad
A_{s}({\bf z}) = (2 \ \ -4 \ \ 2) \, 
\left( \begin{array}{c}
z_1 \\
z_2\\
z_3
\end{array} \right),
$$
and
$$
A_{\ell} (x) = 
\left( \begin{array}{c c c}
2 & -4 & 2 \\
2 & -4 & 2 \\
2 & -4 & 2 
\end{array} \right) \, 
\left( \begin{array}{c}
z_1 \\
z_2\\
z_3
\end{array} \right) +
\left( \begin{array}{c}
0 \\
-1/2\\
-1
\end{array} \right), 
\qquad
\ell = 1, \dotsc, s-1.
$$
Figure \ref{fig:relu_net_h2} shows the graph representation of
$f_{\Phi_{h_2}}(x) = h_2(x)$, with $L=2$ hidden layers. 

\begin{figure}[!h]
\centering
\includegraphics[width=7.2cm,height=5.5cm]{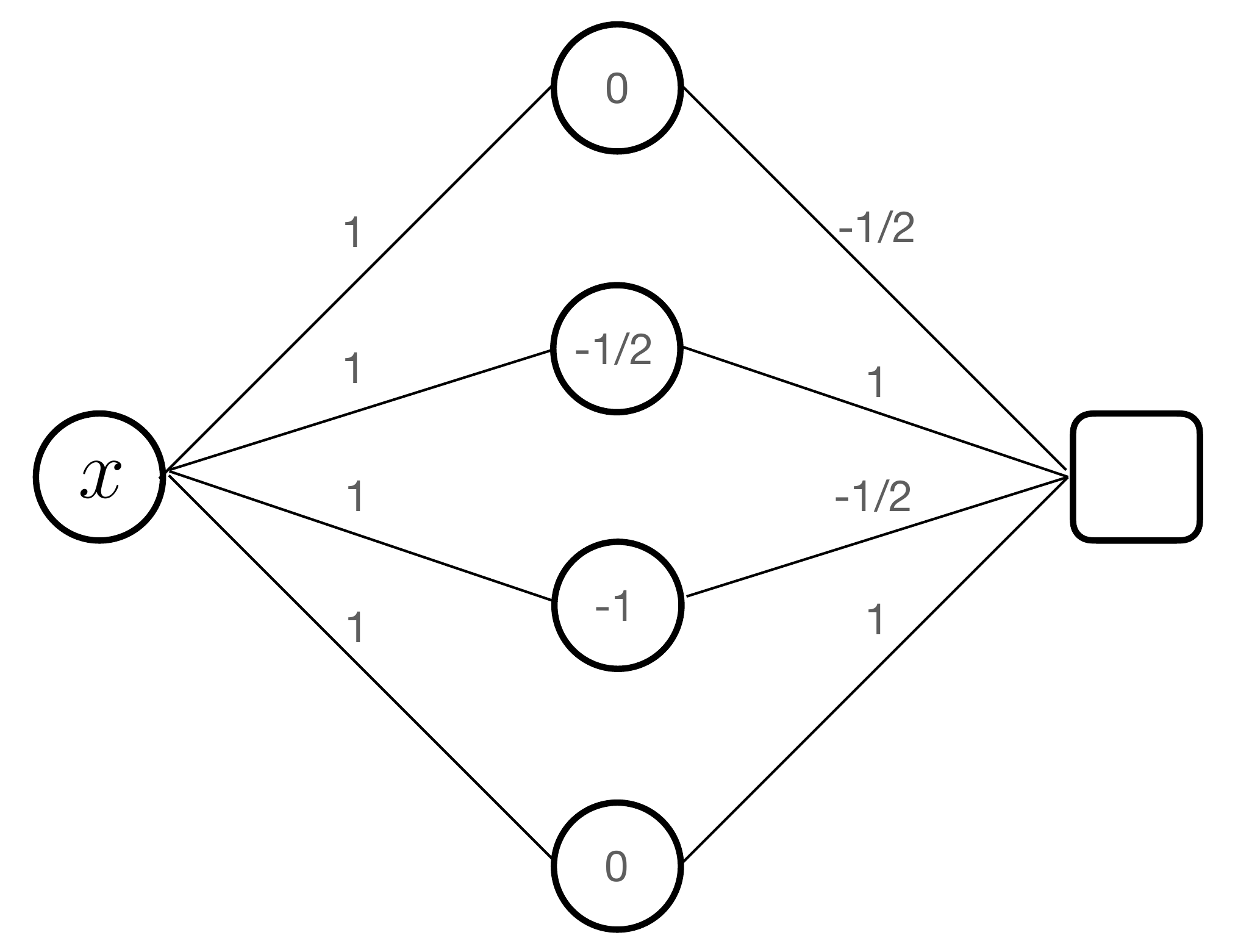}
\includegraphics[width=9.2cm,height=5.5cm]{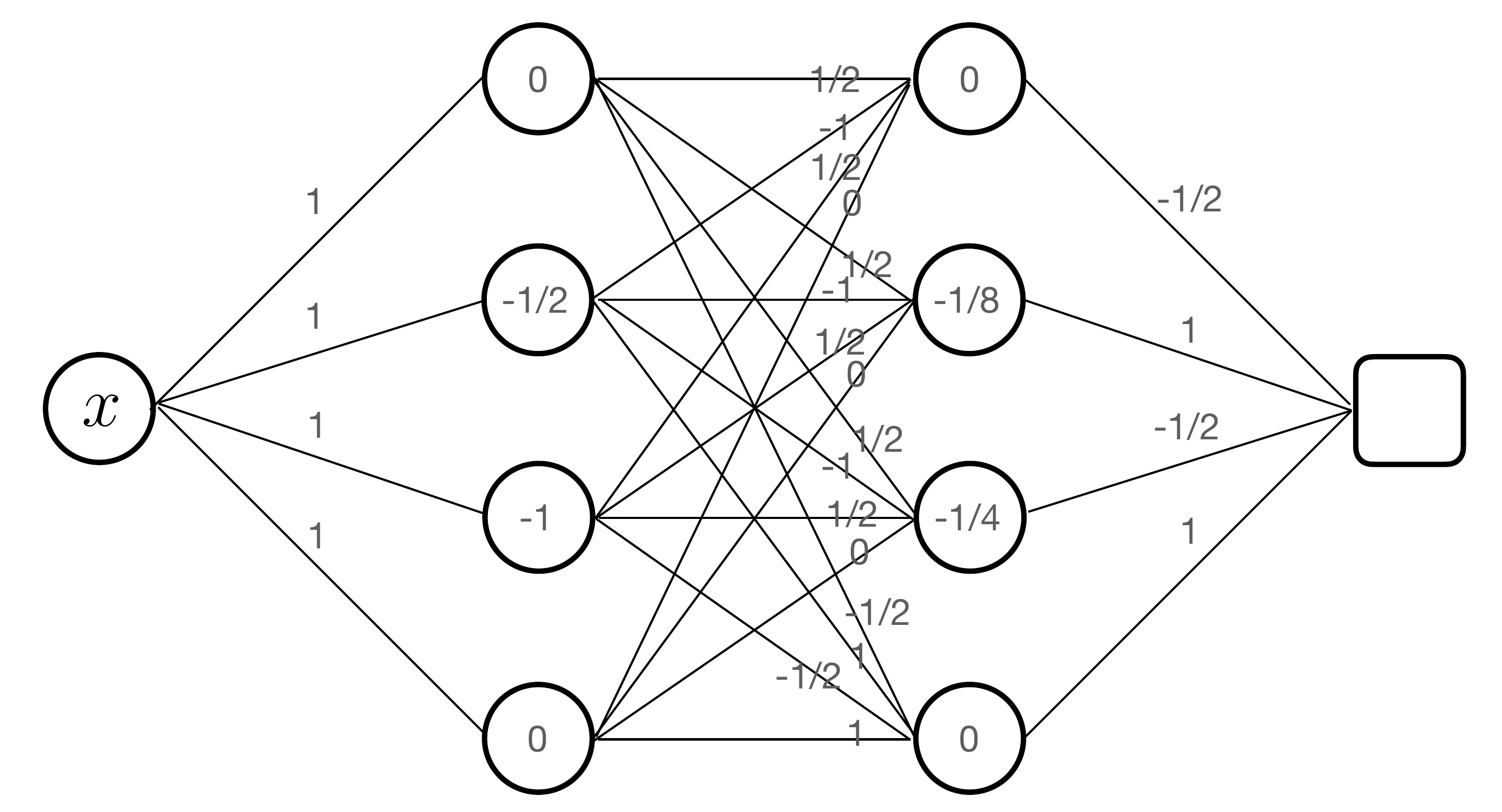}
\vspace{-0.3cm}
\caption{ReLU networks that realize $g_1(x)$ (left) and $g_2(x)$ (right).}
\label{fig:relu_net_x2_m12}
\end{figure}

Finally, we can construct a ReLU network, say $\Phi_{g_m}$, with $L=m$ hidden layers and $W=4$ neurons in the hidden layer (or $W=3$ when $x \in [0,1]$) that realizes $g_m(x)$, for any $m\ge 1$. One such network is given by
$$
f_{\Phi_{g_m}}(x) := A_{m} \circ \sigma \circ 
\dotsc \circ A_1 \circ \sigma \circ A_0(x), 
$$
where
$$
A_0 (x) = 
\left( \begin{array}{c}
1 \\
1\\
1\\
1
\end{array} \right) \, x +
\left( \begin{array}{c}
0 \\
-1/2\\
-1\\
0
\end{array} \right), 
\quad
A_{m}({\bf z}) = (-1/2 \ \ 1 \ \ -1/2 \ \ 1) \, 
\left( \begin{array}{c}
z_1 \\
z_2\\
z_3\\
z_4
\end{array} \right),
$$
and
$$
A_{\ell} ({\bf z}) = 
\left( \begin{array}{c c c c}
1/2 & -1 & 1/2 & 0\\
1/2 & -1 & 1/2 & 0 \\
1/2 & -1 & 1/2 & 0 \\
-1/2 & 1 & -1/2 & 1
\end{array} \right) \, 
\left( \begin{array}{c}
z_1 \\
z_2\\
z_3\\
z_4
\end{array} \right) +
\left( \begin{array}{c}
0 \\
-2^{-2 \, \ell + 1}\\
-2^{-2 \, \ell + 2}\\
0
\end{array} \right), 
\qquad
\ell = 1, \dotsc, m-1.
$$
The graph representations of $f_{\Phi_{g_1}}(x) = x - \frac{1}{4} \, h_1(x)$ and $f_{\Phi_{g_2}}(x) = x -
\frac{1}{4} \, h_1(x) - \frac{1}{16} \, h_2(x)$ are shown in Figure \ref{fig:relu_net_x2_m12}.

We state the main result on the approximation of the quadratic function by ReLU networks.

\medskip
\noindent
\begin{lemma}\label{lem:x2} 
The function $f(x)=x^2$ can be approximated within any small error $\varepsilon \in
(0, 1/2)$ in the sense of \eqref{accuracy_x2} by a standard ReLU network with width $W = 3$ for $x\in [0,1]$, or $W = 4$ for general $x \in {\mathbb R}$, and depth $L = {\mathcal O}(\log_2 \varepsilon^{-1})$.
\end{lemma}

This result illustrates the power of composition in neural networks, making them efficient approximants in the sense that their complexity increases logarithmically with the reciprocal of the approximation error. 
We note that the input dimension is one here, and hence this result does not address the effect of composition on dimension.


\subsection{Approximating product of two numbers by ReLU networks}
\label{sec:relu-xy}

We next consider two real numbers $x \in{\mathbb R}$ and $y \in {\mathbb R}$, where $\max \{ |x|, |y| \} \le D$, with $D > 0$. Utilizing the identity formula $xy=\frac{1}{4} ((x+y)^2 - (x-y)^2)$ and the results of Section \ref{sec:relu-x2} in approximating the quadratic function by ReLU networks, we show that the product $x y$ can also be well approximated by a ReLU network.

\medskip
\noindent
\begin{lemma}\label{lem:xy}
The function $f(x,y)=x y$ on the domain $[-D,D]^2$,
where $D>0$, can be approximated by a ReLU network $\Phi_{D,\varepsilon}$ within any small error $\varepsilon \in
(0, 1/2)$ in the sense 
$$
|| f_{\Phi_{D,\varepsilon}} (x,y) - x y ||_{L^{\infty}([-D,D]^2)} \le \varepsilon,
$$
where the network has two input dimensions (one for $x$ and one for
$y$) and has the complexity 
$$
W(\Phi_{D,\varepsilon}) \le 5, \qquad  
L(\Phi_{D,\varepsilon}) \le C \, (\log_2 \varepsilon^{-1} + \log_2 \lceil D \rceil).
$$
Moreover, if $x=0$ or $y=0$, then $f_{\Phi_{D,\varepsilon}} (x,y) = 0$. 
\end{lemma}

\begin{figure}[!h]
\centering
        \includegraphics[width=0.85\linewidth]{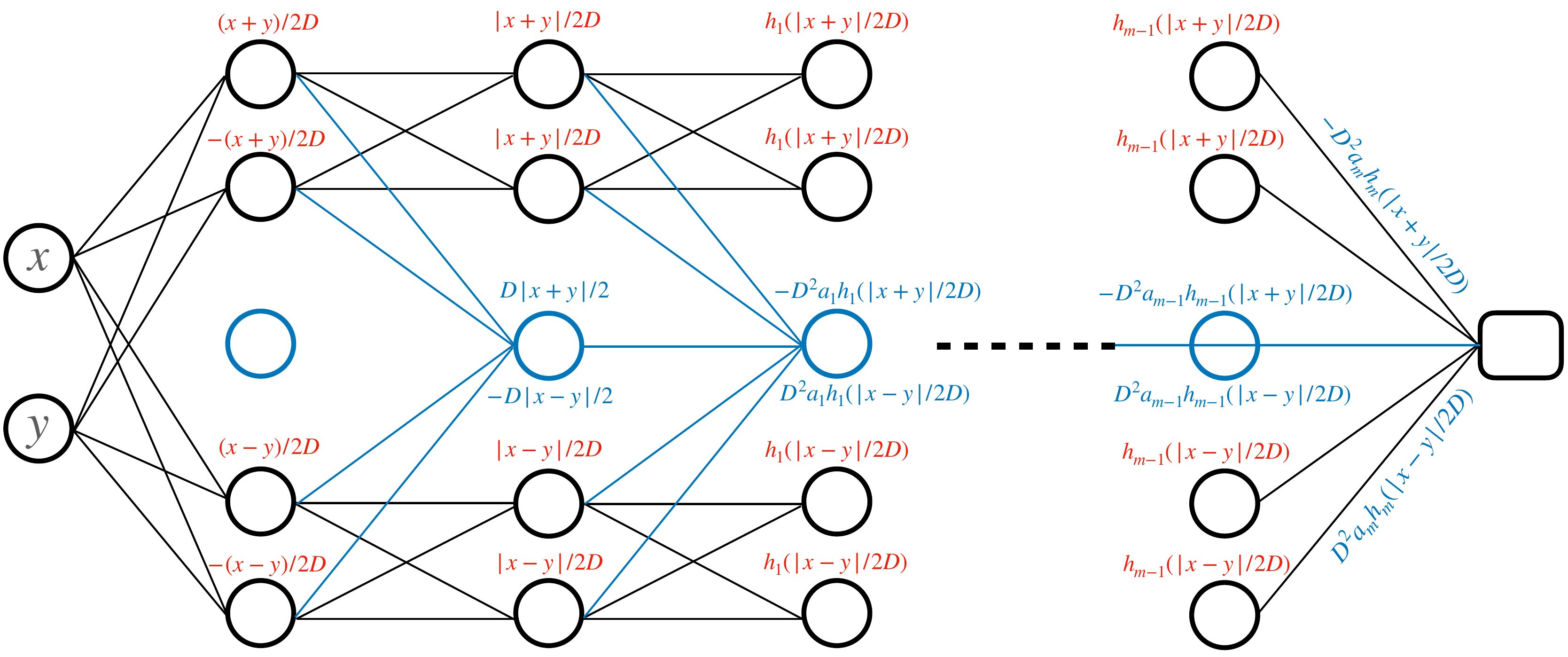}
\caption{Construction of the special network that approximates the product of $x$ and $y$. .}
\label{fig:network_xy}
\end{figure}

\medskip
\noindent
{\it Proof.} Without loss of generality, we can assume that $D \ge 1$. The case $D<1$ can always be replaced by $D=1$ and hence falls into the case $D \ge 1$ that we consider here. Let $\Phi_{\delta}$ be the network from Lemma \ref{lem:x2}, approximating $f(z) = z^2$, with $z \in [0,1]$, such that
$$
|| f_{\Phi_{\delta}} (z) - z^2 ||_{L^{\infty}([0,1])} \le \delta, \qquad z \in [0,1],
$$
with complexity $W(\Phi_{\delta}) = 3$ and $L(\Phi_{\delta}) = {\mathcal O}(\log_2 \delta^{-1})$. 
Noting that 
$$
xy = \frac{1}{4} ((x+y)^2 - (x-y)^2) =
D^2 \left( \left(\frac{x+y}{2D}\right)^2 - \left(\frac{x-y}{2D}\right)^2 \right),
$$
we consider the network $\Phi_{D,\varepsilon}$ that realizes
\begin{equation}\label{eq:subtract}
f_{\Phi_{D,\varepsilon}}(x,y) = D^2 
\left( 
f_{\Phi_{\delta}}\left(\frac{|x+y|}{2D}\right) - f_{\Phi_{\delta}}\left(\frac{|x-y|}{2D}\right)  
\right),
\end{equation}
where each function in the right hand side is given by a linear combination of $m$ sawtooth functions, with $m= L(\Phi_{\delta}) = {\mathcal O}(\log_2 \delta^{-1})$:
\begin{equation}\label{eq:expression}
f_{\Phi_{\delta}}\left(\frac{|x+y|}{2D}\right) = \frac{|x+y|}{2 D} - \sum_{k=1}^m a_k h_k\left(\frac{|x+y|}{2D}\right), \ \ \ \  
f_{\Phi_{\delta}}\left(\frac{|x-y|}{2D}\right) = \frac{|x-y|}{2 D} - \sum_{k=1}^m a_k h_k\left(\frac{|x-y|}{2D}\right), 
\end{equation}
with $a_k = 4^{-k}$. 
With this setup, it easily follows that $f_{\Phi_{D,\varepsilon}} (x,y)
= 0$ if $x=0$ or $y=0$. Moreover, 
\begin{multline*}
|| f_{\Phi_{D,\varepsilon}} (x,y) - x y ||_{L^{\infty}([-D,D]^2)}
\le \\
D^2 \left(
\left\lVert f_{\Phi_{\delta}}\left(\frac{|x+y|}{2D}\right) -
\left(\frac{|x+y|}{2D}\right)^2\right\rVert_{L^{\infty}[-D,D]^2} 
+
\left\lVert f_{\Phi_{\delta}}\left(\frac{|x-y|}{2D}\right) -
  \left(\frac{|x-y|}{2D}\right)^2\right\rVert_{L^{\infty}[-D,D]^2} 
  \right) \le 
  \\
  2 D^2  || f_{\Phi_{\delta}} (z) - z^2 ||_{L^{\infty}([0,1])}  \le 2 D^2 \delta = \varepsilon,
\end{multline*}
provided we choose $\delta = \varepsilon / (2 D^2)$. 
It remains to compute the complexity of the network $\Phi_{D,\varepsilon}$, given in \eqref{eq:subtract}-\eqref{eq:expression}. 
To this end, we first construct a special ReLU network, say $\hat{\Phi}_{D,\varepsilon}$, by stacking two squaring networks on top of each other, each with width $W=2$ following from the fact that $|x+y|/2D, |x-y|/2D \in [0,1]$, and such that they share a collation channel and without any source channel, resulting in a total width of $W=2 \times 2 + 1 =5$. 
Figure \ref{fig:network_xy} shows the construction of $\hat{\Phi}_{D,\varepsilon}$; The top two channels and the bottom two channels represent the two squaring networks with arguments $|x+y|/2D$ and $|x-y|/2D$, respectively. The input to the compute neurons are displayed in red on top of each neuron. The computational weights of the first hidden layer use the identity formula $z = \sigma(z) - \sigma(-z)$ to generate $\pm(x+y)/2D$ and $\pm(x-y)/2D$, and the computational weights of the second hidden layer use the identity formula $|z| = \sigma(z) + \sigma(-z)$ to generate $|x+y|/2D$ and $|x-y|/2D$. The computational weights of the third layer generate $h_1$, and this pattern repeats until finally the computational weights of the output layer carry over $-D^2 a_2 h_2(|x+y|/2D)$ and $D^2 a_2 h_2(|x-y|/2D)$ to the output neuron. These two squaring networks share a collation channel (the middle channel in blue color) that collects intermediate computations, $D^2 (|x+y|/2D - |x-y|/2D)$, $- D^2 a_1 (h_1(|x+y|/2D) - h_1(|x-y|/2D))$, and so forth, and carries them over to the output. The special network $\hat{\Phi}_{D,\varepsilon}$ outputs the right hand side of \eqref{eq:subtract} using their expressions given in \eqref{eq:expression}. 
Clearly, the number of hidden layers of $\hat{\Phi}_{D,\varepsilon}$ is
$$
L(\hat{\Phi}_{D,\varepsilon}) = m  = 
{\mathcal O} (\log_2 \delta^{-1}) = {\mathcal O} (\log_2
\varepsilon^{-1} + \log_2 D),
$$ 
noting that $\delta = \varepsilon / (2 D^2)$. Since there exists a standard ReLU network $\Phi_{D,\varepsilon}$ with the same output and the same complexity as the special network $\hat{\Phi}_{D,\varepsilon}$, the proof is complete. \qed


\subsection{Approximating algebraic polynomials by ReLU networks}
\label{sec:relu-polynom}

Let $X= [0,1]$. Note again that this can be relaxed and extended to any bounded domain $X$ in ${\mathbb R}^d$. Utilizing the results from Sections
\ref{sec:relu-x2}-\ref{sec:relu-xy}, we show that any algebraic polynomial can be well
approximated by a ReLU network.

\smallskip
\noindent
\begin{lemma}\label{lem:polynom}
The polynomial $p_m(x)= a_0 + a_1 x + \dotsc + a_m x^m$ on the domain $[0,1]$ can be approximated by a ReLU network $\Phi_{\varepsilon}$ within any small error $\varepsilon \in
(0, 1/2)$ in the sense 
\begin{equation}\label{eq:lem_accuracy}
|| f_{\Phi_{\varepsilon}} (x) - p_m(x) ||_{L^{\infty}([0,1])} \le \varepsilon,
\end{equation}
where the network has one input dimension and the complexity (width and depth),
\begin{equation}\label{eq:betwork_complexity}
W \le 7, \qquad 
L \le C \, m \, (\log_2 \varepsilon^{-1} + \log_2 m + \log_2 || a ||_{\infty}),
\qquad
|| a ||_{\infty} := \max_i |a_i|.
\end{equation}
\end{lemma}
\begin{proof} 
For $m=1$, where we have a linear function $a_0 + a_1 x$, one can easily construct a network, with one hidden layer ($L=1$) consisting of a single neuron ($W=1$), that outputs the desired linear function, for example with input weight 1 and output weight $a_1$, and with bias 0 in the hidden neuron and bias $a_0$ in the output neuron. Let $m \ge 2$, and set
$$
D_k := 1 + (k-1) \, \delta, \quad k \in {\mathbb N}, \qquad \delta >0. 
$$
Let $\Phi_{D_k,\delta}$ be the product network of Lemma \ref{lem:xy} with the error-complexity,
\begin{equation}\label{eq:prod_aux}
|| f_{\Phi_{D_k,\delta}} (x,y) - x y ||_{L^{\infty}([-D_k,D_k]^2)} \le \delta,
\qquad 
W_{k,\delta} \le 5, 
\qquad  
L_{k, \delta} \le  C (\log_2 \delta^{-1} + \log_2 D_k). 
\end{equation}
We introduce the recursive functions:
\begin{equation}\label{eq:recursive_aux}
f_{0,\delta}(x) = 1, \qquad
f_{1,\delta}(x) = x, \qquad
f_{k,\delta}(x) = f_{\Phi_{D_{k-1},\delta}}(x,f_{k-1,\delta}(x)), \qquad 
k \ge 2.
\end{equation}
We will first show by induction that 
\begin{equation}\label{eq:induct_aux}
|| f_{k,\delta} (x) - x^k ||_{L^{\infty}[0,1]} \le (k-1) \delta, \qquad k \ge 2.
\end{equation}
To this end, let $k=2$. Then, by the recursive formula \eqref{eq:recursive_aux} and \eqref{eq:prod_aux},
$$
|| f_{2,\delta} (x) - x^2 ||_{L^{\infty}[0,1]} 
=
|| f_{\Phi_{D_1,\delta}}(x,x) - x^2 || _{L^{\infty}[0,1]} \le \delta,
$$
and hence \eqref{eq:induct_aux} holds for $k=2$. Now suppose that \eqref{eq:induct_aux} holds for $k-1$ (and $k \ge 3$):
\begin{equation}\label{eq:induct_aux2}
|| f_{k-1,\delta} (x) - x^{k-1} ||_{L^{\infty}[0,1]} \le (k-2) \delta, \qquad k \ge 3.
\end{equation}
Then we have
$$
|| f_{k-1,\delta} (x) ||_{L^{\infty}[0,1]} \le
|| x^{k-1} ||_{L^{\infty}[0,1]} + || f_{k-1,\delta} (x) - x^{k-1} ||_{L^{\infty}[0,1]} \le
1 + (k-2) \delta = D_{k-1},
$$
and hence
\begin{multline*}
|| f_{k,\delta} (x) - x^k ||_{L^{\infty}[0,1]} 
\le 
|| f_{k,\delta} (x) - x \, f_{k-1,\delta} (x) ||_{L^{\infty}[0,1]} +
|| x \, f_{k-1,\delta} (x) - x^k ||_{L^{\infty}[0,1]} \\
\le 
|| f_{\Phi_{D_{k-1},\delta}}(x,f_{k-1,\delta}(x)) - x \, f_{k-1,\delta} (x) ||_{L^{\infty}[0,1]} +
|| x||_{L^{\infty}[0,1]} \,  || f_{k-1,\delta} (x) - x^{k-1} ||_{L^{\infty}[0,1]} \\
\le \delta + (k-2) \, \delta = (k-1) \, \delta.
\end{multline*}
This completes the proof of \eqref{eq:induct_aux}. Next let,
\begin{equation}\label{eq:network_fun}
f_{\Phi_{\varepsilon}} (x) = \sum_{k=0}^m a_k \, f_{k,\delta}(x).
\end{equation}
With this choice, and thanks to \eqref{eq:induct_aux}, we get
$$
|| f_{\Phi_{\varepsilon}} (x) - p_m(x) ||_{L^{\infty}([0,1])} \le \sum_{k=2}^m |a_k| \, || f_{k,\delta} (x) - x^k ||_{L^{\infty}[0,1]} 
\le
|| a ||_{\infty} \, \delta \, \sum_{k=2}^m (k-1) \le || a ||_{\infty} \, \delta \, (m-1)^2.
$$
The desired accuracy \eqref{eq:lem_accuracy} will be achieved if we choose $\delta = \varepsilon / || a ||_{\infty} (m-1)^2$. It remains to show that one can construct a network $\Phi_{\varepsilon}$ that realizes \eqref{eq:network_fun} with desired complexity \eqref{eq:betwork_complexity}. 
Figure \ref{fig:network_polynom} shows the construction of an equivalent special network, say $\hat{\Phi}_{\varepsilon}$, formed by stacking individual product networks $\Phi_{D_1, \delta}, \dotsc, \Phi_{D_{m-1}, \delta}$, with widths $W_{1,\delta}, \dotsc, W_{m-1,\delta}$ and depths $L_{1,\delta}, \dotsc, L_{m-1, \delta}$, satisfying \eqref{eq:prod_aux}. The output of each product network $\Phi_{D_k,\delta}$ is $f_{k, \delta}$. Noting that $f_{k, \delta}$ may take negative values, we utilize the identity relation $\sigma(f_{k,\delta}) - \sigma(- f_{k,\delta}) = f_{k, \delta}$ and introduce intermediate identity networks (displayed in red) to recover $f_{k, \delta}$ and giving it as input to the next product network. We also add a collation channel to collect and carry all intermediate computations $a_2 f_{2,\delta}, \dotsc, a_{m-1} f_{m-1,\delta}$ over to the output. A source channel is also needed as each product network takes both $x$ and the output of the previous product network as inputs. 
\begin{figure}[!h]
\centering
        \includegraphics[width=0.95\linewidth]{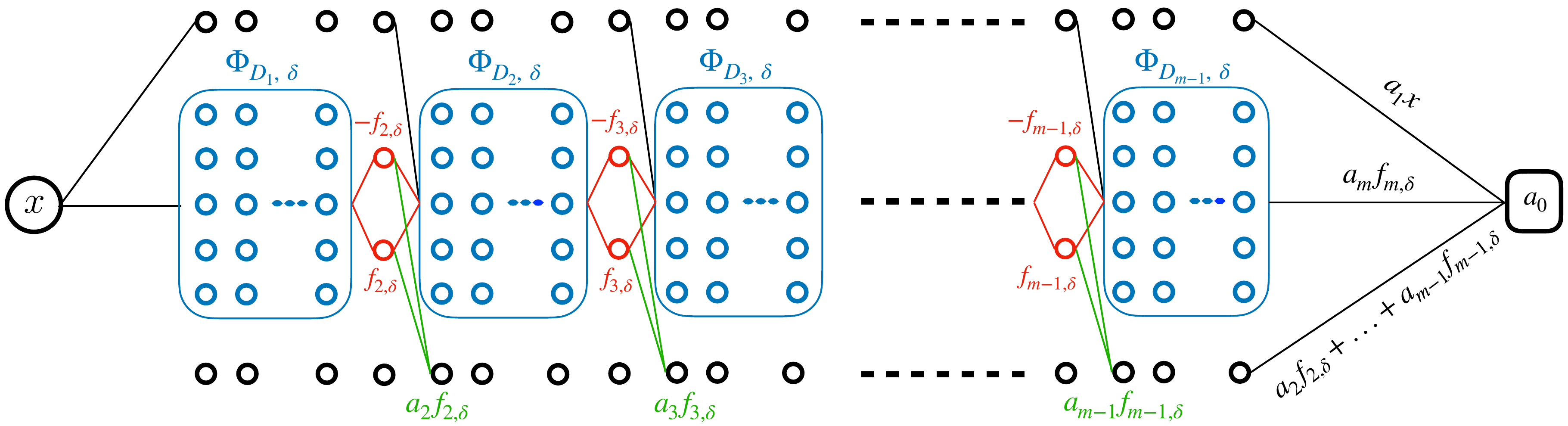}
\caption{Construction of the special network that approximates a polynomial of degree $m$, by stacking individual product networks and adding a source and a collation channel.}
\label{fig:network_polynom}
\end{figure} 
Clearly, such network has a width $W \le 7$ and a depth satisfying
$$
L = m-2 + \sum_{k=1}^{m-1} L_{k, \delta}  \le m-2 + \sum_{k=1}^{m-1} C_k (\log_2 \delta^{-1} + \log_2 D_k) \le C \, m \, (\log_2 \varepsilon^{-1} + \log_2 m + \log_2 || a ||_{\infty}),
$$
as desired, noting that $\delta = \varepsilon / || a ||_{\infty} (m-1)^2$. The proof is complete, as there exists a standard ReLU network $\Phi_{\varepsilon}$ with the same output and the same complexity as the special network $\hat{\Phi}_{\varepsilon}$.
\end{proof}

It is worth mentioning that a similar result can be shown for more general domains $X$, such as $X = [-D,D]$, where $D \in (0, \infty)$.

\smallskip
\noindent
\begin{lemma}\label{lem:polynom_b}
The polynomial $p_m(x)= a_0 + a_1 x + \dotsc + a_m x^m$ on the domain $[-D,D]$, with $D \in (0, \infty)$, can be approximated by a ReLU network $\Phi_{D,\varepsilon}$ within any small error $\varepsilon \in
(0, 1/2)$ in the sense 
$$
|| f_{\Phi_{D,\varepsilon}} (x) - p_m(x) ||_{L^{\infty}([-D,D])} \le \varepsilon,
$$
where the network has one input dimension and the complexity (width and depth),
$$
W \le 7, \quad 
L \le C \, m \, (\log_2 \varepsilon^{-1} + \log_2 m + m \, \log_2 \lceil D \rceil 
+ \log_2 || a ||_{\infty}),
\quad
|| a ||_{\infty} := \max_i |a_i|.
$$
\end{lemma}
Also see Proposition III.5 in \cite{Elbrachter_etal:21}, where a similar result is established with $W \le 9$.

\subsection{Proof sketch of density of deep ReLU networks}
\label{sec:proof-deepReLU}

We are now ready to prove the density of deep ReLU networks. We first recall the Weierstrass theorem that states every continuous function on a closed interval can be approximated to within arbitrary accuracy by a polynomial. Hence, by Lemma \ref{lem:polynom_b}, every continuous function on a closed interval can be approximated to within arbitrary accuracy by a deep ReLU network with a uniform width $W \le 7$. This shows that deep ReLU networks of finite width are dense in the space of continuous functions with respect
to the supremum norm.


\section{The power of depth}
\label{sec:depth-power}

In this section we will further explore the power of depth. 
Specifically, we will demonstrate that a large class of functions exhibiting self-similarity can be approximated by ReLU networks with significantly better approximation rates compared to any other classical approximation methods, such as free-knot linear splines. 

First, we will present a few compositional properties of standard and special ReLU networks (Section \ref{sec:depth1}). Then, we will use these properties to illustrate the power of depth for self-similar functions (Section \ref{sec:depth2}). Throughout this section, we will limit our discussion to one-dimensional input domains where $X = [0,1]$.

\subsection{Compositional properties of ReLU networks}
\label{sec:depth1}

We start with showing that the composition of several networks with the same width can be represented by a single network whose depth is the sum of the networks' depths.

\begin{proposition}\label{prop:composition}
For any $f_{{\Phi}_j} \in {\mathcal N}_{W, L_j}$ with $j=1 ,
\dotsc, J$, the following holds:
$$
f_{{\Phi}_J} \circ \dotsc \circ f_{{\Phi}_1} \in  {\mathcal N}_{W,L_1 + \dotsc + L_J}.
$$
\end{proposition}
\begin{proof} We first concatenate the networks $\Phi_1$ and $\Phi_2$ as follows to construct a network, say $\Phi_{2 \circ 1}$, that outputs $f_{\Phi_2} \circ f_{\Phi_1}$:
\begin{enumerate}
\item The input and the first $L_1$ hidden layers of $\Phi_{2 \circ 1}$ will be the same as the input and $L_1$ hidden layers of $\Phi_1$. 

\item The ($L_1+1$)-st hidden layer of $\Phi_{2 \circ 1}$ is the first
  hidden layer of $\Phi_{2}$. 

\item The weights between hidden layers $L_1$ and $L_1+1$ (a $W \times
  W$ matrix) will be the output weights of $\Phi_1$ (a $1 \times W$
  vector) multiplied from left by the input weights of the first
  hidden layer of $\Phi_2$ (a $W \times 1$ vector). 

\item The bias of hidden layer $L_1+1$ will be the bias of the first
  hidden layer of $\Phi_2$ (a $W \times 1$ vector) plus the product of the output bias of
  $\Phi_1$ (a scalar) and the input weights of the first hidden layer of $\Phi_2$ (a $W \times 1$ vector). 

\item The remaining hidden layers of $\Phi_{2 \circ 1}$ will be the
  same as the remaining hidden layers of $\Phi_2$. 
\end{enumerate}
The resulting network will have $L_1 + L_2$ hidden layers. This
procedure can be applied $J-1$ times to generate $f_{\Phi_{J \circ \dotsc \circ 1}}$ with $L_1 + \dotsc + L_J$ hidden layers. This completes the proof.
\end{proof}

Now, denote by $T^{\circ k}$ the $k$-fold composition
of a function $T$ with itself
$$
T^{\circ k}(x) = \underbrace{{T \circ T \circ \dotsc \circ T}}_{k-1 \, \, \text{compositions}} (x), \qquad s \ge 1. 
$$
Note that $T^{\circ 1}(x) = T(x)$. Two propositions follow.

\begin{proposition}\label{chap3_prop3}
For any $T \in {\mathcal N}_{W, L}$, then 
$f_{\Phi} (x)= \sum_{i=1}^m a_i \, T^{\circ i}(x) \in \hat{\mathcal N}_{W+2, m L}$.
\end{proposition}
\begin{proof} 
We first generate $T^{\circ m}(x)$ as discussed in the proof of Proposition
\ref{prop:composition}, displayed in blue and red in Figure
\ref{fig:sum_compos}. 
\begin{figure}[h!]
\centering
\includegraphics[width=0.9\linewidth]{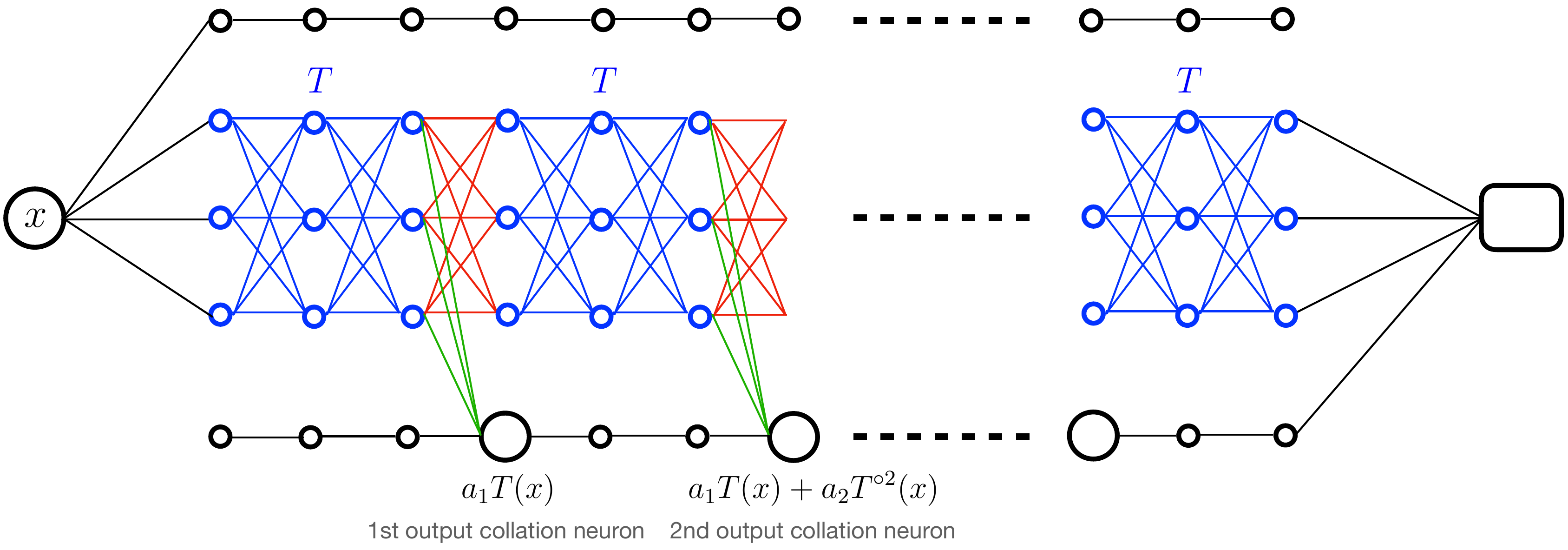}
\caption{Concatenation of $m$ standard ReLU networks, each generating the same function $T$, and addition of a collation channel to output $f_{\Phi} (x)= \sum_{i=1}^m a_i \, T^{\circ i}(x)$.}
\label{fig:sum_compos}
\end{figure} 
We add a collation channel and modify the input
weights (displayed in green) of every $i$-th output collation neuron so that it
produces $a_i \, T^{\circ i}(x)$. The source channel is not needed in
this case. Nevertheless, we include it in case we need to add another
function of $x$ to $\Phi$ later on. The proof is complete.
\end{proof}

\begin{proposition}\label{chap3_prop4}
For any $T \in {\mathcal N}_{W_1, L}$ and $g \in {\mathcal N}_{W_2, L}$, then 
$f_{\Phi_g} (x)= \sum_{i=1}^m a_i \, g \circ T^{\circ i}(x) \in \hat{\mathcal N}_{W_1+W_2+2, (m+1) L}$.
\end{proposition}
\begin{proof} 
The construction of $\Phi_g$ is similar to the construction of $\Phi$
in Proposition \ref{chap3_prop3} with the extra step of adding $m$
copies of $g$ to $T^{\circ m}$ as depicted in Figure \ref{fig:sum_g_compos}. 
\begin{figure}[!h]
\centering
        \includegraphics[width=0.9\linewidth]{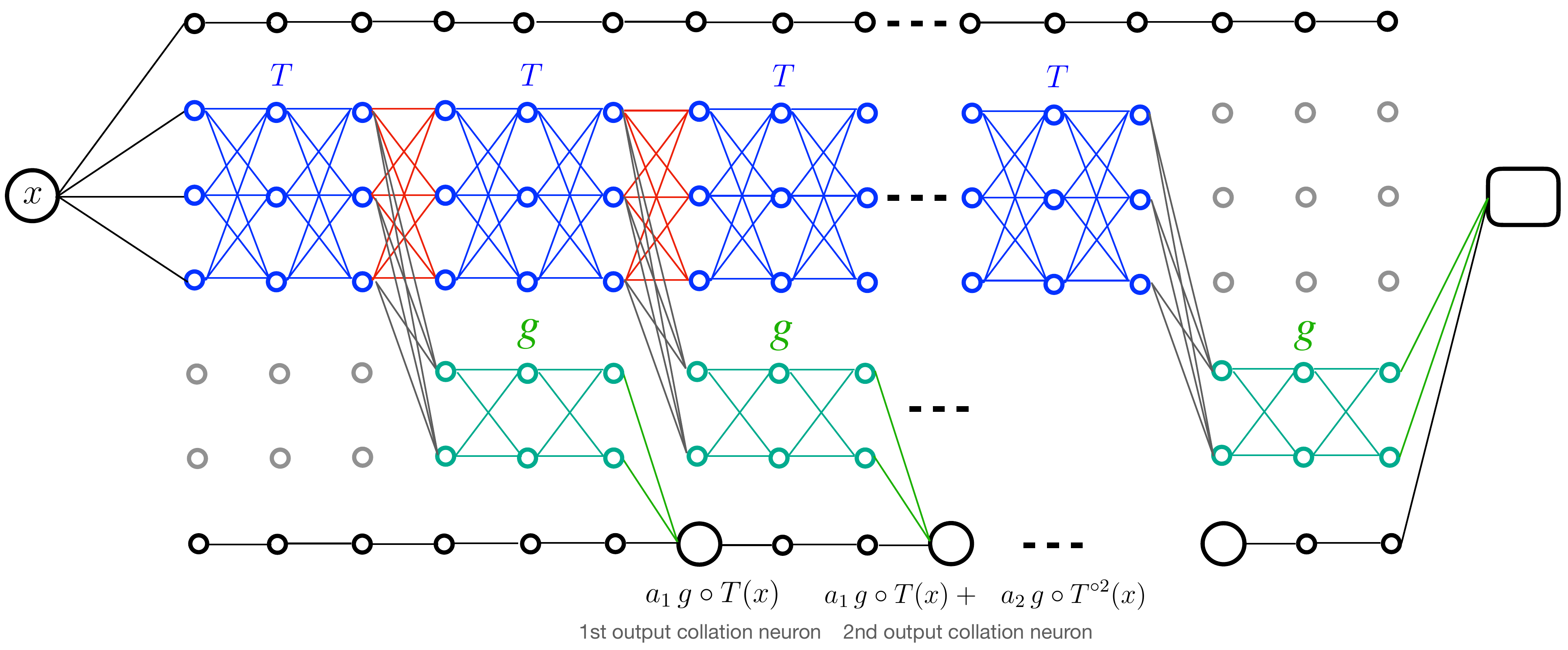}
\caption{Concatenation of $m$ standard ReLU networks, each generating
  the same function $g \circ T$, and addition of a collation channel
  to output $f_{\Phi_g} (x)= \sum_{i=1}^m a_i \, g \circ T^{\circ i}(x)$.}
\label{fig:sum_g_compos}
\end{figure} 
The proof is complete.
\end{proof}

\subsection{Exponential convergence of deep networks in approximating self-similar functions}
\label{sec:depth2}

Consider target functions of the form
$$
f(x) = \sum_{k \ge 1} a^{- k} \, g \circ T^{\circ k}(x), \qquad |a| > 1,
\qquad g: [0,1] \rightarrow {\mathbb R}, \qquad T:[0,1] \rightarrow [0,1].
$$
This is an example of a self-similar function. 
By Proposition \ref{chap3_prop4}, if 
$$
T \in {\mathcal N}_{W_1, L}, \qquad g \in {\mathcal N}_{W_2, L},
$$
then
$$
f_{\Phi_m}(x)= \sum_{k=1}^m a^{- k} \, g \circ T^{\circ k} (x) \ \in \
\hat{\mathcal N}_{W, (m+1) L}, \qquad W=W_1+W_2+2.
$$
We now define the (best) approximation error in approximating $f$ by
special ReLU networks with a fixed width $W$ and depth $(m+1) L$ by
$$
\varepsilon_{(m+1) L} (f) := \inf_{f_{\Phi} \in \hat{\mathcal
    N}_{W, (m+1) L}} || f - f_{\Phi} ||. 
$$
We note that $||\cdot||$ can be any norm as long as $||f||$ and $||f_{\Phi}||$ are bounded. For example, if $f \in C[0,1]$ is continuous, we may consider the uniform norm or the 2-norm.  
Assuming $|| g || = 1$ (this can easily be relaxed), then we will have
\begin{align*}
\varepsilon_{(m+1) L} (f)_{C[0,1]} 
& \le || f - f_{\Phi_m} ||_{C[0,1]} \\
& = || \sum_{k > m} a^{- k} \, g \circ T^{\circ k} ||_{C[0,1]} \\
&\le \sum_{k > m} |a|^{- k} \, || g  \circ T^{\circ k} ||_{C[0,1]} \\
&=  |a|^{-(m+1)} \, \left( 1 + \sum_{k \ge 1} |a|^{- k} \right) 
\le C \, |a|^{-(m+1)}, 
\end{align*}
 where $C = \sum_{k \ge 0}
 |a|^{-k} = 1/(1 - |a|^{-1}) < \infty$ is a bounded constant. Note that the second inequality is
 a simple consequence of triangle inequality. The above estimate
 implies that $f$ can be approximated by a ReLU
 network with exponential accuracy, i.e., the approximation error decays exponentially. Moreover, the deeper the network, that is, the
 more terms $m$ in the sum, the higher the exponential rate of decay. 

Importantly, as we will see in the next section, self-similar functions form a large class of functions, spanning from smooth/analytic functions to functions that are not differentiable anywhere.

\subsection{Takagi functions, an example of self-similar functions}

We now consider a special class of self-similar functions, known as Takagi functions. 
We present two examples of these functions to illustrate that they form a large class of functions that can be well approximated by deep ReLU networks.

\begin{figure}[!h]
\centering
        \includegraphics[width=0.72\linewidth]{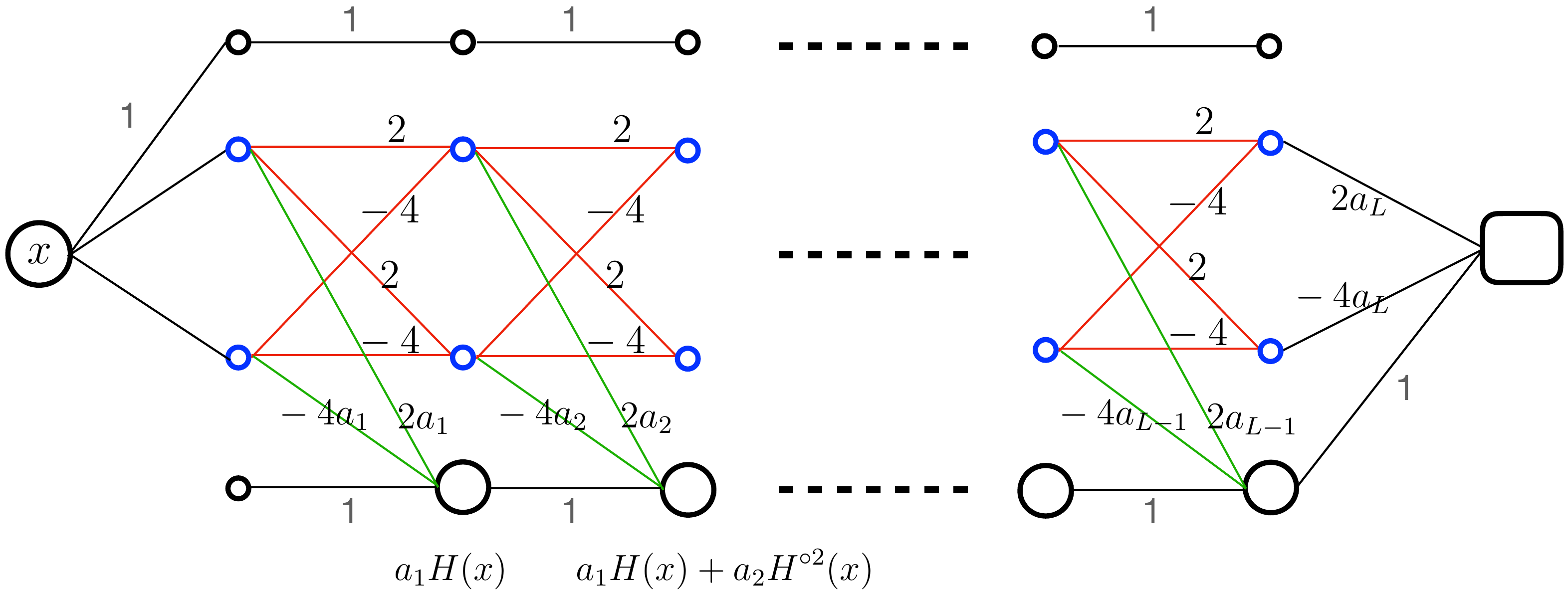}
\vspace{-0.2cm}
\caption{A special ReLU network with width $W=4$ and $L$ hidden layers that outputs $f_{\Phi_L}(x)= \sum_{k=1}^L
  a_k \, H^{\circ k} (x)$, approximating a Takagi function
  $f_{\Phi}(x) = \sum_{ k\ge 1} a_k \, H^{\circ k} (x)$. The deeper the
  network, the higher the rate of decay in approximation error.}
\label{fig:Takagi_network}
\end{figure}

Consider continuous functions of the form
\begin{equation} \label{Takagi_function}
f(x) = \sum_{ k\ge 1} a_k \, H^{\circ k} (x),
\end{equation}
where $\{ a_k \}_{k \ge 1}$ is an absolutely summable sequence of real
numbers, and $H \in {\mathcal N}_{2,1}$ is the hat function. Note that
this is a self-similar function with $g(x) = x$ and $T(x)= H(x)$. By
Proposition \ref{prop:composition}, we will have $H^{\circ k} \in
{\mathcal N}_{2,k}$. 
Following Proposition \ref{chap3_prop3}, a special ReLU network approximating any Takagi function of the form \eqref{Takagi_function} can be easily constructed; see
Figure \ref{fig:Takagi_network}. 

\noindent
The depth $L$ of such special ReLU network, which is equal to the
number of terms in the approximant $f_{\Phi_L}(x)= \sum_{k=1}^L a_k \,
H^{\circ k} (x)$, will determine the exponential decay rate of approximation
error. We will give two specific examples.

\begin{figure}[!h]
\centering
        \includegraphics[width=0.9\linewidth]{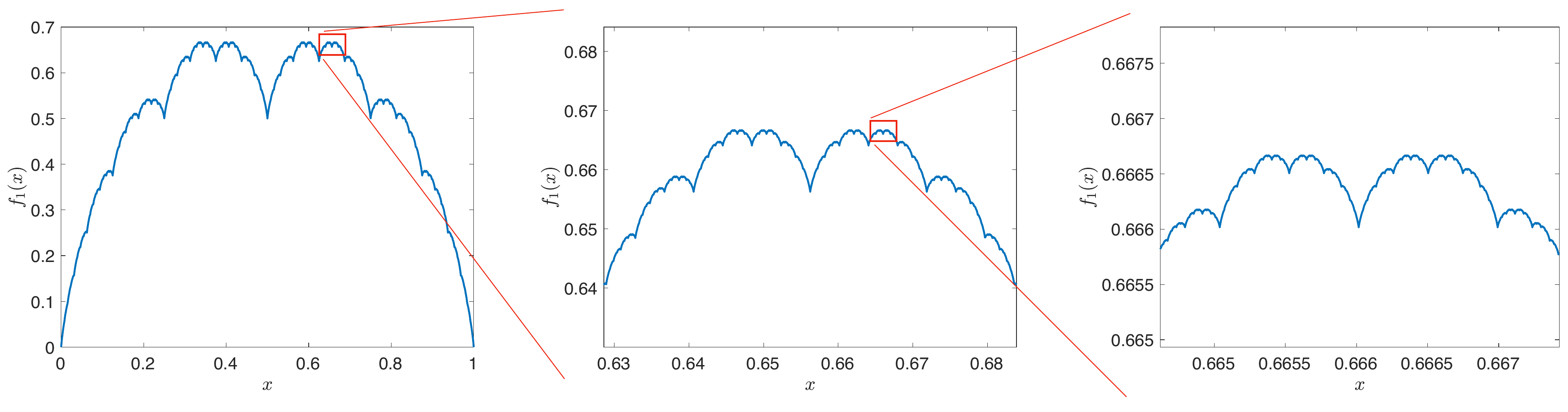}
\caption{The Takagi function $f_1(x) = \sum_{ k\ge 1} 2^{-k} \, H^{\circ k} (x)$ is nowhere differentible and self-similar.}
\label{fig:Takagi_function}
\end{figure}

\medskip
\noindent
{\bf Example 1.} Consider $f_1(x) = \sum_{ k\ge 1} 2^{-k} \, H^{\circ k} (x)$, a Takagi function \eqref{Takagi_function}
with $a_k = 2^{-k}$. 
Figure \ref{fig:Takagi_function} shows the Takagi function with coefficients $a_k = 2^{-k}$. The patterns repeat as we zoom in, which is a characteristic of self-similar functions. 
Let $f_{\Phi_L}(x)= \sum_{k=1}^L 2^{-k} \, H^{\circ k} (x)$ be the ReLU
network approximant of $f_1$, with depth $L \ge 1$. Then, by the error estimate obtained above, we
will have
$$
\varepsilon_{L} (f_1)_{C[0,1]} \le C \, 2^{-L}.  
$$
This implies that $f_1$ can theoretically be approximated with
exponential accuracy by ReLU networks with roughly $W^2 L \sim L$
parameters, given that $W=4$ in the construction above. 
This demonstrates the remarkable power of deep networks, as $f_1$ is nowhere differentiable and thus has very little smoothness in the classical sense. Traditional approximation methods would fail to approximate $f_1$ well. The key to the success of deep networks lies in the self-similarity feature of $f_1$, which possesses a simple compositional/recursive pattern that deep networks can exploit.

\medskip
\noindent
{\bf Example 2.} Consider $f_2(x) = \sum_{ k\ge 1} 4^{-k} \, H^{\circ k} (x) = x (1-x)$, a Takagi function \eqref{Takagi_function} with $a_k = 4^{-k}$. 
Unlike $f_1$ which has very little smoothness, $f_2$ is an analytic
(very smooth) function. 
This representation can be used to show that the quadratic function
$x^2$ can be approximated with exponential accuracy by ReLU networks (as we did earlier). We can
then show that all monomials $x^3, x^4, \dotsc$ can also be approximated with exponential accuracy by ReLU networks. Indeed, we used this strategy in the previous section to prove the density of deep networks, following the density of polynomials in the space of continuous functions. 
In additions, using the additive property of ReLU networks discussed in Section \ref{sec:depth1}, we can conclude that analytic functions (and Sobolev functions) can be approximated by ReLU networks with the same accuracy as their approximation by algebraic polynomials. 

\medskip
\noindent
{\bf A concluding remark.} The above two examples show that the
approximation space of deep ReLU networks is quite large. It contains many functions,
spanning from smooth/analytic functions to functions that lack smoothness in any classical sense. 
This flexibility empowers deep ReLU networks to approximate functions with minimal classical smoothness while maintaining the ability to approximate smooth functions with accuracy comparable to other methods of approximation.

\clearpage

\chapter{Error-complexity analysis for Fourier and ReLU networks}

\bigskip
The study of neural network error-complexity is rapidly evolving, with numerous specialized estimates tailored to specific function spaces. The mathematical techniques employed to derive these estimates are equally diverse. In this chapter, we review a few estimates, emphasizing recent results that apply to a broad class of bounded target functions under minimal regularity assumptions. These results are derived using techniques such as Monte Carlo approximation, optimal control, and an extension of the special ReLU networks introduced earlier.

\section{Target function space}

In this chapter, we focus on error-complexity estimates for neural networks approximating target functions belonging to 
\begin{equation}\label{eqn:approximation_space}
    S=\{f:{\mathbb R}^d\to\mathbb{R}: ||f||_{L^{1}(\mathbb{R}^d)}<\infty, \, ||\hat{f}||_{L^{1}(\mathbb{R}^d)} < \infty, \, f \not\equiv 0 \},
\end{equation}
where $\hat{f}$ is the Fourier transform of $f$. This space consists of all absolutely integrable functions on $\mathbb{R}^d$ with absolutely integrable Fourier transform, excluding the identically zero function. Importantly, functions in $S$ do not need to be continuous; they rather need to be continuous almost everywhere, and there is no requirement of differentiability. We remark here that when $f$ is discontinuous, we consider the relaxed notion of the Fourier transform \cite{grafakos2008classical}, where it is only assumed that
\begin{equation*}
    f(\bm{x}) = \int_{\mathbb{R}^d}\hat{f}(\bm{\omega})e^{i2\pi\bm{\omega} \cdot \bm{x}}\: d\bm{\omega}\qquad \text{ for almost every } \bm{x} \in \mathbb{R}^d.
\end{equation*}

\medskip
\noindent
{\bf Background and overview.} 
Error-complexity estimates in $S$ were initially derived for random Fourier neural networks, a type of neural network that uses randomized complex exponential activation functions. Utilizing Monte Carlo approximation, an estimate for Fourier networks with one hidden layer was derived in \cite{1layerKammonen}. Later, an improved estimate was derived for deep residual Fourier networks \cite{kammonen2023smaller} by leveraging optimal control theory. Subsequently, in \cite{davis2024mathematically}, approximation properties of ReLU networks in the space $S$ were explored, utilizing Fourier networks and a variant of the special ReLU networks introduced in Section \ref{sec:relu-net}, referred to here as generalized special ReLU networks.
Inspired by the works \cite{1layerKammonen,kammonen2023smaller,davis2024mathematically}, we focus the remainder of this chapter on the derivation of error estimates for Fourier networks (in Section \ref{sec:rFNNs}) and ReLU networks (in Section \ref{sec:approximation_errror_ReLU}).

\medskip
\noindent
{\bf Works on regular function spaces.} Over the last decade, there have been several works that derive error-complexity estimates for ReLU neural networks applicable to specific target function spaces, with extra regularity compared to $S$. Recent results include \cite{yarotsky2018optimal,PetVio:18,shen2022optimal,siegel2022optimal,montanelli2019deep}. In \cite{yarotsky2018optimal}, optimal approximation rates are derived for continuous functions that depend on the modulus of continuity of the target function and the complexity of the approximating network. Similar results are derived for piecewise smooth target functions in \cite{PetVio:18} and  H\"{o}lder continuous target functions in \cite{shen2022optimal}. In \cite{montanelli2019deep}, dimension independent approximation rates are derived for bandlimited target functions. Moreover, optimal approximation rates for target functions belonging to Sobolev spaces which depend on the Sobolev norm of the target function and the complexity of the approximating network are derived in \cite{siegel2022optimal, yarotsky2018optimal}. 

\section{Fourier networks}
\label{sec:rFNNs}

In this section, we first introduce Fourier networks (FNs) \cite{1layerKammonen,kammonen2023smaller,Davis_etal:23B} that use complex exponential activation functions, also known as Fourier features: 
$$
s(x) = e^{i \, x}, \qquad x \in {\mathbb R}.
$$
We then study error-complexity estimates for both shallow FNs and deep residual FNs.

\subsection{Structure of Fourier networks}
\label{sec:FN_construction}

\begin{figure}[!htb]
    \centering
    \includegraphics[width = 0.75\textwidth]{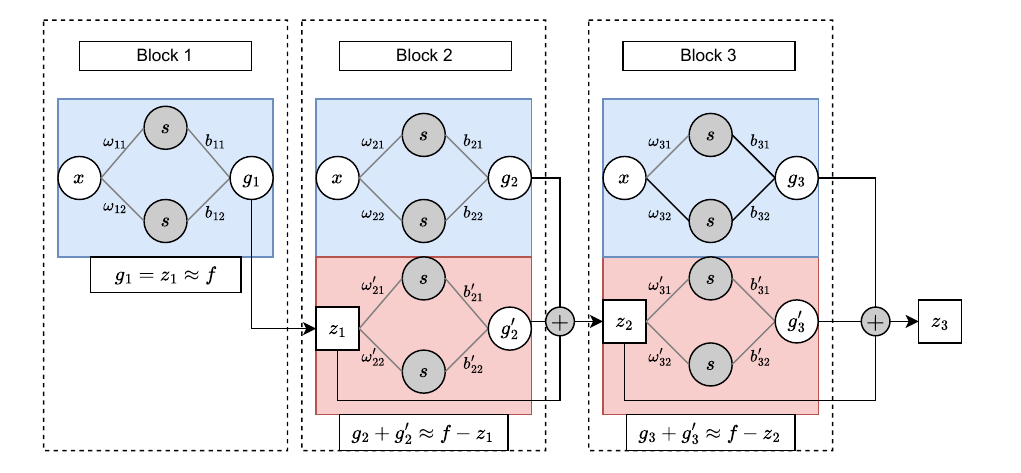}
    \caption{A Fourier neural network with one input and $(W_{F},L_{F}) = (2,3)$.} 
    \label{fig:FF_net_example}
\end{figure}

Following \cite{kammonen2023smaller}, we define a Fourier network of depth $L_{F}\geq 1$ and width $W_{F} \geq 1$ on ${\mathbb R}^d$, with $d$ input neurons, to be composed of $L_{F}$ network blocks, where the first block contains one hidden layer with $W_F$ neurons, and the remaining blocks consist of one hidden layer with $2W_{F}$ neurons. The network realizes the function:
\begin{equation}\label{eq:FN1}
    f_{\Psi}(\bm{x}) =  z_{L_{F}}(\bm{x}), \qquad \bm{x} \in {\mathbb R}^d,
\end{equation}
where $z_{L_{F}}$ results from the recursive scheme:
\begin{align*}
    z_{1}(\bm{x})&=\underbrace{\Re\sum_{j=1}^{W_{F}}b_{1 j}s(\bm{\omega}_{1j} \cdot \bm{x})}_{g_{1}(\bm{x};\: \underline{\bm{\omega}}_{1},\underline{\bm{b}}_{1})};\\
    z_{\ell}(\bm{x}) &= z_{\ell-1}(\bm{x}) +\underbrace{\Re\sum_{j=1}^{W_{F}}b_{\ell j}s(\bm{\omega}_{\ell j} \cdot \bm{x})}_{g_{\ell}(\bm{x};\:\underline{\bm{\omega}}_{\ell},\underline{\bm{b}}_{\ell})}+\underbrace{\Re\sum_{j=1}^{W_{F}}b_{\ell j}'s(\omega_{\ell j}' \, z_{\ell-1}(\bm{x}))}_{g'_{\ell}(z_{\ell-1};\: \underline{\bm{\omega}}'_{\ell},\underline{\bm{b}}'_{\ell})}, \qquad \ell = 2,\dotsc, L_{F}.
\end{align*}
Here, $\bm{\omega}_{\ell j} \in \mathbb{R}^{d}$, $\omega_{\ell j}'\in\mathbb{R}$, and $b_{\ell j},b_{\ell j}'\in\mathbb{C}$ are respectively frequency and amplitude parameters of the network. For each block $\ell$, we denote the collections of these frequency and amplitude parameters by 
$$
    \underline{\bm{\omega}}_{\ell} := \{\bm{\omega}_{\ell j} \}_{j=1}^{W_F}, \qquad  \underline{\bm{b}}_{\ell} := \{b_{\ell j} \}_{j=1}^{W_F}, \qquad \ell = 1, \dotsc, L_F,
$$
$$
    \underline{\bm{\omega}}'_{\ell} := \{\omega_{\ell j}'\}_{j=1}^{W_F}, \qquad 
     \underline{\bm{b}}'_{\ell} := \{b_{\ell j}' \}_{j=1}^{W_F}, \qquad \ell = 2, \dotsc, L_F.
$$
We also denote the collection of these parameters using the notation
$$
    \underline{\bm{\omega}} := \{\underline{\bm{\omega}}_{\ell} \}_{\ell=1}^{L_{F}}, 
    \qquad
     \underline{\bm{\omega}}' := \{\underline{\bm{\omega}}'_{\ell} \}_{\ell=2}^{L_{F}}, 
    \qquad 
    \underline{\bm{b}} :=\{\underline{\bm{b}}_{\ell}\}_{\ell=1}^{L_{F}}, 
     \qquad \underline{\bm{b}}':=\{\underline{\bm{b}}'_{\ell}\}_{\ell=2}^{L_{F}}.
$$
Analogous to the case of ReLU networks, a Fourier network is given by the set of frequencies and amplitudes:
$$
\Psi := \{ (\underline{\bm{\omega}}, \underline{\bm{\omega}}'), \, (\underline{\bm{b}}, \underline{\bm{b}}') \}. 
$$ 
To emphasize the dependence of a Fourier network's output on its frequency and amplitude parameters, we sometimes write:
$$
f_{\Psi}(\bm{x}) = f_{\Psi}(\bm{x}; 
\underline{\bm{\omega}}, \underline{\bm{\omega}}',
\underline{\bm{b}}, \underline{\bm{b}}'), \qquad \bm{x} \in {\mathbb R}^d.
$$
Figure~\ref{fig:FF_net_example} shows an example diagram of a Fourier network taking one-dimensional input with depth $L_F=3$ and width $W_F=2$.

Each block in a Fourier network aims at learning a correction on the output of the previous block. In particular, the output of the first block, $z_{1}$, is intended to be a $W_{F}$ term Fourier-like sum approximation of the target function $f$. Then, at the $\ell$th block, with $\ell \ge 2$, the parameters $\underline{\bm{\omega}}_{\ell}, \underline{\bm{\omega}}_{\ell}', \underline{\bm{b}}_{\ell}, \underline{\bm{b}}_{\ell}'$ are tuned to approximate 
\begin{equation*}
    f(\bm{x}) - z_{\ell-1}(\bm{x})\approx g_{\ell}(\bm{x};\: \underline{\bm{\omega}}_{\ell},\underline{\bm{b}}_{\ell}) + g'_{\ell}(z_{\ell-1};\:\underline{\bm{\omega}}'_{\ell},\underline{\bm{b}}'_{\ell}), \qquad \ell \ge 2.
\end{equation*}
Specifically, we assume that this correction can be efficiently learned as the sum of $g_{\ell}(\bm{x})$, which is just a function of $\bm{x}$, and $g_{\ell}'(z_{\ell -1})$, which is just a function of the output of the previous block $z_{\ell - 1}$.  The output of block $\ell$ is then given by $z_{\ell} = z_{\ell-1} + g_{\ell} + g_{\ell}'$, which is obtained as the sum of the output of the previous block $z_{\ell-1}$ and the approximation of the correction $g_{\ell}+g_{\ell}'$.  Importantly, this makes the output of each block $\ell$ a true approximation of the target function; i.e., $z_{\ell}\approx f$ for all $\ell=1,\dotsc, L_{F}$.

We remark that deep Fourier networks (with $L_F \ge 2$) use a residual network architecture, characterized by the skip-connections that pass along the output from the previous block $z_{\ell-1}$ to the output of the next block $z_{\ell}$ without any computation. Skip-connections similar to this first appeared \cite{he2016deep}. Importantly, residual network architectures have a clear mathematical connection to optimal control problems, and indeed it is this connection that is exploited in \cite{kammonen2023smaller} to derive complexity estimates for deep Fourier networks.

\subsection{Best-approximation error for Fourier networks}

Let $f \in S$ be a target function. Let $f_{\Psi} = f_{\Psi}(\bm{x}; 
\underline{\bm{\omega}}, \underline{\bm{\omega}}',
\underline{\bm{b}}, \underline{\bm{b}}')$ be the function that a Fourier network of depth $L_{F}\geq 1$ and width $W_{F} \geq 1$ with parameters $\underline{\bm{\omega}}, \underline{\bm{\omega}}',
\underline{\bm{b}}, \underline{\bm{b}}'$ realizes on ${\mathbb R}^d$. Consider the risk functional, given by the weighted $L^2$-norm:
$$
\norm{ f(\bm{x}) - f_{\Psi}(\bm{x}; \underline{\bm{\omega}}, \underline{\bm{\omega}}',
\underline{\bm{b}}, \underline{\bm{b}}')}_{L_{\rho}^2({\mathbb R}^d)}^2 := \int_{{\mathbb R}^d}  |f(\bm{x}) - f_{\Psi}(\bf x; \underline{\bm{\omega}}, \underline{\bm{\omega}}',
\underline{\bm{b}}, \underline{\bm{b}}')|^2 \, \rho(\bm{x}) \, d\bm{x},
$$
where $\rho:{\mathbb R}^d \rightarrow [0, \infty)$ is a non-negative, continuous weight function with $\int_{{\mathbb R}^d} \rho(\bm{x}) \, d\bm{x} = 1$. The best-approximation error made by Fourier networks in approximating $f \in S$ is defined by:  
\begin{equation}\label{eqn:opt_error} 
\varepsilon_{\text{opt}} := \min_{\underline{\bm{\omega}},\underline{\bm{\omega}}',\underline{\bm{b}},\underline{\bm{b}}'}
\norm{ f(\bm{x}) - f_{\Psi}(\bm{x};\underline{\bm{\omega}},\underline{\bm{\omega}}',\underline{\bm{b}},\underline{\bm{b}}')}_{L_{\rho}^2({\mathbb R}^d)}^2.
\end{equation}

An interesting approach in obtaining upper bounds for the best-approximation error is to assume that the frequency parameters $\underline{\bm{\omega}}_{\ell}, \underline{\bm{\omega}}'_{\ell}$ at each block $\ell=1,\dotsc, L_{F}$ are independently and identically distributed (i.i.d.) random variables following two specific distributions $p_{\ell}(\bm{\omega}): {\mathbb R}^d \rightarrow [0, \infty)$ and $q_{\ell}(\omega'): {\mathbb R} \rightarrow [0, \infty)$, respectively. 
That is,
\begin{equation}\label{eqn:random_freq}
\bm{\omega}_{\ell j} \overset{\mathrm{iid}}{\sim}  p_{\ell}(\bm{\omega}), 
\qquad
\omega_{\ell j}' \overset{\mathrm{iid}}{\sim}  q_{\ell}(\omega'), \qquad j=1, \dotsc, W_{F},
\end{equation}
where $\bm{\omega}$ and $\omega'$ are arbitrary input variables to the density functions $p_{\ell}$ and $q_{\ell}$ respectively.
Denoting by ${\mathbb E}_{\underline{\bm{\omega}},\underline{\bm{\omega}}'}[\cdot]$ the expectation with respect to frequency parameters, we define the mean squared error with respect to frequency parameters by:
\begin{equation}\label{eqn:MSE_error}
    \varepsilon := \mathbb{E}_{\underline{\bm{\omega}},\underline{\bm{\omega}}'}[\min_{\underline{\bm{b}},\underline{\bm{b}}'}\norm{ f(\bm{x}) - f_{\Psi}(\bm{x};\underline{\bm{\omega}},\underline{\bm{\omega}}',\underline{\bm{b}},\underline{\bm{b}}')}_{L_{\rho}^2({\mathbb R}^d)}^2].
\end{equation}
Since a minimum is always less than or equal to its mean, there holds
$$
\varepsilon_{\text{opt}} \le \varepsilon.
$$
Finding an upper bound for the mean squared error \eqref{eqn:MSE_error} will hence also give an upper bound for the best-apprixmation error \eqref{eqn:opt_error}. 

The optimization problem \eqref{eqn:MSE_error} is a variant of the well known \emph{random Fourier features problem}, which first appeared in \cite{rahimi2007random} and later in \cite{weinan2020comparative,rudi2017generalization, 1layerKammonen, kammonen2023smaller}.

\subsubsection{Shallow Fourier networks} 
We first derive error estimates for shallow Fourier networks, with $L_F = 1$. This was achieved in \cite{1layerKammonen} using a Monte Carlo sampling argument, a result summarized in the following theorem.
\begin{theorem}\label{thm:1hidlayer_rFNN}
Let $f$ be a target function in $S$ as defined in \eqref{eqn:approximation_space}. The mean squared error \eqref{eqn:MSE_error} for a Fourier network of depth $L_F =1$ satisfies
\begin{equation}\label{eqn:1hidlayer_estimate}
    \varepsilon \leq C \, \frac{||\hat{f}||^{2}_{L^{1}(\mathbb{R}^d)}}{W_{F}},
\end{equation}
where $C$ is a positive constant, and $\hat{f}$ is the Fourier transform of $f$.
\end{theorem}
\begin{proof}
    Following \cite{1layerKammonen}, we introduce the Fourier representation of the target function and an associated Monte Carlo estimator $g(\bm{x},\underline{\bm{\omega}}_{1})$
    \begin{equation}
        f(\bm{x}) = (2\pi)^{-d/2}\int_{\mathbb{R}^d}\hat{f}(\bm{\omega})e^{i\bm{\omega}\cdot \bm{x}}d\bm{\omega},\qquad g(\bm{x},\underline{\bm{\omega}}_{1}) =\frac{1}{W_{F}}\sum_{j=1}^{W_{F}}\frac{\hat{f}(\bm{\omega}_{1j})e^{i\bm{\omega}_{1j}\cdot \bm{x}}}{(2\pi)^{d/2}p_{1}(\bm{\omega}_{1j})}, 
    \end{equation}
where $\underline{\bm{\omega}}_{1} = \{ \bm{\omega}_{1j} \}_{j=1}^{W_F}$ is a collection of $W_F$ independent samples drawn from the distribution $p_1(\bm{\omega}): {\mathbb R}^d \rightarrow [0, \infty)$. We notice that $g$ is an unbiased estimator of $f$:
\begin{equation}\label{eq:unbiased_est}
\mathbb{E}_{\underline{\bm{\omega}}_{1}}[g(\bm{x},\underline{\bm{\omega}}_{1})] = f(\bm{x}).
\end{equation}
Importantly, $g$ is also of the same structure as a shallow Fourier network with the particular amplitudes $\underline{\bm{\beta}} = (\beta_{1},\dotsc,\beta_{W_{F}})$ given by
    \begin{equation}
        \beta_{j} = \frac{\hat{f}(\bm{\omega}_{1j})}{W_{F}(2\pi)^{d/2}p_{1}(\bm{\omega}_{1j})}, \qquad j=1,\dotsc, W_F.
    \end{equation}
Hence, in order to study the approximation capability of a shallow Fourier network in an upper bound sense, we can study the particular version which is the Monte Carlo estimator $g$. Using the definition of variance of a Monte Carlo estimator and \eqref{eq:unbiased_est}, we directly calculate
\begin{equation}\label{eq:var_est}
\mathbb{V}_{\underline{\bm{\omega}}_{1}}[g(\bm{x}, \underline{\bm{\omega}}_{1})] =
        \mathbb{E}_{\underline{\bm{\omega}}_{1}}
        [ | g(\bm{x}, \underline{\bm{\omega}}_{1}) - f(\bm{x}) |^2]
        =
        \frac{1}{W_{F}}\mathbb{E}_{\bm{\omega}}\left[\frac{|\hat{f}(\bm{\omega})|^2}{(2\pi)^dp_{1}^2(\bm{\omega})}-f^2(\bm{x})\right].
\end{equation}
We can now write 
    \begin{align*}
        \varepsilon &= \mathbb{E}_{\underline{\bm{\omega}}_{1}}[\min_{\underline{\bm{b}}_{1}}
        \norm{f(\bm{x})-f_{\Psi}(\bm{x}; \underline{\bm{\omega}}_{1},\underline{\bm{b}}_{1})}_{L_{\rho}^2({\mathbb R}^d)}^2]\leq \mathbb{E}_{\underline{\bm{\omega}}_{1}}[ \norm{f(\bm{x})-g(\bm{x},\underline{\bm{\omega}}_{1})}_{L_{\rho}^2({\mathbb R}^d)}^2]\\
        &=  \norm{ \bigl( \mathbb{E}_{\underline{\bm{\omega}}_{1}}
        [ | g(\bm{x}, \underline{\bm{\omega}}_{1}) - f(\bm{x}) |^2] \bigr)^{1/2}}_{L_{\rho}^2({\mathbb R}^d)}^2
        =\frac{1}{W_{F}} 
        \norm{ \bigl( \mathbb{E}_{\bm{\omega}}
        \bigl[\frac{|\hat{f}(\bm{\omega})|^2}{(2\pi)^dp_{1}^2(\bm{\omega})}-f^2(\bm{x})\bigr]
        \bigr)^{1/2}}_{L_{\rho}^2({\mathbb R}^d)}^2
        \\
        &\le \frac{1}{W_{F}} 
        \norm{ \bigl( \mathbb{E}_{\bm{\omega}}
        \bigl[\frac{|\hat{f}(\bm{\omega})|^2}{(2\pi)^dp_{1}^2(\bm{\omega})}\bigr]
        \bigr)^{1/2}}_{L_{\rho}^2({\mathbb R}^d)}^2
        = 
        \frac{1}{W_{F}}\mathbb{E}_{\bm{\omega}}\bigl[\frac{|\hat{f}(\bm{\omega})|^2}{(2\pi)^{d} \, p_{1}^2(\bm{\omega})}\bigr].
    \end{align*}
Here, the first inequality holds because the minimum is less than or equal to any specific value (e.g., at $\underline{\bm{b}}_1 = \underline{\bm{\beta}}$). The first equality follows by exchanging the order of integration in the expectation and the $L^2$-norm. The second equality comes from equation \eqref{eq:var_est}, and the last equality holds since the term inside the norm is independent of $\bm{x}$. 
We next search for the distribution $p_{1}^{*}$ that minimizes the derived upper bound. This involves solving the following optimization problem
$$
    \min_{p_{1}}\bigg\{(2\pi)^{-d}\int_{\mathbb{R}^d}\frac{|\hat{f}(\bm{\omega})|^2}{p_{1}(\bm{\omega})}\:d\bm{\omega}; \int_{\mathbb{R}^d}p_{1}(\bm{\omega})\:d\bm{\omega} = 1 \bigg\}.
$$
In order to solve this problem, we introduce the change of variables $p_{1}(\bm{\omega}) = \bar{p}_{1}(\bm{\omega})/\int_{\mathbb{R}^d}\bar{p}_{1}(\bm{\omega})d\bm{\omega}$, which guarantees $\int_{\mathbb{R}^d}p_{1}(\bm{\omega})d\bm{\omega}=1$ for any $\bar{p}_{1}:\mathbb{R}^d\rightarrow [0,\infty)$. We then aim to optimize over $\bar{p}_{1}$. For any small $\delta>0$ and any $v:\mathbb{R}^d\rightarrow \mathbb{R}$, we introduce the function $Q(\delta)$ defined by 
\begin{equation}
        Q(\delta) = \int_{\mathbb{R}^d}\frac{|\hat{f}(\bm{\omega})|^2}{\bar{p}_{1}(\bm{\omega}) + \delta \,  v(\bm{\omega})}\:d\bm{\omega}\int_{\mathbb{R}^d}\bar{p}_{1}(\bm{\omega}) + \delta \, v(\bm{\omega})\:d\bm{\omega}
\end{equation}
Subsequently, we calculate
$$
        \frac{dQ}{d\delta}(0) =\int_{\mathbb{R}^d}\bigl( c_2-c_1\frac{|\hat{f}(\bm{\omega})|^2}{\bar{p}_{1}^2(\bm{\omega})}\bigr)v(\bm{\omega})\:d\bm{\omega}, \qquad
        c_1 = \int_{\mathbb{R}^d}\bar{p}_{1}(\tilde{\bm{\omega}})\:d\tilde{\bm{\omega}},\qquad c_2=\int_{\mathbb{R}^d}\frac{|\hat{f}(\tilde{\bm{\omega}})|^2}{\bar{p}_{1}(\tilde{\bm{\omega}})}\:d\tilde{\bm{\omega}}.
$$
The optimality condition $Q'(0)=0$ implies $\bar{p}_{1}(\bm{\omega}) = \left(\frac{c_1}{c_2}\right)^{1/2}|\hat{f}(\bm{\omega})|$, which implies the minimizing density is given by 
\begin{equation}
        p_{1}^{*}(\bm{\omega}) = |\hat{f}(\bm{\omega})|/||\hat{f}||_{L^{1}(\mathbb{R}^d)}.
\end{equation}
Replacing $p_{1}$ by $p_{1}^*$ in the derived upper bound on $\varepsilon$ yields the desired result. 
\end{proof}

The result in Theorem~\ref{thm:1hidlayer_rFNN} leads directly to the following corollary. 
\begin{corollary}
Let $f$ be a target function in $S$ as defined in \eqref{eqn:approximation_space}. The best-approximation error \eqref{eqn:opt_error} for a Fourier network of depth $L_F =1$ satisfies
\begin{equation}\label{eqn:shallow_opt}
    \varepsilon_{\text{opt}} \leq C \, \frac{||\hat{f}||^{2}_{L^{1}(\mathbb{R}^d)}}{W_{F}},
\end{equation}
where $C$ is a positive constant, and $\hat{f}$ is the Fourier transform of $f$.    
\end{corollary}

This result establishes a clear connection between network approximation error and complexity. Indeed, for shallow Fourier networks with one hidden layer, the approximation error can be bounded above by a quantity inversely proportional to the network width and proportional to the $L^1$-norm of the Fourier transform of the target function. A similar result was derived for one hidden layer networks with sigmoid activation \cite{Barron:1993}, but only applying to target functions in Barron space (i.e. having Fourier transform with bounded first moment).

\subsubsection{Deep Fourier networks} 
Given the error-complexity estimates \eqref{eqn:1hidlayer_estimate} and \eqref{eqn:shallow_opt} for shallow Fourier networks with one hidden layer $L_F=1$, a natural follow-up question is whether this estimate can be improved if we consider deep neural networks with many hidden layers $L_F \ge 2$. Here, we will show that the answer to this question is yes, a result stated in the following theorem. 
\begin{theorem}\label{thm:ff_main_result} 
Let $f:{\mathbb R}^d \rightarrow {\mathbb R}$ be a function in $S$ as defined in \eqref{eqn:approximation_space}. Let $f_{\Psi}$ be a residual Fourier network \eqref{eq:FN1} with depth $L_{F}\ge 2$ and width $W_{F} \ge 1$ and parameters $(\underline{\bm{\omega}},\underline{\bm{\omega}}',\underline{\bm{b}},\underline{\bm{b}}')$. 
There exists a positive constant $C$ such that for all sufficiently large $W_{F} L_{F}$, with $W_{F}=\mathcal{O}(L_{F})^2$, the best-approximation error \eqref{eqn:opt_error} satisfies 
\begin{equation}\label{eqn:deep_opt}
\varepsilon_{\text{opt}} \leq C\frac{||f||_{L^{\infty}({\mathbb R}^d)}^{2}}{W_{F}L_{F}}\left(1 + \ln \frac{||\hat{f}||_{L^{1}(\mathbb{R}^d)}}{||f||_{L^{\infty}({\mathbb R}^d)}} \right)^{2}.
\end{equation}
\end{theorem}

The proof of Theorem \ref{thm:ff_main_result} uses Theorem 2.1 in \cite{kammonen2023smaller}, that we state here as a lemma.

\begin{lemma}\label{thm:Anders} 
Let $f:{\mathbb R}^d \rightarrow {\mathbb R}$ be a function in $S$ as defined in \eqref{eqn:approximation_space}. 
Let $f_{\Psi}$ be a residual Fourier network \eqref{eq:FN1} of depth $L_{F}\ge 2$ and width $W_{F} \ge 1$, with random frequencies $\bm{\omega}_{\ell k} \in \mathbb{R}^d$ and $\omega_{\ell k}' \in \mathbb{R}$ independently and identically distributed according to the probability density functions $p: \bm{\omega} \in \mathbb{R}^d\to [0,\infty)$ and $p': \omega' \in \mathbb{R}\to [0,\infty)$, respectively. 
Additionally, let $h: z \in {\mathbb R} \rightarrow {\mathbb R}$ be a Schwartz function, and assume that the following quantities are bounded:
\begin{align*}
Q_1 &:= \bigg|\bigg| \frac{|\hat{f}(\bm{\omega})|^2}{p(\bm{\omega})} \bigg|\bigg|_{L^1({\mathbb R}^d)} + \bigg|\bigg| \frac{\hat{f}(\bm{\omega})}{p(\bm{\omega})} \bigg|\bigg|_{L^{\infty}({\mathbb R}^d)} < \infty,\\
Q_2 &:= \bigg|\bigg| \frac{| F \partial_{\omega'} \hat{h}(F \omega')|^2}{p'(\omega')} \bigg|\bigg|_{L^1({\mathbb R})} 
+ \bigg|\bigg| \frac{| F \partial_{\omega'} \hat{h}(F \omega')|^2}{p'(\omega')} \bigg|\bigg|_{L^{\infty}({\mathbb R})} 
+ F || \omega' \partial_{\omega'} \hat{h}(F \omega') ||_{L^2({\mathbb R})} < \infty,\\
Q_3 &:= || \partial_z (z h(z/F))  ||_{L^{\infty}({\mathbb R})} + F \, || \partial_z^2 (z h(z/F))  ||_{L^{\infty}({\mathbb R})} < \infty,
\end{align*}
where $F := || f||_{L^{\infty}(\mathbb{R}^d)} \in (0,\infty)$. 
Then, there are positive constants $C'$ and $c$ such that the mean squared error \eqref{eqn:MSE_error} satisfies:
\begin{equation}\label{eqn:thm2.1b}
\varepsilon = \mathbb{E}_{\underline{\bm{\omega}},\underline{\bm{\omega}}'}\left[\min_{\underline{\bm{b}},\underline{\bm{b}}'}
\norm{f(\bm{x}) - f_{\Psi}(\bm{x}; \underline{\bm{\omega}},\underline{\bm{\omega}}',\underline{\bm{b}},\underline{\bm{b}}')}_{L_{\rho}^2({\mathbb R}^d)}^2\right]\leq \frac{C'}{W_F L_F} + \mathcal{O}\bigl(\frac{1}{W_F^{2}} + \frac{1}{L_F^{4}} + L_F \, e^{-c \, W_F}\bigr).
\end{equation}
Furthermore, the minimum value of the constant $C'$ is obtained for the optimal densities
\begin{equation}\label{eqn:optimal_densities}
p(\bm{\omega}) = \frac{|\hat{f}(\bm{\omega})|}{|| \hat{f}(\bm{\omega}) ||_{L^1({\mathbb R}^d)}}, \qquad p'(\omega') = \frac{| \partial_{\omega'} \hat{h}(F \omega')|}{|| \partial_{\omega'} \hat{h}(F \omega') ||_{L^1({\mathbb R})}},  
\end{equation}
and given by
\begin{equation}\label{eqn:const1}
C' = B^{2}(1 + \ln(A/B))^2,
\end{equation}
where
\begin{equation}\label{eqn:const2}
A=||\hat{f}||_{L^{1}(\mathbb{R}^d)}, 
\qquad 
B=F \, ||\partial_{\omega'}\hat{h}(F\omega')||_{L^{1}(\mathbb{R})} \, \exp\left(||\partial_{z}(zh(z/F))||_{L^{\infty}(\mathbb{R})}\right).
\end{equation}
\end{lemma}

Lemma \ref{thm:Anders} is proven in an elegant way in \cite{kammonen2023smaller} by initially framing the Fourier network optimization problem as a discrete approximation of a continuous-time optimal control problem. This optimal control problem, which represents a network with infinite width and depth, has a known explicit solution. By introducing random frequencies and random times, the integrals appearing in the solution of the optimal control problem are approximated using Monte Carlo importance sampling. Interestingly, the Monte Carlo estimates of the optimal control solution take the form of Fourier networks with density-dependent amplitudes. The error estimate \eqref{eqn:thm2.1b} is then derived by analyzing the errors associated with the Monte Carlo approximation. We encourage interested readers to refer to \cite{kammonen2023smaller} for more details.

\begin{remark} 
Lemma \ref{thm:Anders} presented here is indeed a special case of Theorem 2.1 in \cite{kammonen2023smaller}. Specifically, here we only consider the scenario of noiseless data ($\varepsilon = 0$) and Fourier networks that do not include the regularization parameter ($\delta = 0$). 
Moreover, Theorem 2.1 in \cite{kammonen2023smaller} originally considers the expectation of the error with respect to time in the left hand side of \eqref{eqn:thm2.1b}, highlighting that the proof relies on the continuous-time optimal control problem with random time points. However, since the random Fourier network $f_{\Psi}$ in the left hand side of \eqref{eqn:thm2.1b} is independent of random time---due to the independence of random frequencies and random times---the expectation in time simplifies to the identity map. We note that in Theorem 2.1 in \cite{kammonen2023smaller}, the joint distribution of frequency and time variables factors into the product of the marginal distributions: $\bar{p}(\bm{\omega},t) = p(\bm{\omega}) \, q(t)$, and $\bar{p}'(\omega',t') = p'(\omega') \, q'(t')$,
where $t$ and $t'$ are the random times, supported on $[0,1]$, corresponding to random frequencies $\bm{\omega}$ and $\omega'$, respectively.  
Finally, it is worth noting that the strict positivity of $F$, which is required to obtain \eqref{eqn:thm2.1b}, is mistakenly omitted in Theorem 2.1 of \cite{kammonen2023smaller}, but is correctly included in Lemma \ref{thm:Anders} here. This omission is a typographical error and does not affect the validity of the result.
\end{remark}

\medskip
\noindent
{\bf Proof of Theorem \ref{thm:ff_main_result}.} 
We let $h:\mathbb{R}\to \mathbb{R}$ be a Schwartz function. Then, with $f \in S$ and the optimal densities \eqref{eqn:optimal_densities}, we first show that all assumptions of Lemma \ref{thm:Anders} hold so that \eqref{eqn:thm2.1b} is valid with the optimal constant $C'$ in \eqref{eqn:const1}-\eqref{eqn:const2}. To this end, with the optimal density $p$ in \eqref{eqn:optimal_densities}, we obtain
$$
Q_1 = || \hat{f}(\bm{\omega}) ||_{L^1({\mathbb R}^d)}^2 + || \hat{f}(\bm{\omega}) ||_{L^1({\mathbb R}^d)},
$$
which is bounded since $f \in S$. Next, with $p'$ in \eqref{eqn:optimal_densities}, we obtain
$$
Q_2 = F^2 || \partial_{\omega'} \hat{h}(F \omega') ||_{L^1({\mathbb R})}^2 +  F^2 || \partial_{\omega'} \hat{h}(F \omega') ||_{L^1({\mathbb R})} \, || \partial_{\omega'} \hat{h}(F \omega') ||_{L^{\infty}({\mathbb R})} + F || \omega' \partial_{\omega'} \hat{h}(F \omega') ||_{L^2({\mathbb R})}.
$$
We note that Schwartz functions are smooth functions whose derivatives (including the function itself) decay at infinity faster than any power. We further note that the Fourier transform of a Schwartz function is also a Schwartz function. This implies that 
\begin{equation*}
|| \partial_{\omega'} \hat{h}(F \omega') ||_{L^1({\mathbb R})} = \int_{\mathbb R} | \partial_{\omega'} \hat{h}(F \omega') | \, d\omega' = \int_{\mathbb R} | \partial_u \hat{h}(u) | \, du < \infty,
\end{equation*}
independent of $F$. Also noting that $F<\infty$, which follows from $f \in S$, it further implies 
$$
|| \partial_{\omega'} \hat{h}(F \omega') ||_{L^{\infty}({\mathbb R})} = F \, || \partial_u \hat{h}(u) ||_{L^{\infty}({\mathbb R})} < \infty, 
$$
and
$$
F^2 \, || \omega' \partial_{\omega'} \hat{h}(F \omega') ||_{L^2({\mathbb R})}^2 = F^2 \, \int_{\mathbb R} \omega'^2 \, (\partial_{\omega'} \hat{h}(F \omega'))^2 \, d\omega' = F \, \int_{\mathbb R} u^ 2 \, (\partial_u \hat{h}(u))^2 \, du < \infty.
$$
Hence, $Q_2$ is bounded. We next show that $Q_3$ is bounded. The first term in $Q_3$ can be shown to be bounded independent of $F$:
\begin{multline*}
||\partial_{z}(zh(z/F))||_{L^{\infty}(\mathbb{R})} = ||h(z/F) + z \, \partial_z h(z/F) ||_{L^{\infty}(\mathbb{R})} =\\
= ||h(u) + u \, \partial_u h(u) ||_{L^{\infty}(\mathbb{R})} 
\leq ||h(u) ||_{L^{\infty}(\mathbb{R})} + || u \, \partial_u h(u) ||_{L^{\infty}(\mathbb{R})} < \infty,
\end{multline*}
knowing that $h$ is a Scwhartz function. 
The second term in $Q_3$ can similarly be shown to be bounded:
\begin{multline*}
F \, ||\partial_{z}^2 (zh(z/F))||_{L^{\infty}(\mathbb{R})} = F \, || 2 \, \partial_z h(z/F) + z \, \partial_z^2 h(z/F)||_{L^{\infty}(\mathbb{R})} = \\
= || 2 \, \partial_u h(u) + u \, \partial_u^2 h(u)||_{L^{\infty}(\mathbb{R})} \le 
2 \, || \partial_u h(u) ||_{L^{\infty}(\mathbb{R})} + || u \, \partial_u^2 h(u)||_{L^{\infty}(\mathbb{R})} < \infty.
\end{multline*}
Since all assumptions of Lemma \ref{thm:Anders} hold, the estimate \eqref{eqn:thm2.1b} is valid with the optimal constant $C'$ in \eqref{eqn:const1}-\eqref{eqn:const2}. Moreover, we have already shown that both $||\partial_{\omega'}\hat{h}(F \, \omega')||_{L^{1}(\mathbb{R})}$ and $||\partial_{z}(zh(z/F))||_{L^{\infty}(\mathbb{R})}$ are bounded independent of $F$. This implies that there is a positive constant $C$ such that 
$$
C' = C \, F^2 \, ( 1 + \ln (|| \hat{f} ||_{L^{1}(\mathbb{R^d})} / F) )^2.
$$
From \eqref{eqn:thm2.1b}, we hence obtain
\begin{equation*}
\varepsilon
\leq 
C \frac{F^2}{W_F L_F} \bigl( 1 + \ln \frac{|| \hat{f} ||_{L^{1}(\mathbb{R^d})}}{F}  \bigr)^2 + \mathcal{O}\bigl(\frac{1}{W_F^{2}} + \frac{1}{L_F^{4}} + L_F \, e^{-c \, W_F} \bigr).
\end{equation*}
Now, assuming $W_F L_F\to \infty$ with $W_F = \mathcal{O}(L_F^2)$, the first term in the right-hand-side of the above inequality dominates the second term, and we conclude that there exists a positive constant $C$ such that
\begin{equation*}
\varepsilon
\leq C \frac{F^2}{W_F L_F} 
    \bigl( 1 + \ln \frac{||\hat{f}||_{L^{1}(\mathbb{R}^d)}}{F} \bigr)^{2}.
\end{equation*}
The desired estimate \eqref{eqn:deep_opt} follows noting that a minimum is always less than or equal to its mean. This completes the proof. \qed

\begin{figure}[!h]
    \centering
    \includegraphics[width=0.47\linewidth]{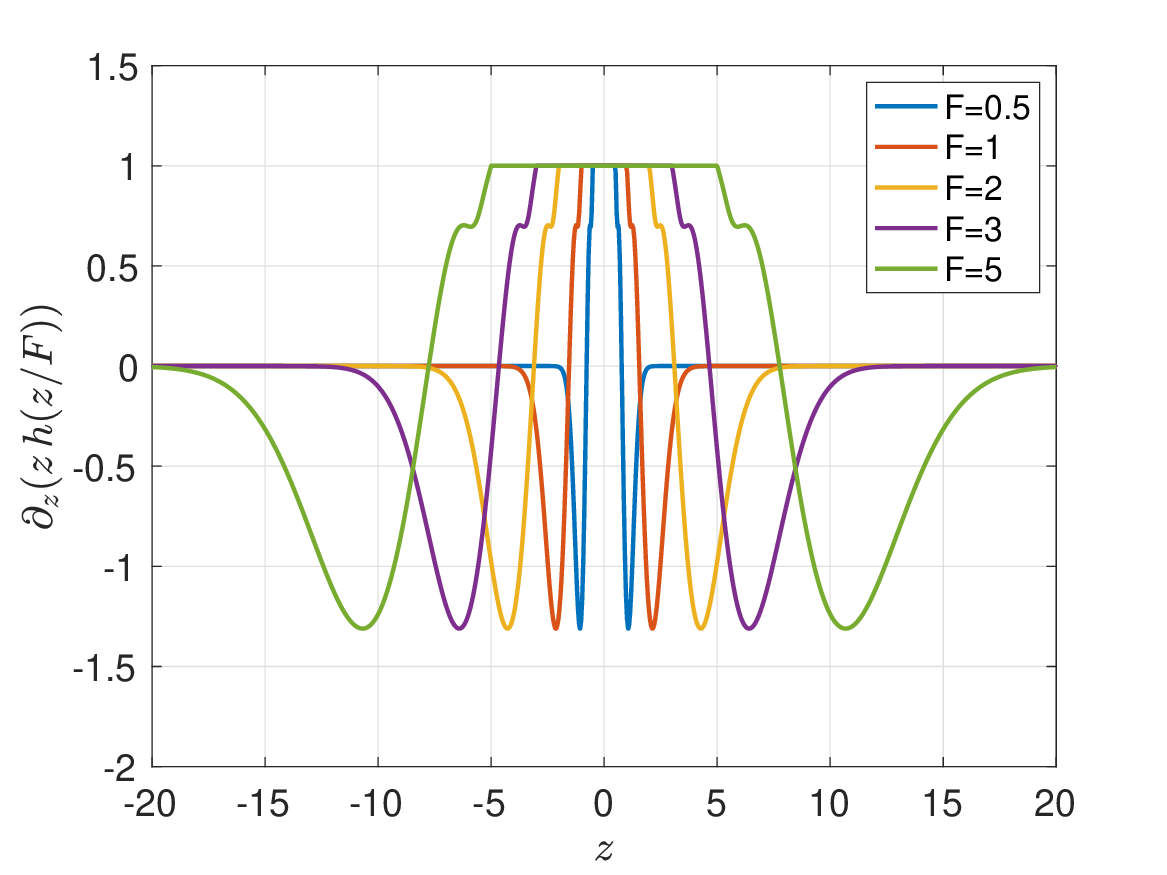} 
    \includegraphics[width=0.47\linewidth]{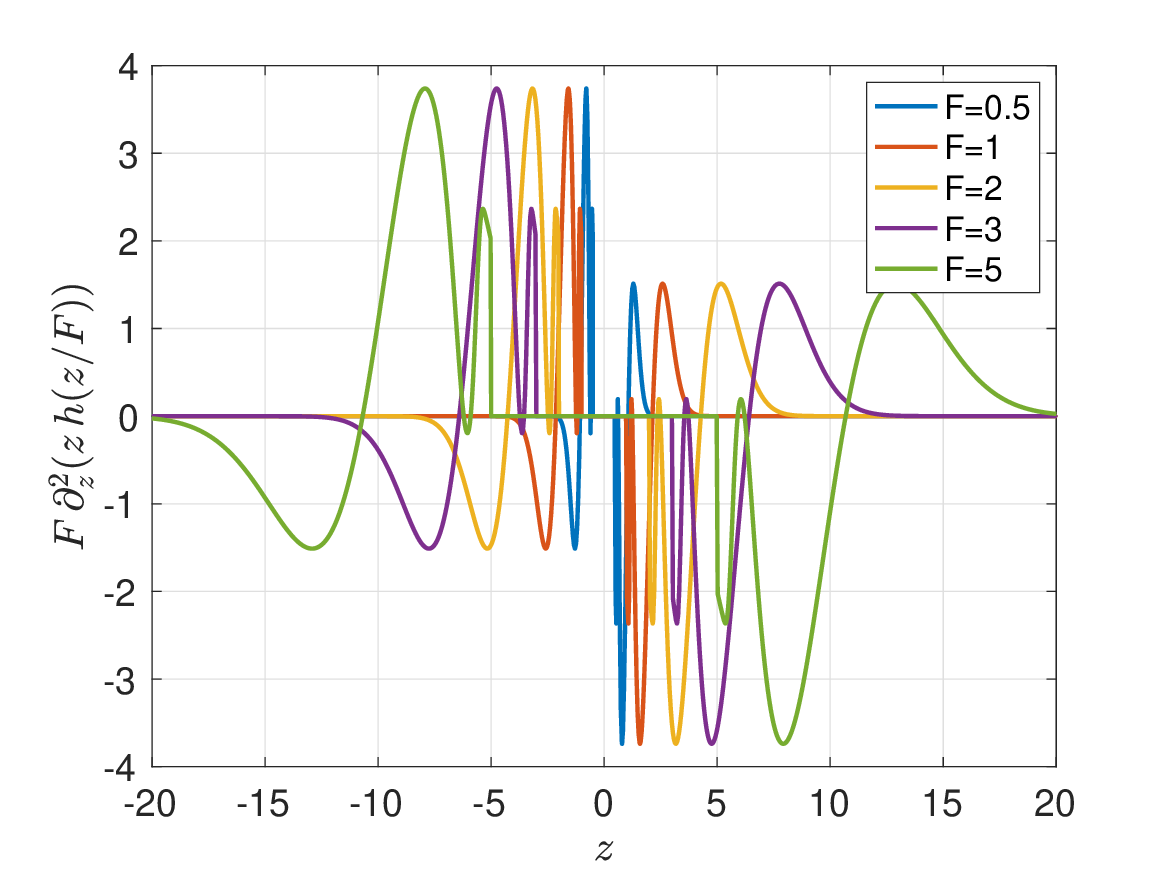} 
    \caption{Profiles of $\partial_z(zh(z/F))$ and $F \, \partial_{z}^2 (zh(z/F))$ for different values of $F$ for the specific Schwartz function defined in \eqref{eqn:specific_h}, illustrating their boundedness independent of $F$.}
    \label{fig:indep_f}
\end{figure}

To further illustrate the boundedness of derivatives of $z h(z/F)$, which was shown in the proof of Theorem \ref{thm:ff_main_result}, in Figure \ref{fig:indep_f} we plot the profiles of $\partial_z(zh(z/F)$ (left) and $F \, \partial_{z}^2 (zh(z/F))$ (right) for a specific Schwartz function defined as in \cite{kammonen2023smaller}:
\begin{equation}\label{eqn:specific_h}
    h(z) = \begin{cases}
                \left(1 + e^{\frac{1}{1-z}}\right)e^{-(1-z)^{2}} & z > 1\\
                1                                                & |z|\leq 1\\
                \left(1 + e^{\frac{1}{1+z}}\right)e^{-(1+z)^{2}} & z < -1,
            \end{cases}
\end{equation}
for different values of $F$. 

The estimate \eqref{eqn:deep_opt} indicates linear convergence in the best-approximation error with respect to the product $W_{F} L_{F}$ of network width and depth. It further shows that the best-approximation error can be bounded above by a quantity proportional to $||f||_{L^{\infty}({\mathbb R}^d)}^2 \, (1+\ln(||\hat{f}||_{L^{1}(\mathbb{R}^d)}/||f||_{L^{\infty}({\mathbb R}^d)}))^2$. The term $(1+\ln(||\hat{f}||_{L^{1}(\mathbb{R}^d)}/||f||_{L^{\infty}({\mathbb R}^d)}))^2$ is related to the regularity of the target function $f$. The ratio $||\hat{f}||_{L^{1}(\mathbb{R}^d)}/||f||_{L^{\infty}({\mathbb R}^d)}$ is always greater than or equal to $1$ following from H{\"o}lder's inequality and the definition of the Fourier transform:
\begin{equation*}
    ||f||_{L^{\infty}({\mathbb R}^d)} =\bigg|\bigg|\int_{\mathbb{R}^d}\hat{f}(\bm{\omega})e^{i2\pi\bm{\omega}\cdot\bm{x}}\:d\bm{\omega}\bigg|\bigg|_{L^{\infty}({\mathbb R}^d)}\leq\int_{\mathbb{R}^d}|\hat{f}(\bm{\omega})|\:d\bm{\omega} = ||\hat{f}||_{L^{1}(\mathbb{R}^d)}.
\end{equation*}
This ratio tends to grow as the target function becomes more irregular. Importantly, we observe logarithmic scaling with respect to this ratio, so especially as $||f||_{L^{\infty}({\mathbb R}^d)}$ becomes small, the scale of the best-approximation error is dominated by $||f||_{L^{\infty}({\mathbb R}^d)}^2$. 

The estimates \eqref{eqn:shallow_opt} and \eqref{eqn:deep_opt} facilitate a comparison between the approximation properties of shallow and deep Fourier networks. Indeed the approximation rate with respect to network complexity is the same, being $\mathcal{O}(1/W_F L_F)$ for both, but the dependence on the size of the target function is very different. While for shallow networks the approximation error depends on $||\hat{f}||_{L^1({\mathbb R}^d)}^2$, for deep networks this dependence is dominated by $||f||_{L^{\infty}({\mathbb R}^d)}^2$, especially for target functions satisfying $||\hat{f}||_{L^1({\mathbb R}^d)} \gg ||f||_{L^{\infty}({\mathbb R}^d)}$, which tend to be more irregular and with complex pattern. Hence, an important interpretation of this result is that more irregular functions are more efficiently approximated by deep networks.

\subsection{Fourier network error estimates for target functions on compact domains}
\label{sec:compactX}

The estimate \eqref{eqn:deep_opt} holds for all functions $f \in S$, defined on the whole ${\mathbb R}^d$. One can show that a similar estimate holds when the target function $f$ is defined on a compact domain $X \subset {\mathbb R}^d$, as long as it is bounded, and there exists an extension $f_e: {\mathbb R}^d \rightarrow {\mathbb R}$ of $f: X \subset{\mathbb R}^d \rightarrow {\mathbb R}$ such that $f_e \in S$ and 
$\norm{f_e}_{L^{\infty}({\mathbb R}^d)} \le c \, \norm{f}_{L^{\infty}(X)}$ for some constant $c \ge 1$. In this case, we may define
$$
\norm{\hat{f}}_{L^1({\mathbb R}^d)} := \inf \{ \norm{\hat{f_e}}_{L^1({\mathbb R}^d)}: f_e \in S \ \text{extends} \ f  \ \text{with} \  \norm{f_e}_{L^{\infty}({\mathbb R}^d)} \le c \, \norm{f}_{L^{\infty}(X)}   \}.
$$
Then, noting that 
$\norm{f}_{L_{\rho}^2(X)} \le \norm{f_e}_{L_{\rho}^2({\mathbb R}^d)}$ and 
$\norm{f}_{L^{\infty}(X)} \le \norm{f_e}_{L^{\infty}({\mathbb R}^d)}$ with $X \subset {\mathbb R}^d$, it is straightforward to show the following result.
\begin{theorem}\label{thm:ff_compactX} 
Let $f: X  \rightarrow {\mathbb R}$ be a bounded function on a compact domain $X \subset {\mathbb R}^d$ that accepts an extension $f_e: {\mathbb R}^d \rightarrow {\mathbb R}$ such that $f_e \in S$ and 
$\norm{f_e}_{L^{\infty}({\mathbb R}^d)} \le c \, \norm{f}_{L^{\infty}(X)}$ for some constant $c \ge 1$. 
There exists a residual Fourier network $f_{\Psi}$ with depth $L_{F}\ge 2$ and width $W_{F} \ge 1$ and parameters $(\underline{\bm{\omega}},\underline{\bm{\omega}}',\underline{\bm{b}},\underline{\bm{b}}')$ such that for all sufficiently large $W_{F} L_{F}$, with $W_{F}=\mathcal{O}(L_{F})^2$: 
\begin{equation}\label{eqn:deep_opt}
\norm{f(\bm{x}) - f_{\Psi}(\bm{x};\underline{\bm{\omega}},\underline{\bm{\omega}}',\underline{\bm{b}},\underline{\bm{b}}')}_{L_{\rho}^2(X)}^2 \leq C \frac{||f||_{L^{\infty}(X)}^{2}}{W_{F}L_{F}}\left(1 + \ln \frac{||\hat{f}||_{L^{1}(\mathbb{R}^d)}}{||f||_{L^{\infty}(X)}} \right)^{2},
\end{equation}
where $C$ is a positive constant.
\end{theorem}

\subsection{Illustrative numerical examples}

Randomness in Fourier networks was first introduced in \cite{1layerKammonen,kammonen2023smaller} for theoretical purposes; to allow Monte Carlo approximation and derive error estimates. Later, \cite{Davis_etal:23B} showed that random frequencies also enable fast randomized algorithms for training Fourier networks, avoiding the need to solve large, non-convex global optimization problems. 
Here, we present two numerical examples to illustrate the effectiveness of random Fourier networks in approximating functions with multiscale and discontinuous features.

\subsubsection{Example 1: A step function in one and two dimensions}

Here, we examine the performance of random Fourier networks in approximating discontinuous target functions. Given that these networks use a sinusoidal activation, their suitability for such tasks may initially seem doubtful. Classical Fourier approximations, for instance, exhibit Gibbs oscillations near discontinuities for any finite number of modes \cite{grafakos2008classical}. However, deep Fourier networks introduce recursive and compositional structures that go beyond traditional Fourier series. Notably, random Fourier networks trained with the randomized algorithm from \cite{Davis_etal:23B} have been shown to approximate discontinuous features sharply, without exhibiting Gibbs phenomena. 
Analytic work exploring these numerical observations is an exciting direction for future work.

\begin{figure}[!htb]
    \centering
    \includegraphics[width=0.7\linewidth]{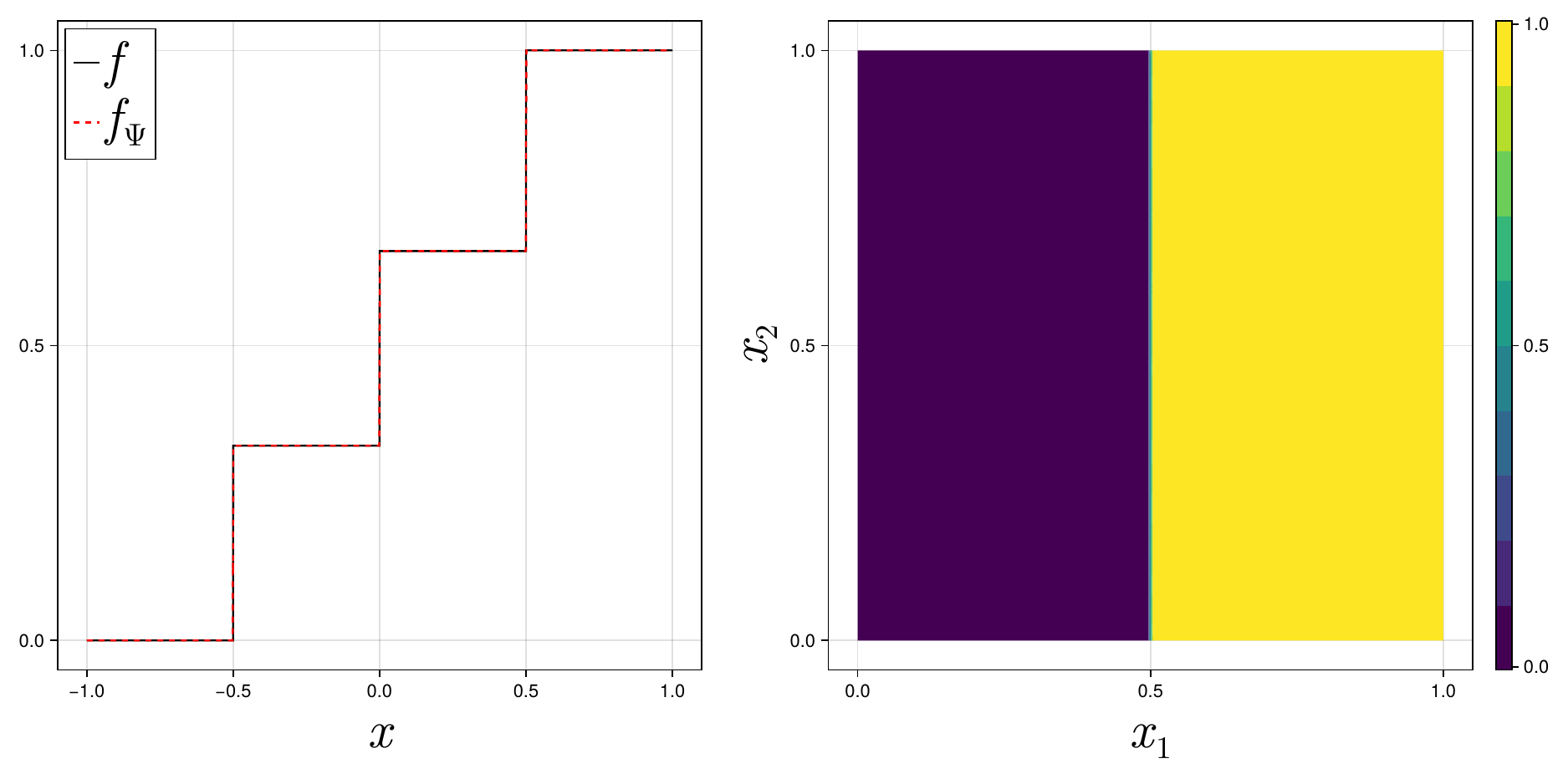}
    \caption{Left: Random Fourier network prediction (red, dashed) in approximating the stairstep function (black, solid). Right: Contour plot of random Fourier network reconstruction of a two-dimensional discontinuous interface.}
    \label{fig:stairstep}
\end{figure}

In Figure~\ref{fig:stairstep}, 
we present two such examples. 
On the left, we plot the target function (black, solid) alongside the random Fourier network prediction (red, dashed) over 10,000 test inputs uniformly distributed in $[-1,1]$. The network employed here is relatively small, with $\mathcal{O}(100)$ parameters, and it was trained using just 1,000 samples drawn randomly from a uniform distribution in $[-1,1]$. 
On the right, we showcase a contour plot depicting the random Fourier network approximation of a two-dimensional function with a jump discontinuity. In this case, the network contains $\mathcal{O}(1000)$ parameters, is trained on 6,400 samples, and is tested on 100,000 points. 
Remarkably, in both cases, despite being trained on finite data and employing sinusoidal activation functions, the random Fourier networks yield sharp predictions at the discontinuous interfaces without Gibbs oscillations, achieving an overall tolerance of $\mathcal{O}(10^{-5})$.

\subsubsection{Example 2: An image with multiscale features}

\begin{figure}[!ht]
    \centering
    \includegraphics[width=0.3\linewidth]{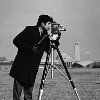}
    \hspace{1cm}
    \includegraphics[width=0.3\linewidth]{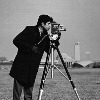}
    \caption{Original image (left) and rFN reconstruction (right)}
    \label{fig:camera_man}
\end{figure}

Random Fourier networks are also adept at conventional deep learning tasks such as image reconstruction. 
Grayscale images, like the ``cameraman'' shown in Figure~\ref{fig:camera_man}, can be represented as functions $f:(x_1,x_2)\in \mathbb{R}^2\rightarrow[0,1]$, where each pixel location $(x_1,x_2)$ corresponds to a scalar value that encodes the intensity or color of that pixel. 
In images like this one, sharp transitions in pixel values across different scales suggest that the function $f$ exhibits both multiscale and irregular features. On a larger scale, there is a distinct contrast between the cameraman's dark coat and the light gray-white background, while at finer scales, we observe sharp transitions in pixel values that capture details such as the cameraman's eyes and the intricate details of the camera itself.

The presence of high-frequency features at multiple scales presents a significant approximation challenge for deep neural networks. However, random Fourier networks trained using the algorithm presented in \cite{Davis_etal:23B} have shown the capability to effectively approximate such functions with low tolerances. The left image displays the original, while the right image shows the reconstruction produced by the random Fourier network. Can you spot the difference?

In Figure \ref{fig:camera_man}, the original image has resolution $512 \times 512$ pixels, but has been coarsened to $256 \times 256$ for efficient training, totaling {\bf 65,536} training data points. The training is supervised on all pixels in this coarsened image and tested on this same coarsened grid. The neural network architecture comprises a width of $W=2$ and a depth of $L=4000$, with a total of {\bf 63,992} parameters, of which 31,996 are frequency parameters.  Interestingly, the network's approximation power enables image compression of approximately $2.4\%$, calculated as $(65,536 - 63,992)/  65,536 \approx 2.4 \%$.

\section{Deep ReLU networks}
\label{sec:approximation_errror_ReLU}

We have discussed error-complexity estimates for Fourier networks, which have appeared in a few isolated works \cite{1layerKammonen,kammonen2023smaller,Davis_etal:23B}. A natural question is whether similar estimates hold for ReLU networks, which are widely studied in the neural network literature. Initial exploration into this question was explored in \cite{davis2024mathematically}. Borrowing strategies from that work, we derive a new complexity result for ReLU networks, formulated in Theorem~\ref{thm:main_theoretical_result}.

\subsection{A complexity estimate}

\smallskip
\noindent
{\bf Assumptions.} 
Throughout this section, we consider bounded target functions $f: X \rightarrow \mathbb{R}$ on a compact domain $X \subset \mathbb{R}^d$, with an extension as described in Section \ref{sec:compactX}. We also assume that all frequencies $(\underline{\bm{\omega}}, \underline{\bm{\omega}}')$ in the residual Fourier network $f_{\Psi}$ of Theorem \ref{thm:ff_compactX} are uniformly bounded. This assumption is motivated by first cutting out very high-frequency components of $f$ before approximating it with a residual Fourier network. The cutoff should be chosen so that the total error remains within a desired tolerance $\varepsilon$.

\smallskip
\noindent
{\bf Space of uniform-width standard ReLU networks.} 
Before presenting the main result, we revisit the space of uniform-width standard ReLU networks, introduced in Section \ref{sec:relu-net}. These networks have $d$ input neurons, one output neuron, and $L$ hidden layers, each with $W$ neurons, and are parameterized in terms of weight-bias tuples $\Phi = \{ (M_{\ell} , b_{\ell}) \}_{\ell
  = 0}^{L}$:
$$
{\mathcal N}_{W,L}^X = \{ f_{\Phi}: X \subseteq {\mathbb R}^d \rightarrow {\mathbb R}, \ \ \Phi = \{ (M_{\ell} , b_{\ell}) \}_{\ell
  = 0}^{L}  \in {\mathbb R}^{(L-1) W^2 + (d+n_L)W} \times {\mathbb R}^{LW + n_L} \}.
$$
We use the notation ${\mathcal N}_{W,L}^X$ to indicate that the network operates on domain $X$, distinguishing it from the previously used ${\mathcal N}_{W,L}$.

\begin{theorem}\label{thm:main_theoretical_result}
Let $f: X \subseteq {\mathbb R}^d \rightarrow {\mathbb R}$ be a bounded function as in Theorem \ref{thm:ff_compactX}. 
For any $\varepsilon \in (0,1/2)$, there exists a ReLU network $f_{\Phi}\in \mathcal{N}_{W,L}^X$ with $L\geq 2$, $W > 2d+2$, and satisfying:
\begin{equation}\label{eqn:main_theoretical_tol}
    ||f - f_{\Phi}||_{L^{2}(X)}\leq\varepsilon,
\end{equation}
such that 
\begin{equation}\label{eqn:main_theoretical_complexity_result}
    W \, L^{1/3} \leq C \, ||f||^{2}_{L^{\infty}(X)} \, \varepsilon^{-2} \, \log_2^{4/3}(\varepsilon^{-1}),
\end{equation}
where $C>0$ is a constant depending linearly on $d$ and logaritmically on $||f||_{L^{\infty}(X)}$ and $||\hat{f}||_{L^{1}(\mathbb{R}^d)}$. 
\end{theorem}

\subsection{Proof of the complexity estimate}

Theorem \ref{thm:main_theoretical_result} can be proved by conducting a constructive approximation of a residual Fourier network by a ReLU network of comparable complexity, a task that leverages a more general type of special ReLU networks introduced in Section \ref{sec:relu-net}, referred here as to generalized special ReLU networks. Below, we provide a proof for the case $d=1$ and $X = [-D,D]$, with $D \ge 1$. We remark here that the proof for $d =1$ can be straightforwardly generalized to the case $d\geq 2$. The proof involves the following four steps:
\smallskip

\noindent{\bf Step 1.} 
Reformulate residual Fourier networks expressed in terms of the cosine function.

\smallskip

\noindent{\bf Step 2.} 
Demonstrate that a generic cosine function, and by extension, a linear combination of cosine functions, can be approximated by a ReLU network with quantifiable complexity.

\smallskip

\noindent{\bf Step 3.} 
Utilize generalized special ReLU networks and the result from Step 2 to construct a ReLU network that approximates a residual Fourier network.

\smallskip

\noindent{\bf Step 4.} 
Combine the constructed ReLU network with Theorem \ref{thm:ff_compactX} to derive the desired result.

\subsubsection{Step 1: A real-valued formulation of Fourier networks} 

We introduce an explicit real-variable formulation of Fourier networks. Keeping the notation consistent with Section~\ref{sec:rFNNs}, a Fourier network of width $W_{F}$ and depth $L_{F}$ realizes the function $f_{\Psi}=z_{L_{F}}(x)$, for $x \in [-D,D]$, where $z_{L_{F}}$ results from the following recursion:
\begin{align}
z_{1}(x) &= g_{1}(x;\underline{\bm{\omega}}_{1},\underline{\bm{b}}_{1}), \label{eq:real_form1}\\
z_{\ell}(x) &= z_{\ell-1}(\bm{x}) + g_{\ell}(x;\underline{\bm{\omega}}_{\ell},\underline{\bm{b}}_{\ell}) + g_{\ell}'(z_{\ell-1};\underline{\bm{\omega}}'_{\ell},\underline{\bm{b}}_{\ell}'), \qquad \ell=2,\dotsc. L_{F}, \label{eq:real_form2}
\end{align}
The explicit real-variable form of $g_{\ell}$ and $g_{\ell}'$ are given by: 
\begin{align}
    g_{\ell}(x;\underline{\bm{\omega}}_{\ell},\underline{\bm{b}}_{\ell}) &= \sum_{j=1}^{W_{F}}|b_{\ell j}|\cos\left(\omega_{\ell j} \, x -\tan^{-1}\left(\frac{-\Im(b_{\ell j})}{\Re(b_{\ell j})}\right)\right),\label{eqn:g_ell}\\
    g_{\ell}'(z_{\ell-1};\underline{\bm{\omega}}_{\ell}',\underline{\bm{b}}_{\ell}') &= \sum_{j=1}^{W_{F}}|b_{\ell j}'|\cos\left(\omega'_{\ell j}\, z_{\ell-1} -\tan^{-1}\left(\frac{-\Im(b_{\ell j}')}{\Re(b_{\ell j}')}\right)\right)\label{eqn:g_ell_prime},
\end{align}
where $|\cdot|$ denotes the complex modulus.

\subsubsection{Step 2: Approximating sum of cosine functions by ReLU networks} 

We first observe that a primary building block in a constructive approximation will be the ability to represent a one-dimensional cosine function by a ReLU network of bounded complexity. 
\begin{lemma}\label{lem:cos}
Let $\alpha \in (0, \infty)$, $\beta \in (-\infty, \infty)$, $x \in [-D,D]$, with $D\ge 1$, and $\varepsilon\in (0,1/2)$. Then there exists a constant $C>0$ and a ReLU network $\Phi \in \mathcal{N}^{[-D,D]}_{W,L}$ with complexity 
$$
W\leq 7, \qquad L \leq C (\log_{2}^2 \varepsilon^{-1} + \log_{2} \lceil \alpha D \rceil),
$$
satisfying
\begin{equation*}
        ||\cos( \alpha \, x + \beta) - f_{\Phi}(x)||_{L^{\infty}[-D,D]}\leq \varepsilon.
\end{equation*}
The same error-complexity result holds for the sine function.  
\end{lemma}
Proof of Lemma \ref{lem:cos} uses the following lemma.

\begin{lemma}\label{lem:smooth}
Let $f \in C^{\infty}[-1,1]$ be a real-valued function with $|| f^{(n)} (x) ||_{L^{\infty}[-1,1]} \le n!$. For every $\varepsilon \in (0,1/2)$, there exists a ReLU network $\Phi \in {\mathcal N}_{W,L}^{[-1,1]}$, with $W \le 7$ and $L \le C \, \log_2^2 \varepsilon^{-1}$, where $C>0$ is a constant, satisfying
$$
|| f(x) - f_{\Phi}(x) ||_{L^{\infty}[-1,1]} \le \varepsilon.
$$
\end{lemma}
\begin{proof}
We start with a known result in polynomial interpolation (see e.g. \cite{Elbrachter_etal:21}): For any function $f$ satisfying the conditions in the Lemma, there exists a polynomial $p_m(x) = \sum_{j=0}^m a_j \, x^j$ of degree $m$ interpolating $f$ at $m+1$ Chebyshev points such that
$$
|| f(x) - p_m(x) ||_{L^{\infty}[-1,1]} \le \frac{|| f^{(m+1)} (x) ||_{L^{\infty}[-1,1]}}{2^m \, (m+1)!} \le 2^{-m}, \qquad || a ||_{\infty} := \max_j |a_j| \le 2(m+1) 3^m.
$$
Applying Lemma \ref{lem:polynom_b} with approximation error $\varepsilon/2$ and $D=1$ to $p_m$, we obtain that there is a ReLU network $\Phi \in {\mathcal N}_{W,L}^{[-1,1]}$, with $W \le 7$ and $L \le C \, m \, (\log_2 \varepsilon^{-1} + m)$, that approximates $p_m$, satisfying
$$
|| f(x) - f_{\Phi}(x) ||_{L^{\infty}[-1,1]} \le || f(x) - p_m(x) ||_{L^{\infty}[-1,1]} + || p_m(x) - f_{\Phi}(x) ||_{L^{\infty}[-1,1]} \le 2^{-m} + \varepsilon/2,
$$
Setting $m = \lceil \log_2 (2/\varepsilon) \rceil$, we get the desired result.
\end{proof}

\medskip
\noindent
{\it Proof of Lemma \ref{lem:cos}.} 
Using the identity $\cos (\alpha \, x + \beta) = \cos (\alpha \, x) \, \cos(\beta) - \sin (\alpha \, x) \, \sin(\beta)$, and thanks to Proposition \ref{prop:scalar_mult}, it suffices to show the desired complexity estimates for $\cos(\alpha \, x)$ and $\sin(\alpha \, x)$, with $\beta =0$. Here, we only prove this for the cosine function. The proof for the sine function follows similarly. Henceforth, we set $\beta = 0$. First let $D=1$ and $\alpha \in (0,\pi]$. The function $f(x) = (6/\pi^3) \, \cos (\alpha \, x)$ satisfies $|| f^{(n)} (x) ||_{L^{\infty}[-1,1]} \le 6 \, \pi^{n-3} \le  n!$, for every $n \in {\mathbb N}$. Hence, by Lemma \ref{lem:smooth}, with approximation error $(6/ \pi^3) \, \varepsilon$, there is a network $\Phi$ with complexity $W \le 7$ and $L \le C \, \log_2^2 \varepsilon^{-1}$, satisfying:
$$
|| \frac{6}{\pi^3} \, \cos (\alpha x) - f_{\Phi} (x) ||_{L^{\infty}[-1,1]} \le \frac{6}{\pi^3} \, \varepsilon, \qquad \alpha \in (0, \pi].
$$
This implies that, by Proposition \ref{prop:scalar_mult}, the desired network $\tilde{\Phi}$ is the scalar multiplication network that outputs $f_{\tilde{\Phi}}(x) = (\pi^3/6) \, f_{\Phi}(x)$, with a complexity equal to the complexity of $\Phi$, satisfying:
$$
|| \cos (\alpha x) - f_{\tilde{\Phi}} (x) ||_{L^{\infty}[-1,1]} = || \cos (\alpha x) - \frac{\pi^3}{6} \, f_{\Phi} (x) ||_{L^{\infty}[-1,1]} \le \varepsilon, \qquad \alpha \in (0, \pi].
$$
Next, let $D=1$ and $\alpha > \pi$. By the periodicity and symmetry of the cosine function, we have:
$$
\cos(\pi \, 2^s \, x) = \cos (\pi \, h_s(|x|)), \qquad s \in {\mathbb N}, \qquad x \in [-1,1],
$$
where $h_s: [0,1] \rightarrow [0,1]$ is the sawtooth function in \eqref{eqn:sawtooth}. Choosing $s = \lceil \log_2 \alpha - \log_2 \pi \rceil \ge 1$, we will have
$$
\tilde{\alpha} := \frac{\alpha}{\pi \, 2^s} \in (\frac{1}{2}, 1].
$$
We hence can write:
$$
\cos (\alpha \, x) = \cos (\pi \, 2^s \, \tilde{\alpha} \, x) = \cos (\pi \, h_s(\tilde{\alpha} \, |x|)) =: \cos (\pi \, y), \qquad x \in [-1,1]. 
$$
Since $y := h_s(\tilde{\alpha} \, |x|) \in [0,1]$, approximating $\cos(\alpha \, x)$ with $\alpha > \pi$ amounts to approximating $\cos(\pi \, y)$, falling into the case where $\alpha = \pi$. There is hence a network $\tilde{\Phi}$, with complexity $\tilde{W} \le 7$ and $\tilde{L} \le C \, \log_2^2 \varepsilon^{-1}$, that takes $y$ as input and outputs an $\varepsilon$-approximation of $\cos(\pi \, y)$, satisfying:
$$
|| \cos (\alpha x) - f_{\tilde{\Phi}} (h_s(\tilde{\alpha} \, |x|)) ||_{L^{\infty}[-1,1]} =
|| \cos (\pi \, y) - f_{\tilde{\Phi}} (y) ||_{L^{\infty}[0,1]}  \le \varepsilon, \qquad \alpha > \pi.
$$
The desired network that takes $x \in [-1,1]$ as input and outputs $f_{\tilde{\Phi}} (h_s(\tilde{\alpha} \, |x|))$ is then given by the composition of the following three networks:
\begin{itemize}
\item $\Phi_1$: the network that takes $x \in [-1,1]$ and outputs $z:= \tilde{\alpha} \, |x| \in [0,1]$. This network is given by $z = \tilde{\alpha} \, \sigma(x) + \tilde{\alpha} \, \sigma(-x)$, and hence has width $W_1 = 2$ and depth $L_1= 1$.

\item $\Phi_2$: the network that takes $z \in [0,1]$ and outputs $y = h_s(z) \in [0,1]$. This network is given in \eqref{eqn:network_hs}, with width $W_2=3$ and depth $L_2 = s \le \log_2 \lceil \alpha \rceil$. 

\item $\tilde{\Phi}$: the network $\tilde{\Phi}$ that takes $y \in [0,1]$ and outputs the desired $f_{\tilde{\Phi}} (y)$, with complexity $\tilde{W} \le 7$ and $\tilde{L} \le C \, \log_2^2 \varepsilon^{-1}$.
\end{itemize}
By Proposition \ref{prop:composition}, the desired ReLU network $\Phi_{\alpha} = \tilde{\Phi} \circ \Phi_2 \circ \Phi_1$, taking input $x$ and outputting an $\varepsilon$-approximation of $\cos (\alpha \, x)$, has complexity 
$$
W \le 7, \qquad L \le C \, \log_2^2 \varepsilon^{-1} + \log_2 \lceil \alpha \rceil + 1 \le \tilde{C} \, (\log_2^2 \varepsilon^{-1} + \log_2 \lceil \alpha \rceil ).
$$
with some constant $\tilde{C} > 0$, as desired. 
We finally let $D \ge 1$ and build a network $\Phi_{\alpha, D}$ that takes $x\in[-D,D]$ as input and outputs an $\varepsilon$-approximation of $\cos (\alpha x)$ for $x \in [-D,D]$ with desired complexity. To this end, we define $y=x/D \in [-1,1]$. Then, noting that
$$
|| \cos (\alpha x) - f_{\Phi_{\alpha,D}} (x) ||_{L^{\infty}[-D,D]} 
= 
|| \cos (\alpha D \, y) - f_{\Phi_{\alpha,D}} (Dy) ||_{L^{\infty}[-1,1]}, 
$$
the desired network $\Phi_{\alpha, D}$ will be $\Phi_{\alpha \, D}$, which takes $y = x/D \in [-1,1]$ as input and outputs an $\varepsilon$-approximation of $\cos (\alpha D \, y)$. This is the network that we constructed for the case $D=1$, with the new frequency $\alpha D$ instead of $\alpha$. This completes the proof. \qed

We next approximate linear combinations of cosine functions, appearing in \eqref{eqn:g_ell}-\eqref{eqn:g_ell_prime}.

\begin{lemma}\label{lem:cos_sum}
Consider the bounded function
$$
g(x) = \sum_{j=1}^{W_F} a_j \, \cos (\alpha_j \, x + \beta_j), \qquad x \in [-D,D], \qquad D\ge 1,
$$
where $a_j \in [0, \infty)$ and $\alpha_j \in (0, \infty)$ are uniformly bounded, and $\beta_j \in (-\infty, \infty)$. 
Then for any $\varepsilon\in (0,1/2)$, there exists a constant $C>0$ and a ReLU network $\Phi \in \mathcal{N}^{[-D,D]}_{W,L}$ with complexity 
$$
W\leq 7 \, W_F, \qquad L \leq C \left( \log_{2}^2 (W_F \, \varepsilon^{-1}) + \log_{2} \lceil \alpha \, D \rceil \right), \qquad \alpha := \max_j \alpha_j,
$$
satisfying
\begin{equation*}
        ||g(x) - f_{\Phi}(x)||_{L^{\infty}[-D,D]}\leq \varepsilon.
\end{equation*}
\end{lemma}
\begin{proof}
Utilizing Lemma \ref{lem:cos} and Proposition \ref{prop:scalar_mult}, we first construct $W_F$ networks $\Phi_j \in {\mathcal N}_{W_j,L_j}^{[-D,D]}$ for $j=1, \dotsc, W_F$, each approximating $a_j \cos(\alpha_j x + \beta_j)$ within tolerance $\varepsilon/W_F$, and with complexities satisfying $W_j \leq 7$ and $L_j \leq C (\log_{2}^2 (W_F \, \varepsilon^{-1}) + \log_{2} \lceil \alpha_j D \rceil)$. 
Next, the desired network $\Phi \in {\mathcal N}_{W,L}^{[-D,D]}$ is constructed as in Proposition \ref{prop:sum_width} by stacking the individual $\Phi_j$ networks vertically. All individual networks $\Phi_j$ can be adjusted to have the same depth $L \le C (\log_{2}^2 (W_F \, \varepsilon^{-1}) + \log_{2} \lceil \alpha D \rceil)$, where $\alpha = \max_j \alpha_j$. This is achieved by extending the shallower networks with additional hidden layers that implement identity mappings using the relation $\sigma(y) - \sigma(-y) = y$, for $y \in {\mathbb R}$. 
As a result, the constructed network $\Phi$ will meet the desired complexity and accuracy.
\end{proof}

\subsubsection{Step 3: Approximating a Fourier network by a ReLU network} 

We first introduce a more general type of special ReLU networks defined in Section \ref{sec:relu-net}, as follows.

\smallskip
\noindent
{\bf Generalized special ReLU networks.} A generalized special ReLU network generalizes a special ReLU network by allowing neurons in the collation channel to feed the collected intermediate computations back into their next immediate layer of the computational channel. This type of ``generalized'' collation channel is key to implementing the recursive linear combination and composition present in residual Fourier networks. 
These networks have $d$ input neurons, one output neuron, and $L$ hidden layers, each with $W$ neurons, and are parameterized in terms of weight-bias tuples $\tilde{\Phi} = \{ (\tilde{M}_{\ell} , \tilde{b}_{\ell}) \}_{\ell
  = 0}^{L}$:
$$
{\mathcal G}_{W,L}^X = \{ f_{\tilde{\Phi}}: X \subseteq {\mathbb R}^d \rightarrow {\mathbb R}, \ \ \tilde{\Phi} = \{ (\tilde{M}_{\ell} , \tilde{b}_{\ell}) \}_{\ell
  = 0}^{L}  \in {\mathbb R}^{(L-1) W^2 + (d+n_L)W} \times {\mathbb R}^{LW + n_L} \}.
$$

Similar to special ReLU networks, generalized special ReLU networks can be re-parameterized into standard ReLU networks. Precisely, given any function $f_{\tilde{\Phi}} \in {\mathcal G}^{X}_{W,L}$ corresponding to a generalized special network $\tilde{\Phi}=\{ \tilde{M}^{(\ell)} , \tilde{\bm{b}}^{(\ell)} \}_{\ell = 0}^{L}$, a standard ReLU network $\Phi = \{ {M}^{(\ell)} , \bm{b}^{(\ell)} \}_{\ell= 0}^{L}$ can be constructed that produces the same function $f_{\Phi} \equiv f_{\tilde\Phi}$, where $f_{\Phi} \in \mathcal{N}^{X}_{W+1, L}$ if the input takes only non-negative values (i.e. $X \subset {\mathbb R}_+^d$), and $f_{\Phi} \in \mathcal{N}^{X}_{W+d+1, L}$ otherwise. The extra neuron added here in each layer is because the values in the collation channel may be negative, and hence we add one more channel so that using the identity relation $\sigma(z) - \sigma(-z) = z$ the negative value can be recovered after applying ReLU activation. For more details we refer to \cite{davis2024mathematically}.

\begin{remark} 
The generalized special ReLU networks introduced here not only facilitate the construction of ReLU networks for deriving error-complexity estimates but also offer a natural framework for designing residual networks (ResNets) through recursive computations. However, we do not further explore this computational advantage in the current work.  
\end{remark}

\begin{lemma}\label{lem:fourier_relu}
Let $W_F \in {\mathbb N}$ and $L_F \in {\mathbb N}$, and consider the function $f_{\Psi} = z_{L_F}: x \in [-D,D] \mapsto {\mathbb R}$ as defined in \eqref{eq:real_form1}-\eqref{eqn:g_ell_prime}, with bounded frequencies and amplitudes. 
For any $\varepsilon \in (0, 1/2)$, there exists a ReLU network $\Phi_{\varepsilon} \in {\mathcal N}_{W,L}^{[-D,D]}$, satisfying 
\begin{equation}\label{eq:accuracy_constraint}
|| z_{L_F} - f_{\Phi_{\varepsilon}} ||_{L^{\infty}[-D,D]} \le \varepsilon,
\end{equation}
such that
\begin{equation}\label{eq:network_complexity_relu}
W \le C_{W} \, W_F, \qquad L \le C_{L} \, L_F^3 \, \log_2^2(W_F L_F) \, \log_2^2(\varepsilon^{-1}),
\end{equation}
where $C_{W}$ and $C_L$ are positive, bounded constants that may depend logarithmically on the frequencies and amplitudes and $D$.
\end{lemma}
\begin{proof}
As the ReLU approximation of $z_{L_F}$ in \eqref{eq:real_form1}-\eqref{eqn:g_ell_prime}, consider the following recursive formula:
\begin{equation}\label{eq:recursive_relu}
f_1(x) = \Phi_{1,\delta}(x), \qquad
f_{\ell}(x) = f_{\ell -1} (x) + \Phi_{\ell,\delta/2}(x) + \Phi'_{\ell,\delta/2}(f_{\ell-1}(x)), 
\quad \ell = 2, 3, \dotsc, L_F,
\end{equation}
where $\Phi_{\ell,\delta} \in {\mathcal N}_{W_{\ell},L_{\ell}}^{[-D,D]}$ and $\Phi'_{\ell,\delta} \in {\mathcal N}_{W_{\ell}',L_{\ell}'}^{[-D_{\ell-1},D_{\ell-1}]}$ are the ReLU networks of Lemma \ref{lem:cos_sum}, satisfying
\begin{equation}\label{eq:error_delta_phi}
||\Phi_{\ell,\delta}(x) - g_{\ell}(x)||_{L^{\infty}[-D,D]}\leq \delta, 
\qquad
||\Phi_{\ell,\delta}'(f_{\ell-1}(x)) - g'_{\ell}(f_{\ell-1}(x))||_{L^{\infty}[-D,D]} \leq \delta,
\end{equation}
with widths $W_{\ell} \le 7 \, W_F$ and $W_{\ell}' \le 7 \, W_F$, and depths
\begin{equation}\label{eq:depths_L_Lprime}
L_{\ell} \le C_{\ell} \left( \log_{2}^2 (W_F \, \delta^{-1}) + \log_{2} \lceil \omega_{\ell} \, D \rceil \right),
\qquad
L_{\ell}' \le C_{\ell}' \left( \log_{2}^2 (W_F \, \delta^{-1}) + \log_{2} \lceil \omega_{\ell}' \, D_{\ell-1} \rceil \right),
\end{equation}
where $\omega_{\ell} := \max_j \omega_{\ell j}$ and $\omega_{\ell}' := \max_j \omega_{\ell j}'$, and assuming $f_{\ell}(x) \in [-D_{\ell}, D_{\ell}]$ for any $x \in [-D,D]$. 
We first show that 
\begin{equation}\label{eq:error_f_ell}
|| f_{\ell}(x) - z_{\ell} (x) ||_{L^{\infty}[-D,D]} \le \delta \, \ell \, (1 + c_0 \, W_F)^{\ell-1}, \qquad c_0 := \max \bigl( 1, \max_{\ell, j} ( |b_{\ell j}'| \, \omega_{\ell j}') \bigr).    
\end{equation}
We note that \eqref{eq:error_f_ell} holds for $\ell=1$ by construction. For $\ell \ge 2$, by triangle inequality, we have:
$$
|| f_{\ell}(x) - z_{\ell} (x) ||_{L^{\infty}[-D,D]} \le \delta_{I} + \delta_{II},
$$
where
\begin{align*}
\delta_I &:= || f_{\ell}(x) - \bigl( f_{\ell-1}(x) + g_{\ell}(x) + g_{\ell}'(f_{\ell-1}(x)) \bigr) ||_{L^{\infty}[-D,D]}, \\ 
\delta_{II} &:= || \bigl( f_{\ell-1}(x) + g_{\ell}(x) + g_{\ell}'(f_{\ell-1}(x)) \bigr) - z_{\ell}(x) ||_{L^{\infty}[-D,D]}.
\end{align*}
By triangle inequality and \eqref{eq:recursive_relu} and \eqref{eq:error_delta_phi}, we obtain
$$
\delta_I \le || \Phi_{\ell, \delta/2} (x) - g_{\ell}(x) ||_{L^{\infty}[-D,D]}  + 
|| \Phi_{\ell, \delta/2}' (f_{\ell-1}(x)) -g_{\ell}'(f_{\ell-1}(x))  ||_{L^{\infty}[-D,D]} \le \frac{\delta}{2} + \frac{\delta}{2} = \delta.
$$
Moreover, by triangle inequality and \eqref{eq:real_form2}, we obtain
$$
\delta_{II} \le  || f_{\ell-1} (x) - z_{\ell-1}(x) ||_{L^{\infty}[-D,D]} + 
|| g_{\ell}'(f_{\ell-1} (x)) - g_{\ell}'(z_{\ell-1}(x)) ||_{L^{\infty}[-D,D]}
$$
The second term in the right hand side of the above inequality can be bounded as:
$$
|| g_{\ell}'(f_{\ell-1} (x)) - g_{\ell}'(z_{\ell-1}(x)) ||_{L^{\infty}[-D,D]} \le W_F \, \max_j (|b_{\ell j}'| \omega_{\ell j}') \, || f_{\ell-1}(x) - z_{\ell-1} (x) ||_{L^{\infty}[-D,D]}.
$$
This follows from \eqref{eqn:g_ell_prime} and inequalities $|\cos \alpha - \cos \beta | \le | \alpha - \beta |$ and $|\sin \alpha - \sin \beta | \le | \alpha - \beta |$, which hold by the mean value theorem. 
We hence obtain the recursive inequality:
$$
|| f_{\ell}(x) - z_{\ell} (x) ||_{L^{\infty}[-D,D]} \le \delta + (1 + W_F \, \max_{\ell, j} ( |b_{\ell j}'| \omega_{\ell j}' )) \, || f_{\ell-1}(x) - z_{\ell-1} (x) ||_{L^{\infty}[-D,D]},
$$
from which \eqref{eq:error_f_ell} follows directly. 
Hence, the desired accuracy \eqref{eq:accuracy_constraint} will be achieved if we choose $\delta$ such that
\begin{equation}\label{eq:tol_selection}
\delta \, L_F \, (1 + c_0 \, W_F)^{L_F-1} = \varepsilon, \qquad c_0 := \max \bigl( 1, \max_{\ell, j} ( |b_{\ell j}'| \, \omega_{\ell j}') \bigr).
\end{equation}
We next return to the assumption $f_{\ell}(x) \in [-D_{\ell}, D_{\ell}]$ and compute $D_{\ell}$, as follows. Using triangle inequality, we write
$$
|| f_{\ell}(x) ||_{L^{\infty}[-D,D]} \le 
|| z_{\ell}(x) ||_{L^{\infty}[-D,D]} + 
|| f_{\ell}(x) - z_{\ell}(x) ||_{L^{\infty}[-D,D]}. 
$$
By \eqref{eq:error_f_ell} and \eqref{eq:tol_selection}, the second term in the above inequality is bounded by $\varepsilon$. The first term in the above inequality can be bounded using the recursive formula \eqref{eq:real_form1}-\eqref{eq:real_form2}:
\begin{align*}
&|| z_1(x) ||_{L^{\infty}[-D,D]} \le W_F \, \max_{j} |b_{1 j}|, \\
&|| z_{\ell}(x) ||_{L^{\infty}[-D,D]} \le 
|| z_{\ell-1}(x) ||_{L^{\infty}[-D,D]} +
W_F \, \max_{\ell,j} (|b_{\ell j}| + |b_{\ell j}'| ), \qquad
\ell = 2,  \dotsc, L_F,
\end{align*}
which implies
$$
|| z_{\ell}(x) ||_{L^{\infty}[-D,D]} \le W_F \max_{j} |b_{1 j}| + (\ell - 1) \, W_F \, \max_{\ell,j} (|b_{\ell j}| + |b_{\ell j}'| ) \le c_1 \, \ell \, W_F, 
$$
with $c_1 := \max (1, 2 \max_{\ell,j} (|b_{\ell j}| , |b_{\ell j}'| ))$. We hence obtain:
\begin{equation}\label{eq:D_ell}
D_{\ell} \le  c_1 \, W_F \, L_F + \varepsilon 
\le 
2 c_1 \, W_F \, L_F, \quad c_1 := \max \bigl( 1, 2 \max_{\ell,j} (|b_{\ell j}| , |b_{\ell j}'| ) \bigr), \quad \ell =1, \dotsc, L_F.
\end{equation}
It remains to show that one can construct a network $\Phi_{\varepsilon} \in {\mathcal N}_{W,L}^{[-D,D]}$ that realizes $f_{L_F}$ with desired complexity \eqref{eq:network_complexity_relu}. 
We first adjust each network pair $( \Phi_{\ell, \delta/2}, \Phi_{\ell, \delta/2}')$, for $\ell = 2, \dotsc, L_F$, so they have the same depth $L_{\ell}'' = \max(L_{\ell}, L_{\ell}')$. This is achieved by extending the shallower networks with additional hidden layers that implement identity mappings. We will show that:
\begin{equation}\label{eq:Ldoubleprime}
L_{\ell}'' = \max(L_{\ell}, L_{\ell}') \le C_L \, L_F^2 \, \log_2^2(W_F L_F) \, \log_2^2(\varepsilon^{-1}) =: L'',
\end{equation}
where $C_L$ is a positive constant that may depend logarithmically on the frequencies and amplitudes and $D$. To this end, by \eqref{eq:depths_L_Lprime} and \eqref{eq:tol_selection} and \eqref{eq:D_ell}, we write:
$$
L_{\ell}'' \le \hat{C} \, \Bigl( \log_2^2 \bigl( W_F \, L_F \, (1 + c_0 \, W_F)^{L_F-1} \, \varepsilon^{-1} \bigr) + \log_2 \bigl( \max ( \lceil \omega \, D \rceil, \lceil 2 \, \omega' \, c_1 \, W_F \, L_F  \rceil)  \bigr) \Bigr) , 
$$
where $\hat{C} := \max_{\ell} (C_{\ell}, C_{\ell}')$, $\omega := \max_{\ell} \omega_{\ell}$, and $\omega' := \max_{\ell} \omega_{\ell}'$. 
From this, we obtain \eqref{eq:Ldoubleprime} after simple algebraic manipulations using logarithm properties and noting that with $\alpha \ge 1$ and $\beta \ge 0$ we have $\alpha + \beta \le (1 + \beta) \, \alpha$ and  $(\alpha + \beta)^2 + \alpha \le 2 \, (\alpha + \beta)^2$. 
Figure \ref{fig:network_relu_fourier} shows the construction of a generalized special network $\tilde{\Phi}_{\varepsilon} \in {\mathcal G}_{\tilde{W},\tilde{L}}^{[-D,D]}$, realizing $f_{L_F}$ and formed by stacking networks $\Phi_{1, \delta}, \Phi_{2, \delta/2} \dotsc, \Phi_{L_F, \delta/2}$ and $\Phi_{2, \delta/2}' \dotsc, \Phi_{L_F, \delta/2}'$, all with the same widths bounded by $7 W_F$, and with depths bounded by $L''$.
\begin{figure}[!h]
\centering
        \includegraphics[width=0.85\linewidth]{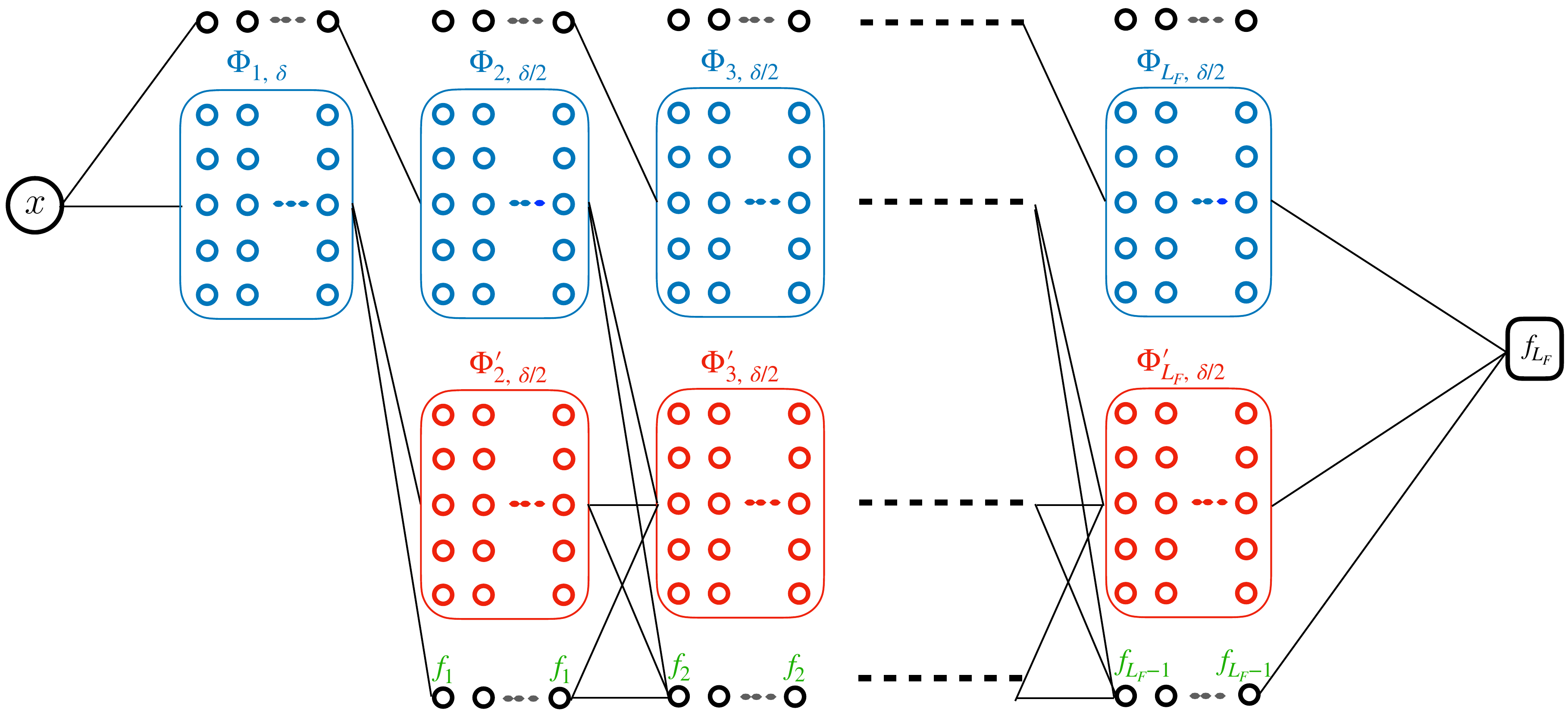}
\caption{Construction of the generalized special network that outputs $f_{L_F}$.}
\label{fig:network_relu_fourier}
\end{figure} 
This network has the complexity
$$
\tilde{W} \le 14 \, W_F + 2, \qquad \tilde{L} \le L_F \, L'' = C_L \, L_F^3 \, \log_2^2(W_F L_F) \, \log_2^2(\varepsilon^{-1}). 
$$
The proof is complete, noting that $\tilde{\Phi}_{\varepsilon} \in {\mathcal G}_{\tilde{W},\tilde{L}}^{[-D,D]}$ can be re-parameterized into a standard ReLU network $\Phi_{\varepsilon} \in {\mathcal N}_{W,L}^{[-D,D]}$ with the same output $f_{L_F}$ and the same depth $L=\tilde{L}$, and with a width $W=\tilde{W}+2 \le 18 \, W_F$. 
\end{proof}

\subsubsection{Step 4: Proof of the desired estimate} 

We are now ready to prove Theorem \ref{thm:main_theoretical_result}. 
Let $X=[-D,D]$, with $D \ge 1$. 
By Theorem \ref{thm:ff_compactX}, there exists a residual Fourier network $f_{\Psi}$ with depth $L_{F}$ and width $W_{F}$ satisfying: 
\begin{equation}\label{eq:proof_aux1}
W_{F} \, L_{F} = C \, ||f||_{L^{\infty}(X)}^{2} \, \varepsilon^{-2}, 
\end{equation}
where $C$ is a positive constant that may depend logarithmically on $||\hat{f}||_{L^{1}(\mathbb{R})}$ and $||f||_{L^{\infty}[-D,D]}$, and such that
\begin{equation}\label{eq:half_tol_1}
|| f(x) - f_{\Psi}(x) ||_{L^2[-D,D]} \le \frac{\varepsilon}{2}.
\end{equation}
Then, by Lemma \ref{lem:fourier_relu}, there exists a ReLU network $\Phi \in {\mathcal N}_{W,L}^{[-D,D]}$, satisfying 
\begin{equation}\label{eq:half_tol_2}
|| f_{\Psi}(x) - f_{\Phi}(x) ||_{L^{\infty}[-D,D]} \le \frac{\varepsilon}{2},
\end{equation}
such that
\begin{equation}\label{eq:proof_aux2}
W \le C_{W} \, W_F, \qquad L \le C_{L} \, L_F^3 \, \log_2^2(W_F L_F) \, \log_2^2(\varepsilon^{-1}).
\end{equation}
The desired accuracy \eqref{eqn:main_theoretical_tol} follows from \eqref{eq:half_tol_1} and \eqref{eq:half_tol_2} by the triangle inequality and the compactness of $X=[-D,D]$. To show the desired complexity \eqref{eqn:main_theoretical_complexity_result}, we note that by \eqref{eq:proof_aux1} and using the properties of logarithm, we have:  
$$
\log_2^2 (W_F \, L_F) = \log_2^2 (C \, ||f||_{L^{\infty}(X)}^{2} \, \varepsilon^{-2}) \le C_1 \, \log_2^2 (\varepsilon^{-1}),
$$
where $C_1$ is a positive constant that may depend logarithmically on $||\hat{f}||_{L^{1}(\mathbb{R})}$ and $||f||_{L^{\infty}[-D,D]}$. By \eqref{eq:proof_aux2}, we hence obtain:
$$
W \le C_{W} \, W_F, \qquad L \le C_L \, C_1 \, L_F^3 \, \log_2^4(\varepsilon^{-1}).
$$
From this and \eqref{eq:proof_aux1}, the desired complexity \eqref{eqn:main_theoretical_complexity_result} follows, and this completes the proof. \qed

\clearpage

\chapter{Further reading and concluding remarks}

\bigskip

The mathematics of deep learning is in its early stages, with foundational results in place but much yet to be explored. This survey primarily focuses on the approximation capabilities of ReLU and Fourier networks for continuous and bounded (possibly discontinuous) functions. 
While key concepts and recent findings in the approximation theory of neural networks have been introduced, many promising directions for future research remain. Further investigations into advanced neural network architectures and approximation methods could significantly enhance our understanding of this rapidly evolving field. The strategies discussed here can help derive new results by designing more sophisticated architectures, addressing questions such as error and complexity estimates for different network types and a broader range of function classes.

Neural network approximation is a nonlinear method and should be compared with other nonlinear techniques. Two important references are \cite{Daubechies_etal:21}, which compares ReLU networks with free-knot splines and N-term Fourier-like approximations, and \cite{Elbrachter_etal:21}, which provides an overview of Kolmogorov-Donoho nonlinear approximation, linking function class complexity to network complexity. For deeper insights into these topics, familiarity with nonlinear approximation (see \cite{DeVore_Acta:98}) is recommended.

We highlight two important topics that have not been addressed in detail: i) the curse of dimensionality, and ii) spectral bias. 
The curse of dimensionality refers to the exponential increase in computational complexity as the input dimension grows, which makes high-dimensional problems particularly challenging for approximation methods. However, certain deep network architectures may mitigate this by exploiting structure in the data, such as low-dimensional manifolds or local compositional properties. For example, convolutional neural networks (CNNs), with their local receptive fields and parameter sharing, can effectively capture such structures in high-dimensional tasks. Convolutional networks may also outperform standard feed-forward networks in specific tasks, especially when the target functions possess hierarchical or geometric properties like translation, permutation, or rotational invariance. For further details on these topics, we refer readers to \cite{Mhaskar_Poggio:16,Poggio_Mhaskar_etal:17,bietti2021on,E_etal:20}. 
Spectral bias, on the other hand, refers to the tendency of neural networks to prioritize learning low-frequency components of a function before high-frequency ones. This behavior can influence both the speed of convergence and the accuracy of approximation, particularly for functions with significant high-frequency content. We refer to \cite{Rahaman_etal:19,tancik2020fourier} for a more detailed discussion on spectral bias.

Finally, we note some potential research directions that build on the topics discussed in this survey. One avenue involves exploring the connections between residual Fourier networks and other sinusoidal activation networks, such as sinusoidal representation networks (SIRENs) \cite{sitzmann2020implicit} and Fourier Neural Operators (FNOs) \cite{li2020fourier}. Understanding these relationships may provide further insights into the applicability of sinusoidal activations in diverse settings. Another direction involves the challenge of optimal experimental design: how to construct network architectures and select data effectively for a given target function space to achieve accurate approximations with minimal computational cost. While these directions are intriguing, their broader relevance and impact warrant further exploration.

\clearpage

\hrule
\vspace{0.2cm}
\noindent
{\bf Acknowledgments.} 
This survey draws heavily from the selected topics course on {\it Mathematics of Deep Learning} offered by the second author at The University of New Mexico and Uppsala University, Sweden, from 2021 to 2024. It also incorporates insights from the PhD dissertation of the first author. The authors express their gratitude to all the students and researchers who actively participated in these courses, as well as to colleagues and collaborators whose valuable discussions contributed to the development of this survey. The second author would like to extend a special thanks to Dr. Gunilla Kreiss for her support during several visits to Uppsala University.
\smallskip

Sandia National Laboratories is a multi-mission laboratory managed and operated by National Technology \& Engineering Solutions of Sandia, LLC (NTESS), a wholly owned subsidiary of Honeywell International Inc., for the U.S. Department of Energy’s National Nuclear Security Administration (DOE/NNSA) under contract DE-NA0003525. This written work is authored by an employee of NTESS. The employee, not NTESS, owns the right, title and interest in and to the written work and is responsible for its contents. Any subjective views or opinions that might be expressed in the written work do not necessarily represent the views of the U.S. Government. The publisher acknowledges that the U.S. Government retains a non-exclusive, paid-up, irrevocable, world-wide license to publish or reproduce the published form of this written work or allow others to do so, for U.S. Government purposes. The DOE will provide public access to results of federally sponsored research in accordance with the DOE Public Access Plan.

\vspace{0.2cm}

\hrule

\bigskip
\bibliography{refs.bib}
\bibliographystyle{unsrt}

\end{document}